\documentclass[runningheads]{llncs}
\RequirePackage{silence}  
\usepackage{graphicx}
\usepackage{comment}
\usepackage{amsmath,amssymb}
\usepackage{color}
\usepackage{url}
\usepackage{hyperref}

\usepackage{booktabs}
\usepackage{tikz}
\usepackage{subcaption}
\usepackage{pifont} 
\newcommand{\cmark}{\ding{51}} 
\newcommand{\xmark}{\ding{55}} 

\usepackage{orcidlink}

\usepackage{tikz}
\usepackage{booktabs}
\usepackage{amsmath}
\usepackage{subcaption}
\usepackage{graphicx}
\usetikzlibrary{arrows.meta,positioning,fit,calc,backgrounds}

\definecolor{backbone}{RGB}{52,73,94}
\definecolor{inputc}{RGB}{52,152,219}
\definecolor{distill}{RGB}{46,204,113}
\definecolor{contrast}{RGB}{241,196,15}
\definecolor{regular}{RGB}{230,126,34}
\definecolor{recon}{RGB}{155,89,182}
\definecolor{predict}{RGB}{231,76,60}
\definecolor{neutral}{RGB}{127,140,141}

\tikzset{
  block/.style={rounded corners=2pt, draw=black!70, thick, align=center, font=\scriptsize, minimum height=8mm, text width=0.88\linewidth},
  smallblock/.style={rounded corners=2pt, draw=black!70, align=center, font=\scriptsize, minimum height=6.5mm},
  arrow/.style={-{Latex[length=2mm]}, thick, draw=black!70},
  note/.style={font=\scriptsize, align=left}
}

\newif\ifreview
\reviewfalse

\ifreview
	\usepackage{lineno}

	\linenumbers
\fi

\begin{document}


\def\SubNumber{95}

\def\GCPRTrack{Fast Review Track}

\title{Physics-Aligned Self-Supervised Learning for Scientific Imaging}

\ifreview
\else

	\author{Bashir Kazimi\inst{1}\thanks{Corresponding Author}\orcidlink{0000-0003-1802-7511} \and
	Stefan Sandfeld\inst{1,2}\orcidlink{0000-0001-9560-4728}}
	
	\authorrunning{B. Kazimi and S. Sandfeld}
	
	\institute{Institute for Materials Data Science and Informatics (IAS-9), Forschungszentrum Jülich GmbH, 52425, Jülich, Germany\\ \and Chair of Materials Data Science and Materials Informatics, Faculty 5 – Georesources and Materials Engineering, RWTH Aachen University, 52056, Aachen, Germany\\
	\email{\{b.kazimi,s.sandfeld\}@fz-juelich.de}}
\fi

\maketitle              

\begin{abstract}
Data augmentations define the invariances learned by self-supervised learning (SSL). Standard augmentation pipelines were designed for natural images, yet scientific imaging modalities are governed by physical measurement processes with distinct symmetry and acquisition constraints. Enforcing invariances that contradict these constraints can distort learned representations and limit downstream performance, but practitioners moving from machine learning into a new scientific modality currently have little guidance beyond transferring natural-image pipelines unexamined. We address this gap with a principled, reproducible procedure for augmentation design in scientific SSL: we formalise the physics-aligned augmentation set as a union of measurement-consistent symmetries and acquisition-driven perturbations, and we give a concrete, largely label-free workflow --- enumerate candidates, label each by the measurement operator, validate with representation-geometry diagnostics, and confirm by single-factor ablation --- for selecting them. We instantiate the procedure for real-space electron microscopy and reciprocal-space 4D-STEM diffraction, and evaluate it across five SSL paradigms (DINOv2, SimCLR, MAE, VICRegL, I-JEPA) on classification and crystal-orientation regression. Physics-aligned augmentations substantially improve downstream performance for objectives relying on cross-view consistency, reduce geodesic error and improve robustness under realistic acquisition variability (detector gain, resolution loss), and systematically reshape representation geometry. While our experiments use electron microscopy, the procedure is modality-agnostic and applies to other measurement-driven domains such as medical and remote-sensing imaging. These results position augmentation design as a primary, and controllable, source of inductive bias in scientific self-supervised learning. Code and pretrained models are available on \href{https://github.com/DL4EM/physics-aligned-ssl}{Github} and \href{https://huggingface.co/DL4EM/physics-aligned-ssl}{Hugging Face}.

\end{abstract}

\section{Introduction}
\label{sec:intro}

Self-supervised learning (SSL) enables representation learning without manual annotation and has become dominant in computer vision. A central component is data augmentation: multiple transformed views of an input are generated and the objective enforces consistency between their representations, so augmentations define the invariances the model encodes. This holds across SSL families---contrastive~\cite{chen2020simclr,he2020moco}, teacher--student~\cite{grill2020byol,oquab2024dinov2}, redundancy-reduction~\cite{bardes2022vicreg,bardes2023vicregl}, masked~\cite{he2022mae}, and predictive~\cite{assran2023ijepa}---which differ in objective but share this reliance on augmentation to define invariance.

Standard augmentation pipelines were developed for natural images, where transformations such as cropping, flipping, color jittering, or blur reflect common scene variability. Scientific imaging modalities, however, are governed by structured physical measurement processes: in electron microscopy (EM) and diffraction imaging, observed signals arise from interactions between the imaging system and the underlying physical structure, so transformations benign for photographs may be physically invalid (e.g., arbitrary flips or photometric changes can contradict the geometry of reciprocal-space diffraction patterns or the mapping between image structure and crystal orientation). Although SSL is increasingly applied to scientific and medical imaging, augmentation pipelines are often transferred directly from natural-image practice~\cite{huang2023medicalsslreview,azizi2021bigselfsupervised,kazimi2024self}. Prior work shows that augmentation choice strongly influences the invariances learned by SSL models~\cite{tian2020views} and that representation quality can be characterized through geometric properties such as alignment and uniformity~\cite{wang2020alignment}, but these studies focus on natural images and do not consider whether augmentation-induced invariances remain valid under domain-specific measurement constraints.

The problem is neither niche nor trivial. It is not niche because the same issue arises wherever SSL meets a measurement-driven modality---MRI/CT (reconstructed from $k$-space and projection data), ultrasound, remote sensing, and astronomical imaging are all governed by acquisition physics under which common natural-image augmentations may be invalid. It is not trivial because the ``correct'' augmentation set cannot be read off the data: it depends on the measurement operator and the downstream task, so a transformation can be visually innocuous yet destroy task-relevant structure (e.g., a spatial flip silently relabels a crystal orientation). The practitioner therefore needs a \emph{procedure} for deciding which transformations are admissible, not a longer list of augmentations.

This raises a central question for scientific SSL: \emph{how does mismatch between augmentation design and measurement physics affect learned representations?} We hypothesize that augmentations violating measurement symmetries impose incorrect invariances---degrading representation quality, downstream performance, and robustness to physically meaningful distribution shifts---while augmentations aligned with the measurement process provide a physically grounded inductive bias that preserves task-relevant structure. To study this, we investigate augmentation--physics alignment across two complementary modalities, real-space electron microscopy and reciprocal-space 4D-STEM diffraction, introducing physics-aligned pipelines that encode measurement-consistent transformations and acquisition variability (physically motivated noise, reciprocal-space scaling from camera-length variation, diffraction tilt) while excluding transformations that violate measurement constraints. We evaluate five representative SSL paradigms---DINOv2~\cite{oquab2024dinov2}, SimCLR~\cite{chen2020simclr}, MAE~\cite{he2022mae}, VICRegL~\cite{bardes2022vicreg,bardes2023vicregl}, and I-JEPA~\cite{assran2023ijepa}---under both regimes, spanning downstream classification in real-space EM and crystal-orientation regression in 4D-STEM, and analyze representation geometry using effective rank, embedding uniformity, collapse ratio, and kNN evaluation.

\noindent\textbf{Contributions.}
This is an empirical methodology study, not a new SSL algorithm or a leaderboard benchmark: our goal is a principled, reproducible procedure for choosing self-supervised augmentations in scientific imaging. We \textbf{(i)} formalise augmentation design as aligning enforced invariances with the measurement process, via the decomposition $\mathcal{T}_{\text{phys}} = \mathcal{T}_{\text{sym}} \cup \mathcal{T}_{\text{acq}}$ where membership is decided by the imaging operator (commutes with the measurement $\rightarrow \mathcal{T}_{\text{sym}}$; models an acquisition parameter $\rightarrow \mathcal{T}_{\text{acq}}$; else excluded); \textbf{(ii)} provide a practitioner's procedure (Section~\ref{sec:procedure}) that turns this into actionable steps---label candidates by the operator, screen with \emph{label-free} diagnostics, confirm by single-factor ablation before committing compute; \textbf{(iii)} quantify the consequences of augmentation--physics (mis)alignment across five SSL paradigms and two complementary modalities, via linear-probe, low-label, robustness, and representation-geometry analyses; and \textbf{(iv)} show the procedure is modality-agnostic, applying to any measurement-driven domain (MRI/CT, ultrasound, remote sensing, astronomy) and giving practitioners a concrete starting point rather than ad hoc trial and error.

\section{Augmentation as Inductive Bias in Scientific SSL}

SSL typically learns representations by enforcing consistency between augmented views. Let $x \in \mathcal{X}$ be an image and $f_\theta:\mathcal{X}\rightarrow\mathbb{R}^d$ a representation function; during training, two transformations $t_1,t_2 \sim \mathcal{T}$ are applied to $x$ and the objective encourages similar representations:

\begin{equation}
f_\theta(t_1(x)) \approx f_\theta(t_2(x)), \quad t_1,t_2 \sim \mathcal{T}.
\end{equation}

However the objective is implemented (contrastive, teacher--student, redundancy reduction, masked reconstruction), the augmentation distribution $\mathcal{T}$ defines which transformations are treated as invariant, acting as an implicit inductive bias~\cite{tian2020views,wang2020alignment}.

\subsection{Physical Symmetry and Measurement Constraints}

In scientific imaging, valid invariances are determined by the measurement physics: images arise from structured interactions between the imaging system and the sample, so only certain transformations correspond to physically meaningful variability. We distinguish $\mathcal{T}_{\text{orig}}$ (standard natural-image augmentations---cropping, flipping, blur, photometric perturbations) from $\mathcal{T}_{\text{phys}}$, which excludes transformations that violate measurement constraints (e.g., arbitrary flips inconsistent with reciprocal-space diffraction geometry) and retains physically meaningful variability:

\begin{itemize}
    \item \textbf{Real-space EM:} valid transformations include sample rotations and moderate contrast/illumination changes; electron-counting statistics and detector electronics motivate controlled rotations, bounded intensity scaling, and physically motivated noise.
    \item \textbf{Reciprocal-space 4D-STEM:} diffraction patterns depend on beam and detector geometry, where camera length and beam tilt produce structured reciprocal-space changes (isotropic scaling, peak distortions)~\cite{ophus2019fourdstem}; arbitrary spatial flips can be physically invalid when predicting crystal orientation.
\end{itemize}

We therefore decompose the physics-aligned set as
\begin{equation}
\mathcal{T}_{\text{phys}} = \mathcal{T}_{\text{sym}} \cup \mathcal{T}_{\text{acq}},
\end{equation}
where $\mathcal{T}_{\text{sym}}$ contains symmetry-consistent transformations (e.g., rotations when valid) and $\mathcal{T}_{\text{acq}}$ models acquisition variability (noise, camera-length scaling, diffraction tilt).

\subsection{A Procedure for Physics-Aligned Augmentation Design}
\label{sec:procedure}
 
The decomposition above is not merely descriptive; it defines a concrete, reproducible procedure that a practitioner without prior augmentation intuition for a new modality can follow. We make it explicit, as it is the methodological core of this work; crucially, no step requires downstream labels until the final validation, which matters where annotation is scarce.

\begin{enumerate}
\item \textbf{Specify the measurement operator.} Note how the observed signal is formed from the physical sample (e.g., a real-space projection for EM, a Fourier-domain transform for 4D-STEM diffraction)---domain knowledge the practitioner already has or can obtain from instrument documentation.
\item \textbf{Enumerate a candidate pool.} Begin from a standard natural-image pool (crops, flips, rotations, photometric jitter, blur, noise) plus any modality-specific perturbations of interest (e.g., detector noise, camera-length scaling).
\item \textbf{Label each candidate by the operator.} Ask whether the transform commutes with the measurement ($\rightarrow \mathcal{T}_{\text{sym}}$), models a physical acquisition parameter ($\rightarrow \mathcal{T}_{\text{acq}}$), or neither (exclude). This test is operational and falsifiable: it depends on the operator and sample symmetry, not on taste.
\item \textbf{Validate label-free.} Pretrain briefly and inspect representation-geometry diagnostics (effective rank, uniformity, collapse ratio, kNN), which expose invariance mismatch \emph{without} downstream labels (Section~\ref{subsec:reprgeom}).
\item \textbf{Confirm by single-factor ablation.} Add or remove one transform at a time, holding all else fixed (Section~\ref{subsec:ablation}); this isolates which choices help and surfaces interaction effects that make individually plausible transforms jointly harmful.
\item \textbf{Iterate, then commit.} Refine the pipeline before spending compute on full downstream finetuning.
\end{enumerate}

This procedure is \emph{not} an automated augmentation-search algorithm (learning policies under measurement constraints is future work), nor a claim that domain expertise can be eliminated---as in loss, architecture, or hyperparameter choice, the practitioner must understand their problem. What it supplies is the missing \emph{structure} that turns that expertise into a reproducible, falsifiable workflow, replacing the current default of transferring natural-image pipelines unexamined. The remainder of the paper instantiates it for two modalities and quantifies the difference it makes.

\section{Experimental Setup}

We evaluate augmentation–physics alignment across two scientific imaging modalities using five representative SSL methods. This section describes datasets, tasks, augmentation regimes, and training protocols.

\subsection{Datasets and Tasks}

\paragraph{Real-Space Electron Microscopy.}
Models are pretrained on \textbf{CEM500K}~\cite{Conrad2021} and evaluated on the \textbf{NFFA} dataset~\cite{aversa2018nffa} for multi-class classification using top-1 accuracy. We additionally report a supervised \emph{scratch} baseline trained directly on NFFA.

\paragraph{Reciprocal-Space 4D-STEM.}
Models are pretrained on simulated \textbf{4D-STEM diffraction data}~\cite{scheunert_2025_cnns,scheunert2026determining} and evaluated on a quaternion regression task for crystal orientation estimation. Performance is measured using mean geodesic error (degrees) and angular accuracy thresholds (Acc@5$^\circ$, Acc@10$^\circ$). A supervised scratch baseline is also included. These modalities provide complementary evaluation settings: real-space semantic classification and reciprocal-space orientation regression.

\subsection{Self-Supervised Learning Methods}

We evaluate five SSL methods: SimCLR~\cite{chen2020simclr}, DINOv2~\cite{oquab2024dinov2}, VICRegL~\cite{bardes2023vicregl}, MAE~\cite{he2022mae}, and I-JEPA~\cite{assran2023ijepa}, representing contrastive learning, teacher–student joint embedding, redundancy reduction, masked modeling, and predictive joint embedding. All models use identical backbone architectures within each modality and are finetuned for the downstream tasks.

\subsection{Augmentation Regimes}

Following Section~\ref{sec:intro}, we compare two pretraining regimes: the natural-image pipeline $\mathcal{T}_{\text{orig}}$ (random cropping, horizontal flipping, Gaussian blur, photometric perturbations) and the physics-aligned pipeline $\mathcal{T}_{\text{phys}}$, which removes transformations violating measurement constraints and adds acquisition-aware perturbations. For real-space EM, $\mathcal{T}_{\text{phys}}$ adds controlled rotations, bounded intensity scaling, Gaussian/Poisson noise, and domain artifacts (scanline dropout, charging streaks). For 4D-STEM, which must preserve the diffraction-to-orientation mapping, spatial flips are removed and reciprocal-space scaling (camera-length variation), diffraction tilt (beam misalignment), and physically consistent noise are added. Table~\ref{tab:augmentations} summarizes the components (full parameters in the supplementary); all other hyperparameters are identical across regimes to isolate the effect of augmentation alignment.

\begin{table}[htb!]
\centering
\caption{Augmentation pipelines used during SSL pretraining for real-space EM
and reciprocal-space 4D-STEM data.}
\label{tab:augmentations}
\setlength{\tabcolsep}{5pt}
\begin{tabular}{lcccc}
\toprule
 & \multicolumn{2}{c}{Real-space EM} & \multicolumn{2}{c}{4D-STEM} \\
\cmidrule(lr){2-3} \cmidrule(lr){4-5}
Augmentation & $\mathcal{T}_{\text{orig}}$ & $\mathcal{T}_{\text{phys}}$ & $\mathcal{T}_{\text{orig}}$ & $\mathcal{T}_{\text{phys}}$ \\
\midrule
Random crop & \cmark & \cmark & \cmark & \cmark \\
Horizontal flip & \cmark & \xmark & \cmark & \xmark \\
Vertical flip & \xmark & \xmark & \xmark & \xmark \\
Rotation ($90^\circ$) & \xmark & \cmark & \xmark & \cmark \\
Gaussian blur & \cmark & \cmark & \cmark & \cmark \\
Intensity scaling & \cmark & \cmark & \cmark & \cmark \\
Intensity bias & \xmark & \cmark & \xmark & \cmark \\
Gaussian noise & \xmark & \cmark & \xmark & \cmark \\
Poisson noise & \xmark & \cmark & \xmark & \cmark \\
Solarization & \cmark & \xmark & \cmark & \xmark \\
\midrule
Brightness/contrast jitter & \xmark & \cmark & \xmark & \cmark \\
Reciprocal-space scaling & \xmark & \xmark & \xmark & \cmark \\
Diffraction tilt & \xmark & \xmark & \xmark & \cmark \\
Scanline artifact & \xmark & \cmark & \xmark & \xmark \\
Charging artifact & \xmark & \cmark & \xmark & \xmark \\
\bottomrule
\end{tabular}
\end{table}

\subsection{Training Protocol}

SSL models are pretrained independently under $\mathcal{T}_{\text{orig}}$ and $\mathcal{T}_{\text{phys}}$. Downstream evaluation uses full finetuning on labeled data. Results are reported as mean $\pm$ standard deviation over three random seeds. Supervised models trained from scratch are included for comparison.

\section{Results}

We evaluate augmentation--physics alignment on downstream performance, then analyze representation geometry, augmentation ablations, and robustness under physically meaningful distribution shifts. Physics-aligned augmentations generally improve performance for objectives relying on cross-view consistency (e.g., DINOv2, I-JEPA), while contrastive and variance-regularized methods show smaller changes, indicating that augmentation design interacts with SSL objective structure.

\subsection{Downstream Performance: Real-Space Classification}
\label{subsec:dsresultsrealspace}

Table~\ref{tab:nffa_results} reports NFFA classification accuracy after pretraining on CEM500K under natural-image augmentations ($\mathcal{T}_{\text{orig}}$) and physics-aligned augmentations ($\mathcal{T}_{\text{phys}}$). Physics-aligned augmentations improve performance for most SSL objectives. DINOv2 shows the largest gain (66.70$\rightarrow$76.67), while VICRegL and I-JEPA exhibit smaller but consistent improvements. MAE also improves substantially and shows reduced variance across training seeds, indicating more stable training under $\mathcal{T}_{\text{phys}}$.

In contrast, SimCLR remains largely unchanged between augmentation regimes, suggesting that the contrastive objective is comparatively insensitive to these invariance differences in this real-space classification setting. Relative to training from scratch, SSL pretraining is beneficial overall but depends on the augmentation regime. DINOv2 and MAE trained with $\mathcal{T}_{\text{orig}}$ perform close to the supervised baseline, whereas physics-aligned augmentations move both clearly above it. SimCLR substantially outperforms the scratch model under both regimes.


\begin{table}[t]
\centering
\caption{Downstream classification on NFFA after pretraining on CEM500K. Accuracy is reported as mean$\pm$std over seeds (in \%). Higher is better.}
\label{tab:nffa_results}
\setlength{\tabcolsep}{6pt}
\footnotesize
\begin{tabular}{lccc}
\toprule
Method & Original & Physics-aligned & $\Delta$ (Phys$-$Orig) \\
\midrule
Scratch & \multicolumn{3}{c}{66.59$\pm$2.76} \\
\midrule
DINOv2  & 66.70$\pm$0.83 & 76.67$\pm$0.39 & +9.97 \\
I-JEPA  & 72.27$\pm$1.15 & 73.41$\pm$0.75 & +1.14 \\
MAE     & 67.29$\pm$12.72 & 79.10$\pm$3.07 & +11.81 \\
SimCLR  & 89.68$\pm$0.37 & 89.39$\pm$0.16 & -0.29 \\
VICRegL & 69.83$\pm$2.68 & 73.42$\pm$3.00 & +3.59 \\
\bottomrule
\end{tabular}
\end{table}


To assess representation quality independently of finetuning, we additionally
evaluate a frozen-encoder linear probe on NFFA using the same CEM500K
checkpoints (Table~\ref{tab:nffa_linprobe}). Physics-aligned augmentations
match or improve $\mathcal{T}_{\text{orig}}$ for every method. The gains are
smaller than under finetuning, consistent with the view that for some
objectives the benefit of physically consistent invariances is realised
during task-specific adaptation. The exception is VICRegL, whose
covariance-regularised features are already linearly structured and for which
the linear probe ($77.2\%$) exceeds finetuning ($73.4\%$) --- a known property
of variance--covariance objectives. Taken together with the kNN results (Section~\ref{subsec:reprgeom}), frozen-feature, linear-probe, and finetuning evaluations form a consistent picture in which no method is clearly harmed by $\mathcal{T}_{\text{phys}}$.
 
\begin{table}[htb!]
\centering
\caption{Additional NFFA evaluations (top-1 accuracy \%, mean$\pm$std over 3 seeds), using the same pretrained checkpoints as Table~\ref{tab:nffa_results}. \emph{Left}: frozen-encoder linear probing. \emph{Right}: finetuning with only 25\% of labels (Scratch@25\%: 59.42$\pm$2.70).}
\label{tab:nffa_linprobe}
\setlength{\tabcolsep}{5pt}
\footnotesize
\begin{tabular}{l|ccc|ccc}
\toprule
 & \multicolumn{3}{c|}{Linear probe} & \multicolumn{3}{c}{Finetune @ 25\% labels} \\
Method & Orig & Phys & $\Delta$ & Orig & Phys & $\Delta$ \\
\midrule
DINOv2  & 42.96$\pm$1.0 & 43.00$\pm$0.4 & +0.0 & 56.44$\pm$1.3 & 64.59$\pm$0.7 & +8.2 \\
I-JEPA  & 41.09$\pm$0.7 & 41.01$\pm$0.8 & $-$0.1 & 63.60$\pm$2.5 & 64.54$\pm$0.5 & +0.9 \\
MAE     & 57.04$\pm$0.7 & 58.55$\pm$0.9 & +1.5 & 68.63$\pm$7.5 & 71.98$\pm$1.0 & +3.4 \\
SimCLR  & 72.46$\pm$0.9 & 73.14$\pm$0.9 & +0.7 & 80.76$\pm$0.9 & 80.42$\pm$1.0 & $-$0.3 \\
VICRegL & 76.20$\pm$0.6 & 77.17$\pm$0.8 & +1.0 & 62.91$\pm$2.8 & 71.22$\pm$7.4 & +8.3 \\
\bottomrule
\end{tabular}
\end{table}

\paragraph{Low-label finetuning}:
Because labelled data is often scarce in scientific imaging, we repeat NFFA finetuning using only 25\% of the training labels (Table~\ref{tab:nffa_linprobe}, right). Physics-aligned augmentations retain their advantage in this regime and additionally stabilise training (e.g., MAE standard deviation $\pm7.5$ under $\mathcal{T}_{\text{orig}}$ vs.~$\pm1.0$ under $\mathcal{T}_{\text{phys}}$). The supervised scratch baseline falls from $66.59\%$ (full labels) to $59.42\%$, while SSL$+\mathcal{T}_{\text{phys}}$ remains well above it, and the relative ranking is preserved from the full-label setting.

\subsection{Downstream Performance: Reciprocal-Space Orientation Regression}
\begin{figure}[htb!]
\centering
\begin{subfigure}[t]{0.48\linewidth}
\centering
\includegraphics[width=\linewidth]{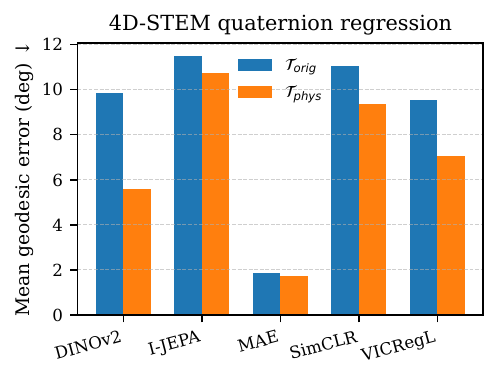}
\caption{Mean geodesic error.}
\label{fig1}
\end{subfigure}
\hfill
\begin{subfigure}[t]{0.48\linewidth}
\centering
\includegraphics[width=\linewidth]{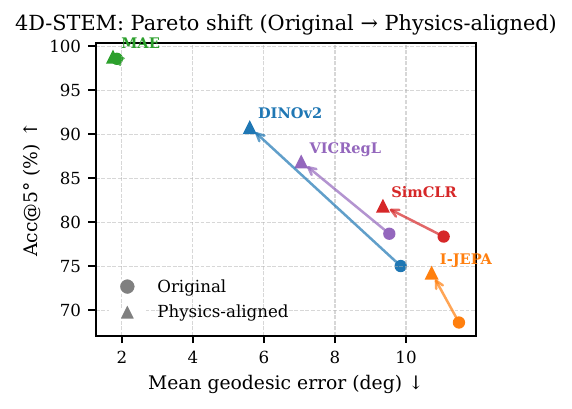}
\caption{Error–accuracy tradeoff.}
\label{fig2}
\end{subfigure}
\caption{Quaternion regression results on 4D-STEM under natural-image augmentations ($\mathcal{T}_{\rm orig}$) and physics-aligned augmentations ($\mathcal{T}_{\rm phys}$). Physics-aligned augmentations consistently improve performance, particularly for teacher–student objectives such as DINOv2.}
\label{fig:quat_results}
\end{figure}

Table~\ref{tab:quat_results} reports quaternion regression on 4D-STEM data. Augmentation alignment has a stronger effect here than in real-space classification. DINOv2 shows the largest improvement, reducing mean geodesic error from $9.85^\circ$ to $5.60^\circ$; VICRegL and SimCLR also improve under $\mathcal{T}_{\text{phys}}$, and I-JEPA shows smaller but consistent gains. MAE behaves differently, achieving geodesic error near $2^\circ$ under both regimes. We attribute this near-constant performance to the nature of its objective rather than to a trivial task: pixel reconstruction does not enforce cross-view invariance, so augmentation choice barely shapes its invariance set. The task itself is not trivial---the supervised scratch baseline reaches only $10.01^\circ$ and the cross-view objectives (SimCLR, I-JEPA) hover near it under $\mathcal{T}_{\text{orig}}$---but masked reconstruction happens to be a near-perfect inductive bias for orientation regression on these diffraction patterns, where every Bragg peak is informative. Consistent with this, MAE still degrades less under blur when pretrained with $\mathcal{T}_{\text{phys}}$ (supplementary; $5.71^\circ$ vs.\ $7.00^\circ$ at $\sigma{=}2$). By contrast, cross-view objectives (DINOv2, SimCLR, VICRegL) rely on invariance between augmented views: when augmentations distort reciprocal-space geometry they suppress orientation-sensitive structure, and physics-aligned augmentations mitigate this. Figures~\ref{fig1} and~\ref{fig2} visualize the error reduction and the shift in the geodesic--accuracy plane; the largest improvements are for DINOv2 and VICRegL.
\begin{table}[htb!]
\centering
\caption{Quaternion regression on 4D-STEM. Mean geodesic error (degrees) and angular accuracy (\%). Lower error and higher accuracy are better.}
\label{tab:quat_results}
\setlength{\tabcolsep}{4pt}
\resizebox{\columnwidth}{!}{%
\begin{tabular}{lccc|cc|cc}
\toprule
 & \multicolumn{3}{c|}{Geodesic Error ($^\circ$)} & \multicolumn{2}{c|}{Acc@5$^\circ$} & \multicolumn{2}{c}{Acc@10$^\circ$} \\
\cmidrule(lr){2-4} \cmidrule(lr){5-6} \cmidrule(lr){7-8}
Method & Orig & Phys & $\Delta$ & Orig & Phys & Orig & Phys \\
\midrule
Scratch
& 10.01$\pm$1.89 & -- & -- 
& 75.61$\pm$6.26 & -- 
& 84.93$\pm$3.69 & -- \\
\midrule
DINOv2
& 9.85$\pm$0.96 & 5.60$\pm$0.58 & +4.25
& 75.03$\pm$5.84 & 90.81$\pm$1.42
& 84.85$\pm$1.96 & 93.26$\pm$0.99 \\
I-JEPA
& 11.49$\pm$0.51 & 10.72$\pm$0.51 & +0.77
& 68.72$\pm$3.07 & 74.28$\pm$3.15
& 81.24$\pm$1.78 & 83.38$\pm$1.00 \\
MAE
& 1.86$\pm$0.01 & 1.75$\pm$0.02 & +0.11
& 98.58$\pm$1.08 & 98.83$\pm$0.65
& 98.85$\pm$1.05 & 99.03$\pm$0.83 \\
SimCLR
& 11.06$\pm$0.32 & 9.35$\pm$0.93 & +1.71
& 78.38$\pm$1.71 & 81.88$\pm$3.16
& 83.56$\pm$0.58 & 86.59$\pm$1.91 \\
VICRegL
& 9.53$\pm$1.97 & 7.05$\pm$1.09 & +2.48
& 78.7$\pm$5.7 & 86.91$\pm$3.19
& 85.44$\pm$4.65 & 90.67$\pm$1.88 \\
\bottomrule
\end{tabular}
}
\end{table}

\subsection{Representation Geometry Analysis}\label{subsec:reprgeom}

To understand the mechanism behind these gains---and to instantiate the label-free validation step of our procedure---we analyze pretrained-embedding geometry using four metrics: effective rank (spectral utilization), uniformity~\cite{wang2020alignment} (dispersion on the hypersphere), collapse ratio (variance in the top principal component), and kNN accuracy (frozen-feature separability; classification only). Full per-method tables for both modalities are in the supplementary; we summarize the findings here.

\paragraph{NFFA Representation Geometry:}

Physics-aligned augmentations slightly increase effective rank for teacher--student and predictive objectives (DINOv2: $1.4\rightarrow1.8$; I-JEPA: $4.5\rightarrow4.7$) and more strongly for MAE ($29.5\rightarrow36.6$), while contrastive and redundancy-reduction methods (SimCLR, VICRegL) show higher effective rank under natural-image augmentations---an objective-dependent response. They also reduce collapse for DINOv2 ($0.964\rightarrow0.917$), I-JEPA ($0.557\rightarrow0.519$), and MAE ($0.500\rightarrow0.443$), coinciding with improved accuracy for several methods (though not universally: SimCLR is strong despite modest geometric change). Uniformity magnitudes are set by each objective's loss---most strongly by VICRegL, whose covariance term explicitly maximizes dispersion---so uniformity is a within-method comparison across regimes, not an across-method ranking. Frozen-feature kNN partly mirrors finetuning: SimCLR ($0.679\rightarrow0.707$) and I-JEPA improve under $\mathcal{T}_{\text{phys}}$, whereas DINOv2 and MAE are slightly higher under natural-image augmentations despite improving after finetuning---indicating that for these two methods the benefit emerges during task-specific adaptation rather than from frozen-feature separability, consistent with the linear-probe results.

\paragraph{4D-STEM Representation Geometry:}

The same objective-dependent pattern holds for reciprocal-space embeddings used in quaternion regression (full table in the supplementary). Augmentation alignment again produces the strongest geometric effects for teacher--student methods: for DINOv2, physics-aligned augmentations substantially increase effective rank ($1.2\rightarrow3.1$) while reducing collapse ($0.975\rightarrow0.830$), coinciding with improved regression accuracy. Masked modeling behaves differently---for MAE, natural-image augmentations yield higher effective rank, consistent with augmentation playing a smaller role when representations are learned through reconstruction---and objectives with explicit variance or redundancy regularization (SimCLR, VICRegL) already maintain high-rank, low-collapse embeddings, so alignment produces smaller changes.

\paragraph{Cross-Objective Interpretation:}

Overall, augmentation alignment affects representation geometry in objective-dependent ways. Methods relying on cross-view consistency (e.g., DINOv2) exhibit the largest geometric changes, particularly in reciprocal-space representations, while predictive approaches such as I-JEPA show smaller but consistent shifts. In contrast, contrastive and redundancy-reduction objectives already enforce dispersion through negative sampling or covariance regularization, limiting the effect of augmentation alignment. The supplementary material provides the full 4D-STEM geometry table and qualitative examples for all three downstream analyses.

\subsection{Robustness to Acquisition Variability}

Scientific imaging exhibits variability in acquisition conditions (detector gain, exposure, resolution). We assess robustness under controlled test-time perturbations---global intensity scaling ($g \in \{0.8,\dots,1.2\}$, simulating detector gain/dose) and Gaussian blur ($\sigma \in \{0,\dots,2.0\}$, simulating resolution loss)---evaluating finetuned models without retraining and also reporting degradation normalized to the nominal setting ($g{=}1.0$, $\sigma{=}0$) to isolate robustness from absolute performance.

Under intensity gains, most objectives pretrained with $\mathcal{T}_{\text{phys}}$ maintain lower geodesic error than natural-image ones (e.g., DINOv2 stays near $5^\circ$ across the range while $\mathcal{T}_{\text{orig}}$ remains above $10^\circ$; I-JEPA improves similarly, SimCLR modestly). Normalized curves (supplementary) remain nearly flat for most $\mathcal{T}_{\text{phys}}$ models, indicating near-invariance to intensity changes and representations driven by diffraction structure rather than absolute magnitude.

We next evaluate robustness to Gaussian blur, which approximates resolution degradation due to detector point-spread or defocus, across blur levels $\sigma \in [0,2]$. As blur increases, orientation prediction becomes harder for all models, but $\mathcal{T}_{\text{phys}}$ models degrade more gracefully: DINOv2 with physics-aligned augmentations rises from roughly $5^\circ$ at $\sigma=0$ to about $16^\circ$ at $\sigma=2$, whereas the natural-image model rises from about $11^\circ$ to more than $26^\circ$. Normalized degradation curves (supplementary) show slower error growth for most $\mathcal{T}_{\text{phys}}$ models.

\paragraph{Summary}

Overall, physics-aligned augmentations improve robustness to acquisition variability, yielding lower error under perturbations and reduced sensitivity to degradation (raw and normalized curves for both perturbations are provided in the supplementary). These results suggest that enforcing physically consistent invariances encourages representations that capture diffraction structure rather than incidental imaging artifacts.

\begin{table}[htb!]
\centering
\caption{Single-factor augmentation ablation on NFFA classification using DINOv2 pretrained on CEM500K. Starting from the original augmentation pipeline $\mathcal{T}_{\text{orig}}$, we introduce or remove individual transformations and evaluate the effect on downstream accuracy. Reported accuracies are averaged over three runs.}
\label{tab:augmentation_ablation}
\small
\setlength{\tabcolsep}{6pt}
\begin{tabular}{p{0.72\linewidth}c}
\toprule
Augmentation configuration & Accuracy (\%) \\
\midrule
Scratch (supervised) & 66.59 \\
$\mathcal{T}_{\text{orig}}$ & 66.70 \\
\midrule
$\mathcal{T}_{\text{orig}}$ + Gaussian noise & 67.01 \\
$\mathcal{T}_{\text{orig}}$ + Rotation $90^\circ$ & 57.75 \\
$\mathcal{T}_{\text{orig}}$ + Vertical flip & 31.32 \\
\midrule
$\mathcal{T}_{\text{orig}}$ + Brightness/contrast jitter & 77.32 \\
$\mathcal{T}_{\text{orig}}$ + Scanline dropout & 63.46 \\
$\mathcal{T}_{\text{orig}}$ + Charging artifact & 46.19 \\
$\mathcal{T}_{\text{orig}}$ + Charging artifact + Brightness/contrast jitter & 76.04 \\
$\mathcal{T}_{\text{orig}}$ + Scanline dropout + Brightness/contrast jitter & 76.6 \\
$\mathcal{T}_{\text{orig}}$ + Scanline dropout + Charging artifact & 62.98 \\
\midrule
$\mathcal{T}_{\text{phys}}$ core: $-$flip $-$solarize $+$noise (Gauss/Poisson) $+$intensity bias $+$bright/contrast & 76.67 \\
\;\;+ rotation $90^\circ$ (best) & 80.36 \\
\;\;+ rotation $90^\circ$, keeping solarization & 72.75 \\
\bottomrule
\end{tabular}
\end{table}
\subsection{Interventional Augmentation Ablation}\label{subsec:ablation}
To identify which augmentations drive the improvements from physics-aligned training, we perform a single-factor ablation using DINOv2. Starting from the original pipeline $\mathcal{T}_{\text{orig}}$, we introduce or remove individual transformations, holding all else fixed, and evaluate NFFA classification accuracy (Table~\ref{tab:augmentation_ablation}). The baseline model trained with $\mathcal{T}_{\text{orig}}$ achieves $66.70\%$, similar to the supervised scratch baseline ($66.59\%$). Adding Gaussian noise alone has little effect ($67.01\%$).
In contrast, augmentations that conflict with measurement constraints degrade performance: a $90^\circ$ rotation reduces accuracy to $57.75\%$ and vertical flips collapse it to $31.32\%$. Augmentations reflecting realistic acquisition variability show the opposite trend---brightness/contrast jitter improves accuracy to $77.32\%$---while domain artifacts (scanline dropout, charging streaks) hurt when applied alone but are largely mitigated when combined with brightness perturbations. Progressively modifying $\mathcal{T}_{\text{orig}}$ toward the physics-aligned pipeline (removing horizontal flips and solarization, adding noise and intensity perturbations) yields $76.67\%$; adding further acquisition variability produces the best result ($80.36\%$).

The rotation result illustrates why single-factor ablation is part of the procedure. Adding rotation to $\mathcal{T}_{\text{orig}}$ lowers accuracy ($57.75\%$), yet rotation is in the physics-aligned pipeline. The resolution is interaction: $\mathcal{T}_{\text{orig}}$ still contains horizontal flip and solarization, so adding rotation enforces the full reflection group while solarization imposes a non-physical inversion. Cellular EM is rotation-equivariant, so rotation alone is valid; once the conflicting transforms are removed, adding rotation gives the strongest configuration ($80.36\%$). A transformation can thus be individually valid yet harmful in combination---the failure mode the ablation is designed to expose, and why the procedure validates the \emph{pipeline} rather than transforms in isolation.

Overall, the ablation confirms that improvements arise from enforcing invariances consistent with imaging physics: transformations that conflict with measurement constraints degrade representation quality, while acquisition-consistent perturbations improve self-supervised pretraining.

\section{Discussion and Conclusion}

Our results show that augmentation design acts as a primary inductive bias in scientific SSL, and---more importantly---that it can be controlled by the reproducible procedure of Section~\ref{sec:procedure} rather than by intuition. Instantiating it for two modalities yields consistent gains for cross-view-consistency objectives (e.g., DINOv2: $+9.97$ points on NFFA, $9.85^\circ\rightarrow5.60^\circ$ on 4D-STEM), while reconstruction- or dispersion-based methods (SimCLR, MAE) are less sensitive; beyond accuracy, alignment reduces collapse, improves spectral utilization, and increases robustness to acquisition variability. The study is limited to two modalities and a fixed augmentation pool; natural extensions are learning augmentation policies under measurement constraints and testing transfer to dense prediction tasks (segmentation, detection, tracking).

%
%
%
%
\bibliographystyle{splncs04}
\bibliography{egbib}

\clearpage
\appendix
\renewcommand{\thesubsection}{S\arabic{section}.\arabic{subsection}}
\setcounter{section}{0}
\renewcommand{\thesection}{S\arabic{section}}
\renewcommand{\thefigure}{S\arabic{figure}}
\renewcommand{\thetable}{S\arabic{table}}
\setcounter{figure}{0}
\setcounter{table}{0}
\renewcommand{\theequation}{S\arabic{equation}}
\setcounter{equation}{0}

\section*{Supplementary Material}
\addcontentsline{toc}{section}{Supplementary Material}

This supplementary material provides additional dataset examples, implementation details, augmentation visualizations, qualitative results, training dynamics, additional quantitative results, full robustness tables, and representation visualizations supporting the results in the main paper titled \textit{Physics-Aligned Self-Supervised Learning for Scientific Imaging}. Code and pretrained models are available on \href{https://github.com/DL4EM/physics-aligned-ssl}{Github} and \href{https://huggingface.co/DL4EM/physics-aligned-ssl}{Hugging Face}. \\

\section{Experiment Details}

\subsection{Dataset Overview}
\label{sec:supp_dataset_overview}

This section provides additional details about the datasets used for self-supervised pretraining and downstream evaluation.

\begin{figure*}[htb!]
\centering
\includegraphics[width=0.95\textwidth]{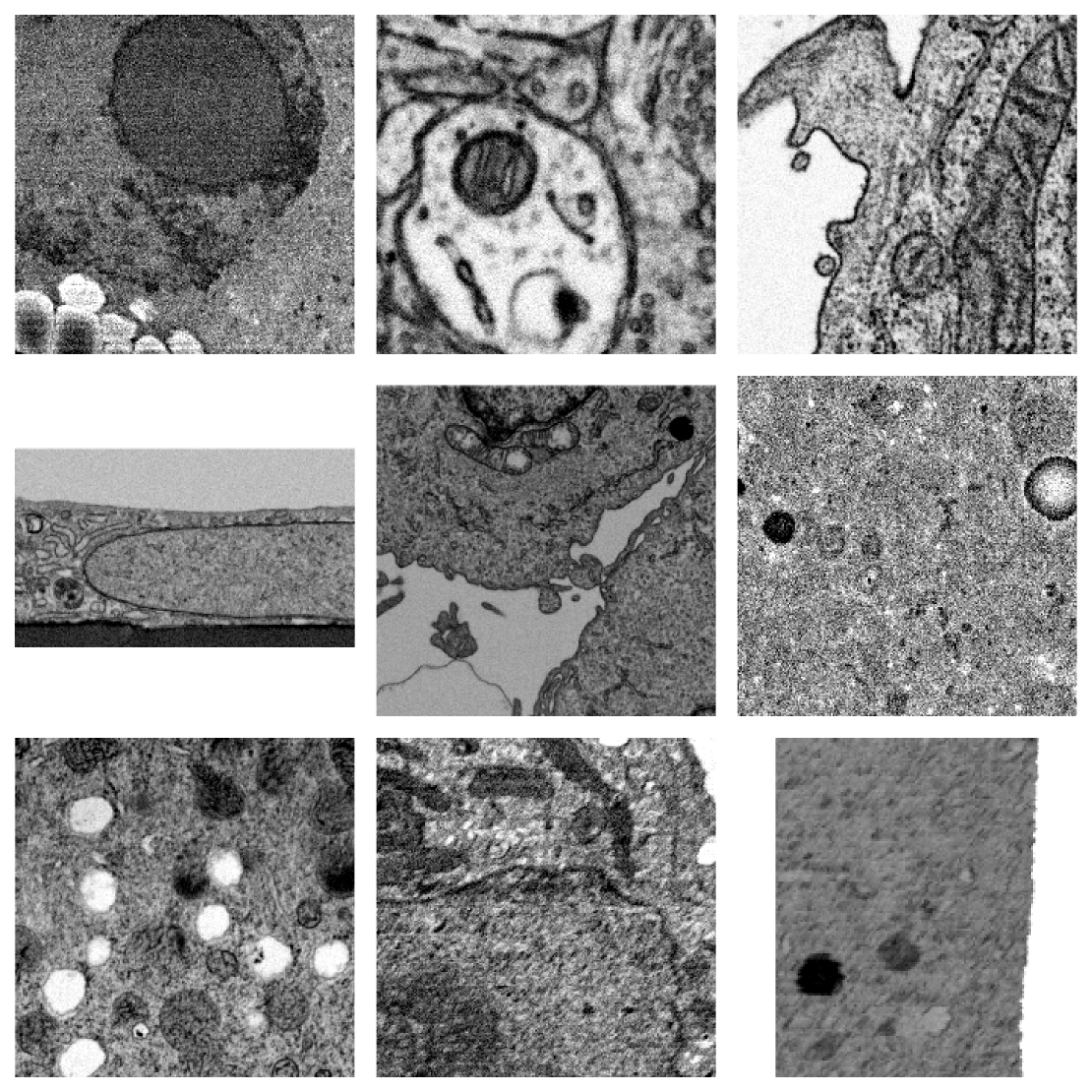}
\caption{Examples from the CEM500K dataset used for pretraining.}
\label{fig:supp_cem500k_examples}
\end{figure*}

\subsubsection{CEM500K Dataset:}

Self-supervised pretraining on real-space electron microscopy images is performed using the CEM500K dataset \cite{Conrad2021}. CEM500K is a large-scale collection of more than 500,000 unlabeled cellular electron microscopy images compiled to support representation learning from heterogeneous EM data.

Due to the large number of SSL models and experimental configurations evaluated in this work, we pretrain on a randomly sampled subset of CEM500K consisting of 10,000 images for training and 2,000 images for validation. The subset is sampled uniformly from the full dataset to preserve its structural diversity. To facilitate reproducibility, the list of selected image filenames for each split will be provided upon release. Representative examples from the dataset are shown in Figure~\ref{fig:supp_cem500k_examples}.

\subsubsection{NFFA Dataset:}

Downstream real-space classification experiments are performed on the NFFA dataset \cite{aversa2018nffa}, a scanning electron microscopy (SEM) image collection provided by the NFFA-Europe project. The dataset contains images of nanostructures and materials across ten classes. The class distribution is show in Table \ref{tab:supp_nffa_classes}:

\begin{table}[htb!]
\centering
\caption{Class distribution of the NFFA dataset used for downstream classification.}
\label{tab:supp_nffa_classes}
\small
\begin{tabular}{lc}
\toprule
Class & Number of Images \\
\midrule
Biological & 962 \\
Fibres & 150 \\
Films\_Coated\_Surface & 309 \\
MEMS\_devices\_and\_electrodes & 4583 \\
Nanowires & 3815 \\
Particles & 3905 \\
Patterned\_surface & 4752 \\
Porous\_Sponge & 174 \\
Powder & 898 \\
Tips & 1621 \\
\bottomrule
\end{tabular}
\end{table}

\begin{figure*}[htb!]
\centering
\includegraphics[width=0.95\textwidth]{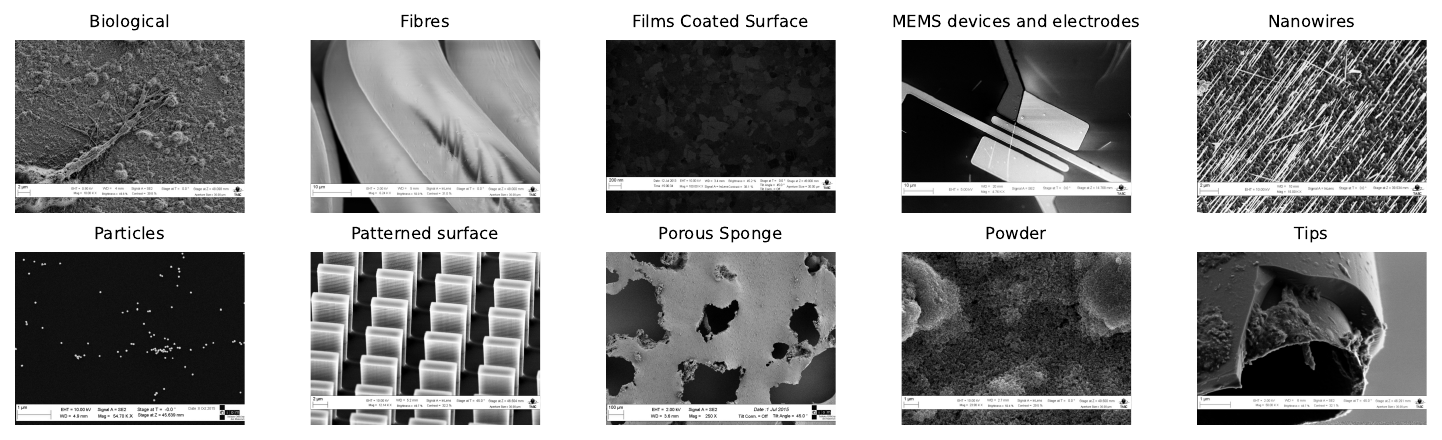}
\caption{Examples from the NFFA Dataset.}
\label{fig:supp_nffa_examples}
\end{figure*}

Following standard practice, the dataset is split into training, validation, and test sets using a ratio of 75\%, 10\%, and 15\%, respectively. Self-supervised models pretrained on CEM500K are finetuned on the labeled NFFA training set and evaluated on the test split using top-1 classification accuracy. Example images from several classes are shown in Fig.~\ref{fig:supp_nffa_examples}.

\subsubsection{4D-STEM Orientation Regression Dataset:}

Reciprocal-space experiments are conducted using simulated 4D-STEM diffraction data
provided by Scheunert et al.~\cite{scheunert_2025_cnns,scheunert2026determining}.
The dataset consists of simulated electron diffraction patterns of LiNiO$_2$
generated using the Bloch-wave algorithm implemented in the \texttt{py4DSTEM} package.

\begin{figure*}[htb!]
\centering
\includegraphics[width=0.95\textwidth]{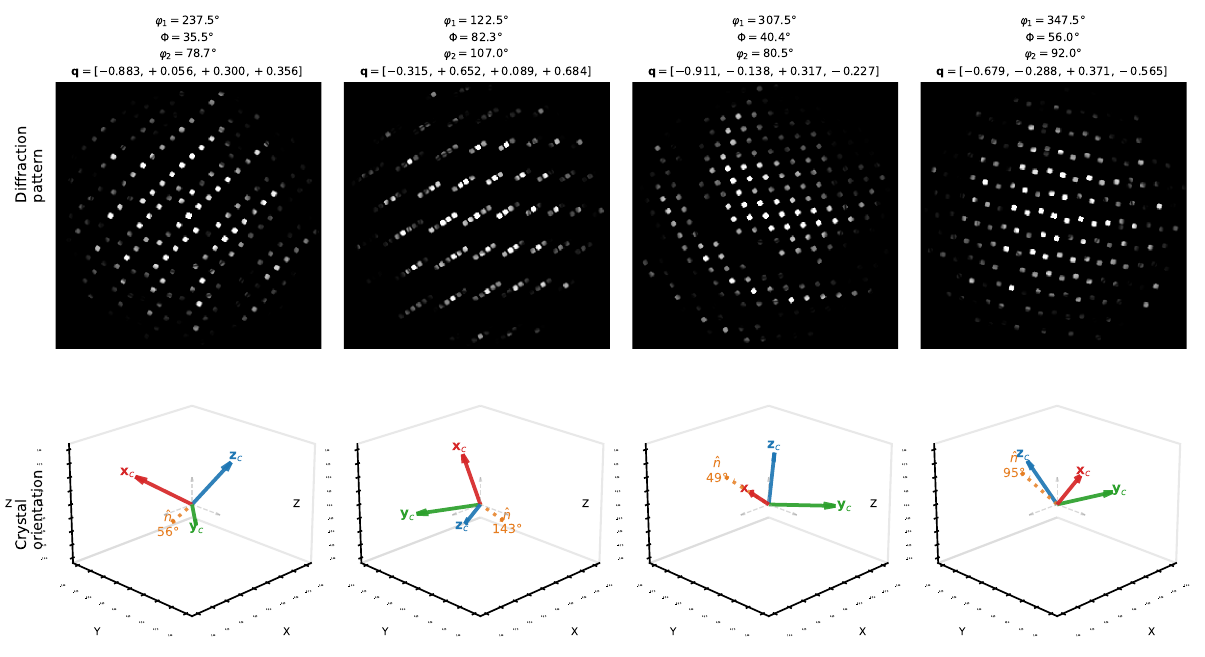}
\caption{%
  Representative examples from the 4D-STEM orientation regression dataset.
  \textbf{Top row:} Simulated electron diffraction patterns of LiNiO$_2$
  (Bloch-wave, 300\,kV, semi-convergence angle 1.5\,mrad).
  Each panel is annotated with the corresponding Bunge--Euler angles
  $(\varphi_1,\,\Phi,\,\varphi_2)$ and the unit quaternion
  $\mathbf{q}{=}[w,x,y,z]$ used as the regression target (see
  Eq.~\ref{eq:euler_to_quat}).
  \textbf{Bottom row:} 3-D visualization of the crystal orientation for each
  pattern.  The rotated crystal-frame axes
  $\mathbf{x}_c$ (red), $\mathbf{y}_c$ (green), and $\mathbf{z}_c$ (blue)
  are shown in the fixed lab frame; faint grey arrows indicate the lab-frame
  reference axes for comparison.
  The orange dashed arrow shows the equivalent rotation axis $\hat{n}$,
  and the adjacent label gives the corresponding rotation angle $\theta$,
  i.e.\ the rotation that maps the lab frame onto the crystal frame.
  The variety of diffraction patterns and orientations illustrates the
  broad coverage of the fundamental orientation zone in the dataset.
}
\label{fig:supp_stem4d_examples}
\end{figure*}

The simulated dataset contains 581,328 diffraction patterns covering the fundamental
orientation zone of the material.  The simulations assume an acceleration voltage of
300\,kV, a semi-convergence angle of 1.5\,mrad, and a maximum field of view of
8\,mrad.  The dataset also includes variations in camera length, beam shifts, and
amorphous background contributions.

For self-supervised pretraining we use a randomly sampled subset consisting of 10,000
training images and 2,000 validation images.  For downstream orientation regression we
use 232,531, 29,067, and 29,067 images for training, validation, and test,
respectively.

\paragraph{Orientation representation.}
Each diffraction pattern is annotated with a crystal orientation given as three
Bunge--Euler angles $(\varphi_1,\,\Phi,\,\varphi_2)$ following the ZXZ convention,
i.e.\ the orientation is the composition of three elemental rotations:
\begin{equation}
  R = R_z(\varphi_1)\,R_x(\Phi)\,R_z(\varphi_2).
  \label{eq:bunge_euler}
\end{equation}
Because Euler angles suffer from gimbal-lock singularities and are discontinuous over
orientation space, we convert them to unit quaternions $\mathbf{q} = [w,\,x,\,y,\,z]$
before regression.  Given half-angles
$c_1{=}\cos(\varphi_1/2)$, $s_1{=}\sin(\varphi_1/2)$,
$c{=}\cos(\Phi/2)$, $s{=}\sin(\Phi/2)$,
$c_2{=}\cos(\varphi_2/2)$, $s_2{=}\sin(\varphi_2/2)$,
the quaternion is
\begin{equation}
  \mathbf{q} =
  \frac{1}{\|\cdot\|}
  \begin{bmatrix}
    c_1 c\, c_2 - s_1 c\, s_2 \\
    c_1 s\, c_2 + s_1 s\, s_2 \\
   -c_1 s\, s_2 + s_1 s\, c_2 \\
    c_1 c\, s_2 + s_1 c\, c_2
  \end{bmatrix},
  \label{eq:euler_to_quat}
\end{equation}
where the overall sign is fixed so that $w \geq 0$ (canonical half-space).
Every unit quaternion encodes a unique axis--angle rotation
$\mathbf{q} = [\cos(\theta/2),\;\sin(\theta/2)\,\hat{n}]$,
giving rotation angle $\theta = 2\arccos(w)$ about the unit axis $\hat{n} =
(x,y,z)/\sin(\theta/2)$.  Quaternion representations avoid the discontinuities and
singularities associated with Euler-angle parameterisations and provide a smooth,
compact representation of rotations suitable for learning-based regression.
Representative diffraction patterns and their corresponding crystal-frame orientations
are shown in Fig.~\ref{fig:supp_stem4d_examples}.

\begin{table}[htb!]
\centering
\caption{Summary of datasets used in the experiments.}
\label{tab:supp_dataset_summary}
\resizebox{\columnwidth}{!}{%
\begin{tabular}{l|c|c|c|c}
\toprule
Dataset & Modality & Task & Pretraining split & Downstream split \\
\midrule
CEM500K & EM images & Pretraining & 10000/2000 & -- \\
NFFA & SEM images & Multi-class classification & -- & 15876/2116/3177 \\
4D-STEM & Diffraction patterns & Pretraining \& Orientation regression & 10000/2000 & 232531/29067/29067 \\
\bottomrule
\end{tabular}
}
\end{table}

\begin{figure}[htb!]
\centering
\begin{tikzpicture}[x=1cm,y=1cm]
\draw[rounded corners=3pt, thick] (0,0) rectangle (8.4,15.3);

\node[font=\bfseries\small, anchor=west] at (0.3,14.9) {A. Input \& augmentation regimes};
\node[smallblock, fill=inputc!12, text width=7.5cm] at (4.2,13.95)
{\textbf{CEM500K (real-space)}: resize $224\times224$ $\rightarrow$ random crop $128\times128$};
\node[smallblock, fill=inputc!12, text width=7.5cm] at (4.2,12.95)
{\textbf{4D-STEM (reciprocal-space)}: resize $256\times256$ $\rightarrow$ crop $224\times224$};
\node[smallblock, fill=neutral!12, text width=7.5cm] at (4.2,11.95)
{Across $\mathcal{T}_{\mathrm{orig}}$ and $\mathcal{T}_{\mathrm{phys}}$: backbone and training are fixed; only augmentation policy changes.};

\node[font=\bfseries\small, anchor=west] at (0.3,10.95) {B. Shared backbone};
\node[smallblock, fill=backbone!15, text width=7.5cm] at (4.2,10.0)
{\textbf{ViT-Base-EM} backbone, single-channel input, no dropout};
\node[smallblock, fill=neutral!10, text width=7.5cm] at (4.2,9.0)
{CEM500K patching: $16\times16$ (DINOv2/SimCLR/VICRegL/MAE), $8\times8$ (I-JEPA)};
\node[smallblock, fill=neutral!10, text width=7.5cm] at (4.2,8.0)
{4D-STEM patching: $16\times16$ for all methods, producing a $14\times14$ token grid};

\node[font=\bfseries\small, anchor=west] at (0.3,7.0) {C. SSL objective and method head (by row)};

\node[smallblock, fill=distill!14, text width=7.5cm] at (4.2,6.15)
{\textbf{DINOv2}: 2 global + 8 local crops $\rightarrow$ teacher--student distillation $\rightarrow$ MLP $768\to2048\to2048\to8192$, bottleneck 384};
\node[smallblock, fill=contrast!14, text width=7.5cm] at (4.2,5.15)
{\textbf{SimCLR}: two augmented views $\rightarrow$ contrastive NT-Xent $\rightarrow$ MLP $768\to2048\to128$};
\node[smallblock, fill=regular!14, text width=7.5cm] at (4.2,4.15)
{\textbf{VICRegL}: two views + local matching $\rightarrow$ variance/invariance/covariance regularization $\rightarrow$ MLP $768\to8192\to8192\to128$};
\node[smallblock, fill=recon!14, text width=7.5cm] at (4.2,3.15)
{\textbf{MAE}: single masked view $\rightarrow$ reconstruction $\rightarrow$ decoder (CEM500K: dim 512, depth 8; 4D-STEM: dim 384, depth 6)};
\node[smallblock, fill=predict!14, text width=7.5cm] at (4.2,2.15)
{\textbf{I-JEPA}: context--target blocks $\rightarrow$ predictive representation learning $\rightarrow$ predictor depth 6};
\end{tikzpicture}
\caption{Overview of pretraining architectures and objectives. All methods share the same ViT-Base-EM backbone (single-channel input, no dropout); differences arise from view generation and SSL objective-specific heads. Across $\mathcal{T}_{\mathrm{orig}}$ and $\mathcal{T}_{\mathrm{phys}}$, backbone and training settings are fixed and only the augmentation pipeline changes.}
\label{fig:supp_arch_overview}
\end{figure}
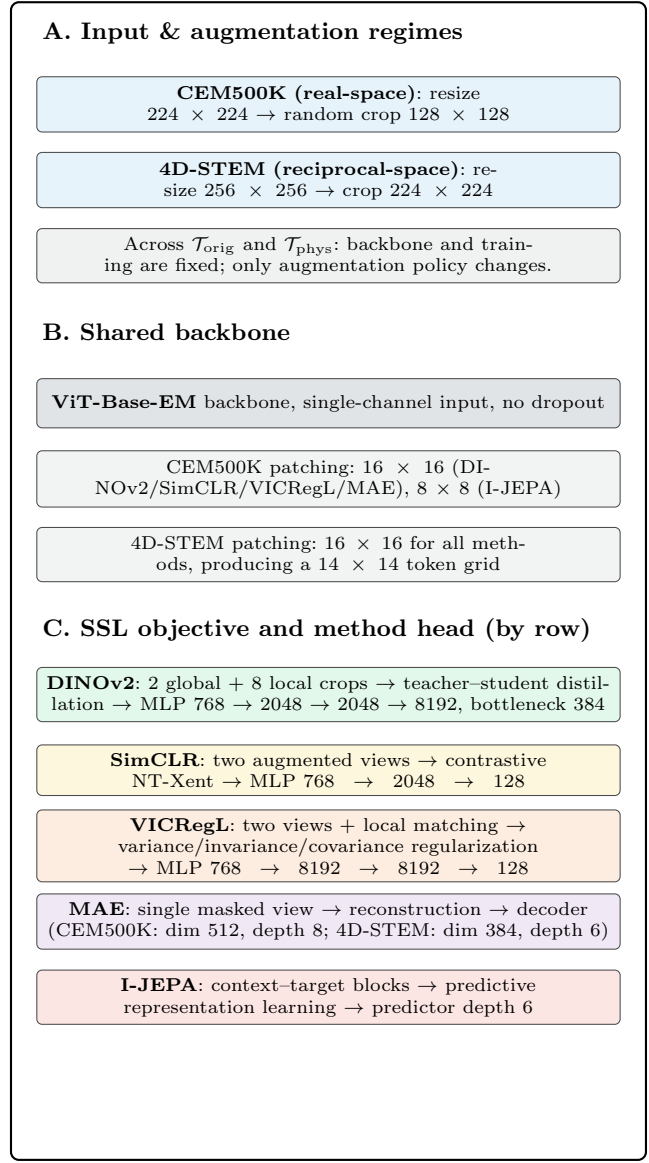

\begin{figure}[htb!]
\centering
\begin{tikzpicture}[x=1cm,y=1cm]

\draw[rounded corners=3pt, thick] (0,0) rectangle (8.4,10.0);
\node[font=\bfseries\small, anchor=west] at (0.3,9.6) {Figure S2. View / token geometry};

\draw[rounded corners=2pt, draw=black!55] (0.5,7.05) rectangle (7.9,9.2);
\draw[rounded corners=2pt, draw=black!55] (0.5,4.8) rectangle (7.9,6.95);
\draw[rounded corners=2pt, draw=black!55] (0.5,2.65) rectangle (7.9,4.7);
\draw[rounded corners=2pt, draw=black!55] (0.5,0.5) rectangle (7.9,2.55);

\node[font=\scriptsize\bfseries, anchor=west] at (0.8,8.95) {A. CEM500K crop \& patching};
\node[font=\scriptsize\bfseries, anchor=west] at (0.8,6.7) {B. 4D-STEM crop \& token grid};
\node[font=\scriptsize\bfseries, anchor=west] at (0.8,4.45) {C. DINOv2 multi-crop};
\node[font=\scriptsize\bfseries, anchor=west] at (0.8,2.3) {D. MAE + I-JEPA geometry};

\node[smallblock, fill=inputc!10, minimum width=2.0cm] at (2.0,8.2) {$224\times224$};
\node[smallblock, fill=inputc!10, minimum width=1.1cm] at (4.2,8.2) {$\rightarrow$};
\node[smallblock, fill=inputc!10, minimum width=2.0cm] at (6.4,8.2) {$128\times128$};
\node[note, anchor=west] at (0.9,7.45) {Patch options: \\ DINOv2 / SimCLR / VICRegL / MAE: $16\times16$  \\ I-JEPA: $8\times8$};

\node[smallblock, fill=inputc!10, minimum width=2.0cm] at (2.0,5.95) {$256\times256$};
\node[smallblock, fill=inputc!10, minimum width=1.1cm] at (4.2,5.95) {$\rightarrow$};
\node[smallblock, fill=inputc!10, minimum width=2.0cm] at (6.4,5.95) {$224\times224$};
\node[note, anchor=west] at (0.9,5.2) {Patch size $16\times16$ for all methods; token grid $14\times14$};

\node[smallblock, fill=distill!12, text width=6.7cm] at (4.2,3.55)
{DINOv2 view generation: 2 global crops + 8 local crops};

\node[smallblock, fill=recon!12, text width=3.1cm] at (2.6,1.45)
{MAE: single masked view};
\node[smallblock, fill=predict!12, text width=3.1cm] at (5.8,1.45)
{I-JEPA: context--target blocks};

\end{tikzpicture}
\caption{Operational geometry of crops, patches, and view generation used by different SSL formulations.}
\label{fig:supp_view_geometry}
\end{figure}
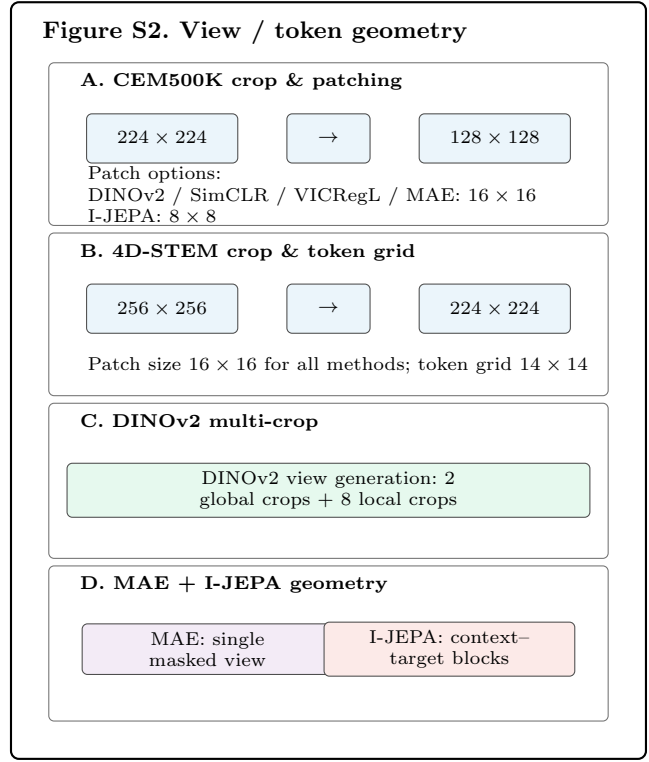

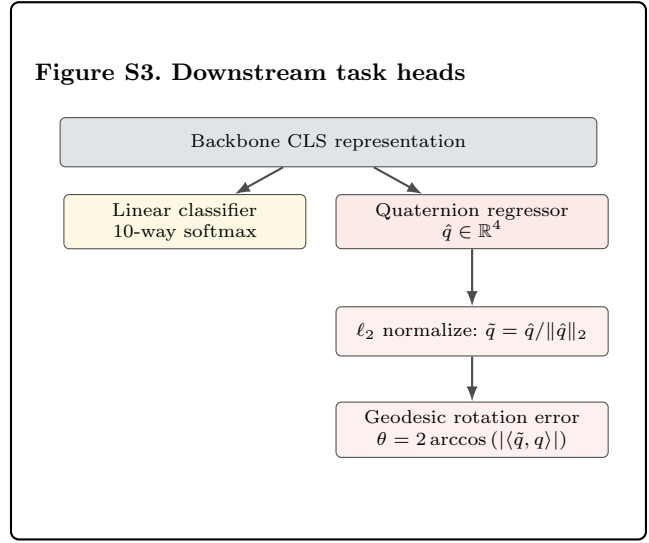
\begin{figure}[htb!]
\centering
\begin{tikzpicture}[x=1cm,y=1cm]

\draw[rounded corners=3pt, thick] (0,0) rectangle (8.4,7.1);
\node[font=\bfseries\small, anchor=west] at (0.2,6.15) {Figure S3. Downstream task heads};

\node[smallblock, fill=backbone!15, minimum width=7.1cm] (cls) at (4.2,5.25)
{Backbone CLS representation};

\node[smallblock, fill=contrast!12, minimum width=3.2cm] (nffa) at (2.3,4.2)
{Linear classifier\\10-way softmax};
\node[smallblock, fill=predict!12, minimum width=3.6cm] (quat) at (6.1,4.2)
{Quaternion regressor\\$\hat{q}\in\mathbb{R}^4$};

\draw[arrow] (cls) -- (nffa);
\draw[arrow] (cls) -- (quat);


\node[smallblock, fill=predict!8, minimum width=3.6cm] (norm) at (6.1,2.75)
{$\ell_2$ normalize: $\tilde{q}=\hat{q}/\|\hat{q}\|_2$};
\node[smallblock, fill=predict!8, minimum width=3.6cm] (geo) at (6.1,1.45)
{Geodesic rotation error\\$\theta = 2\arccos\left(|\langle \tilde{q}, q\rangle|\right)$};
\draw[arrow] (quat) -- (norm);
\draw[arrow] (norm) -- (geo);

\end{tikzpicture}
\caption{Task-specific downstream heads attached to the shared CLS representation: 10-class linear classification for NFFA, and quaternion regression with unit-sphere normalization and geodesic error for orientation estimation.}
\label{fig:supp_downstream_heads}
\end{figure}



\subsection{Model Architectures}
\label{sec:supp_models}

All methods use the same backbone family, \texttt{vit\_base\_em}, with single-channel input and no dropout during pretraining. Unless otherwise noted, models share the same backbone architecture and training configuration across the two augmentation regimes ($\mathcal{T}_{\text{orig}}$ and $\mathcal{T}_{\text{phys}}$); the only difference between these settings lies in the augmentation pipeline applied during pretraining (Fig.~\ref{fig:supp_arch_overview}).

For the real-space CEM500K experiments, input images are resized to $224\times224$ and randomly cropped to $128\times128$. The backbone patch size is $16\times16$ for DINOv2, SimCLR, VICRegL, and MAE, while I-JEPA uses a smaller patch size of $8\times8$ to provide finer spatial granularity for the predictive objective (Figs.~\ref{fig:supp_arch_overview} and~\ref{fig:supp_view_geometry}).

For the reciprocal-space 4D-STEM experiments, images are resized to $256\times256$ and cropped to $224\times224$. All methods use a patch size of $16\times16$, producing $14\times14$ token grids (Fig.~\ref{fig:supp_view_geometry}).

Method-specific differences arise primarily from the self-supervised learning objective and the associated projection or prediction heads (Fig.~\ref{fig:supp_arch_overview}). DINOv2 uses teacher--student self-distillation with a multi-crop view strategy and a high-dimensional MLP projector; SimCLR uses a two-view contrastive (NT-Xent) objective with an MLP projector; VICRegL uses variance--invariance--covariance regularization with local matching; MAE uses masked reconstruction with a lightweight decoder; and I-JEPA uses predictive representation learning with context--target blocks and a predictor head.

\subsubsection{Downstream heads}
For the NFFA classification task, a linear classification layer is attached to the backbone CLS token to predict the ten target classes. For the 4D-STEM orientation task, a regression head maps the CLS representation to a four-dimensional quaternion output (Fig.~\ref{fig:supp_downstream_heads}). Predicted quaternions are $\ell_2$-normalized to lie on the unit sphere and are evaluated using geodesic rotation error.

\subsection{Training Protocol}
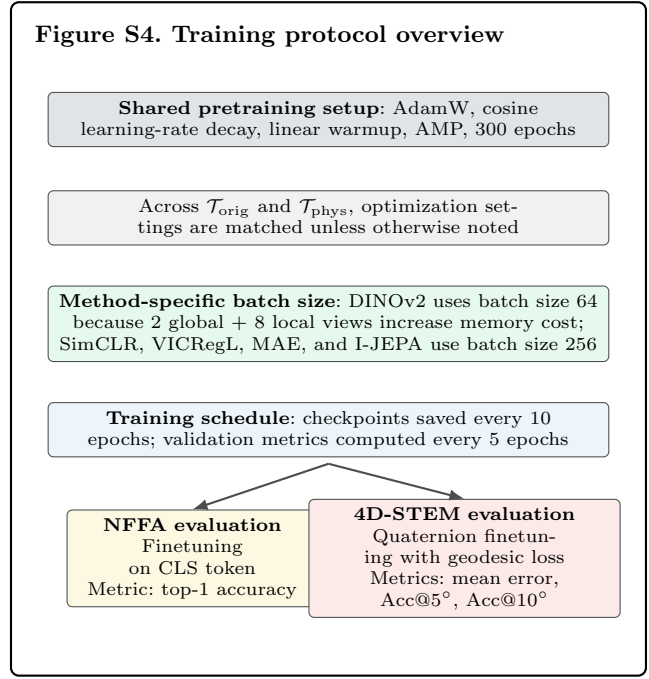
\begin{figure}[t]
\centering
\begin{tikzpicture}[x=1cm,y=1cm]

\draw[rounded corners=3pt, thick] (0,0) rectangle (8.4,8.9);
\node[font=\bfseries\small, anchor=west] at (0.2,8.45) {Figure S4. Training protocol overview};

\node[smallblock, fill=backbone!15, text width=7.2cm] at (4.2,7.35)
{\textbf{Shared pretraining setup}: AdamW, cosine learning-rate decay, linear warmup, AMP, 300 epochs};

\node[smallblock, fill=neutral!12, text width=7.2cm] at (4.2,6.05)
{Across $\mathcal{T}_{\mathrm{orig}}$ and $\mathcal{T}_{\mathrm{phys}}$, optimization settings are matched unless otherwise noted};

\node[smallblock, fill=distill!12, text width=7.2cm] at (4.2,4.65)
{\textbf{Method-specific batch size}: DINOv2 uses batch size 64 because 2 global + 8 local views increase memory cost; SimCLR, VICRegL, MAE, and I-JEPA use batch size 256};

\node[smallblock, fill=inputc!10, text width=7.2cm] at (4.2,3.25)
{\textbf{Training schedule}: checkpoints saved every 10 epochs; validation metrics computed every 5 epochs};

\node[smallblock, fill=contrast!12, text width=3.1cm] (nffa_eval) at (2.4,1.55)
{\textbf{NFFA evaluation}\\Finetuning on CLS token\\Metric: top-1 accuracy};

\node[smallblock, fill=predict!12, text width=3.9cm] (stem_eval) at (6.0,1.55)
{\textbf{4D-STEM evaluation}\\Quaternion finetuning with geodesic loss\\Metrics: mean error, Acc@5$^\circ$, Acc@10$^\circ$};

\draw[arrow] (4.2,2.8) -- (nffa_eval.north);
\draw[arrow] (4.2,2.8) -- (stem_eval.north);

\end{tikzpicture}
\caption{Overview of the shared pretraining protocol and downstream evaluation setup. Exact method-specific hyperparameters are listed in Table~\ref{tab:supp_pretrain_hparams}.}
\label{fig:supp_training_protocol}
\end{figure}

All models are pretrained using the AdamW optimizer with cosine learning-rate decay and linear warmup. Mixed-precision training (AMP) is used for all experiments. Unless otherwise specified, the same optimization settings are used for both augmentation regimes ($\mathcal{T}_{\text{orig}}$ and $\mathcal{T}_{\text{phys}}$), so that performance differences arise solely from the augmentation pipeline rather than from changes in optimization (Fig.~\ref{fig:supp_training_protocol}).

Figure~\ref{fig:supp_training_protocol} summarizes the shared training pipeline and downstream evaluation setup, while Table~\ref{tab:supp_pretrain_hparams} lists the exact pretraining hyperparameters for each method. DINOv2 uses a smaller batch size because the multi-crop strategy produces ten views per image and substantially increases memory consumption. All other methods use a batch size of 256.

\subsubsection{Training schedule}
All models are trained for 300 epochs. Model checkpoints are saved every 10 epochs, and validation metrics are computed every 5 epochs.

\subsubsection{Downstream evaluation}
For the NFFA classification task, a linear classifier is trained on top of the pretrained backbone using the CLS-token representation, and performance is reported using top-1 accuracy. For the 4D-STEM orientation task, the pretrained backbone is finetuned with a quaternion regression head. The loss is defined as the geodesic angular distance between predicted and ground-truth rotations on $SO(3)$, and performance is reported using mean geodesic error together with angular accuracy thresholds (Acc@5$^\circ$ and Acc@10$^\circ$), as summarized in Fig.~\ref{fig:supp_training_protocol}.

\begin{table}[htb!]
\centering
\caption{Pretraining hyperparameters shared across datasets.}
\label{tab:supp_pretrain_hparams}
\small
\setlength{\tabcolsep}{5pt}
\begin{tabular}{lcccccc}
\toprule
Method & Batch & LR & WD & Warmup & Min LR & Grad clip \\
\midrule
DINOv2 & 64  & $4\times10^{-4}$ & 0.04 & 20 & $10^{-6}$ & 1.0 \\
SimCLR & 256 & $8\times10^{-4}$ & $10^{-6}$ & 40 & $10^{-6}$ & 1.0 \\
VICRegL & 256 & $8\times10^{-4}$ & 0.05 & 30 & $10^{-6}$ & -- \\
MAE & 256 & $8\times10^{-4}$ & 0.05 & 30 & $10^{-6}$ & 1.0 \\
I-JEPA & 256 & $4\times10^{-4}$ & 0.05 & 25 & $10^{-6}$ & 1.0 \\
\bottomrule
\end{tabular}
\end{table}

\subsection{Augmentation Pipelines}
\label{sec:supp_augmentations}

\begin{table*}[htb!]
\centering
\caption{Augmentation pipelines used during pretraining. Parameter ranges are
shown where applicable, with the per-view application probability $p$ given for
each $\mathcal{T}_{\text{phys}}$ transform. $\mathcal{T}_{\text{orig}}$ follows
the augmentation strategies of the respective SSL method papers (natural-image
settings); $\mathcal{T}_{\text{phys}}$ replaces or supplements these with
acquisition-aware perturbations. All non-augmentation hyperparameters are
identical between regimes.}
\label{tab:supp_aug_summary}
\small
\setlength{\tabcolsep}{5pt}
\resizebox{\textwidth}{!}{%
\begin{tabular}{lcc}
\toprule
Augmentation & $\mathcal{T}_{\text{orig}}$ & $\mathcal{T}_{\text{phys}}$ \\
\midrule

Random resized crop
& scale $[0.08,1.0]$ (contrastive), $[0.2,1.0]$ (MAE)
& scale $[0.2,1.0]$ (real-space), $[0.4,1.0]$ (4D-STEM); $p{=}1.0$ \\

Horizontal / vertical flips
& enabled
& disabled for diffraction data; replaced by $90^\circ$ rotations \\

Discrete $90^\circ$ rotations
& --
& enabled; $p{=}0.5$ \\

Gaussian blur
& $\sigma \sim U(0.1,2.0)$
& $\sigma \in [0.1,1.5]$ (real-space), $[0.1,1.2]$ (4D-STEM); $p{=}0.5$ \\

Intensity scaling (gain)
& $[0.8,1.2]$
& $[0.8,1.2]$ (real-space), $[0.85,1.15]$ (4D-STEM); $p{=}0.5$ \\

Intensity bias
& --
& $[-0.1,0.1]$; $p{=}0.5$ \\

Brightness / contrast jitter
& --
& $b\in[-0.1,0.1],\,c\in[0.8,1.2]$ (real-space), $p{=}0.5$; $b\in[-0.08,0.08],\,c\in[0.9,1.1]$ (4D-STEM), $p{=}0.4$ \\

Gaussian read noise
& --
& $\sigma \sim U(0,0.08)$; $p{=}0.5$ \\

Poisson shot noise
& --
& scale $0.2$; $p{=}0.5$ \\

Solarization
& used by DINOv2 / VICRegL
& disabled \\

Scanline dropout
& --
& up to 3 lines (real-space only); $p{=}0.3$ \\

Charging streaks
& --
& up to 2 streaks, intensity $[0.05,0.15]$ (real-space only); $p{=}0.3$ \\

Reciprocal-space scaling
& --
& scale $[0.92,1.08]$ (camera length variation; 4D-STEM only); $p{=}0.5$ \\

Diffraction tilt
& --
& $\pm4^\circ$ beam tilt (4D-STEM only); $p{=}0.4$ \\

Virtual aperture mask
& --
& radius $[0.1,0.3]$ of image extent (4D-STEM only); $p{=}0.3$ \\

\bottomrule
\end{tabular}
}
\end{table*}

\begin{figure}[b]
\centering
\includegraphics[width=0.98\textwidth]{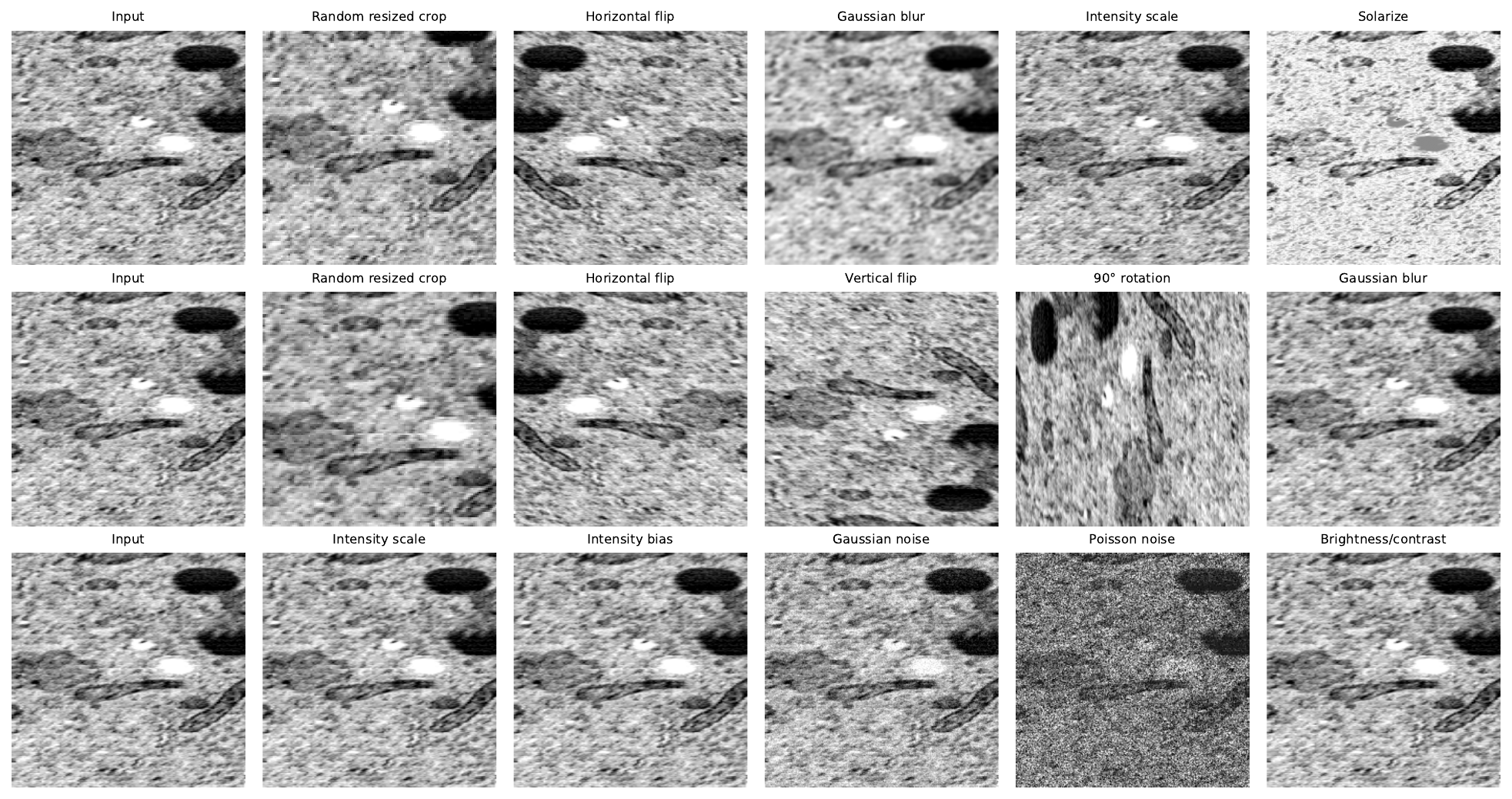}
\caption{\textbf{CEM500K augmentation effects (individual transforms).} Each row contains six panels (input + five augmentations). The first row shows representative transforms from $\mathcal{T}_{\mathrm{orig}}$, while the second and third rows show $\mathcal{T}_{\mathrm{phys}}$ transforms. This visualization highlights the shift from natural-image style perturbations to EM acquisition-aware augmentations.}
\label{fig:supp_aug_cem500k}
\end{figure}

We compare two augmentation regimes during self-supervised pretraining. The original pipeline ($\mathcal{T}_{\text{orig}}$) follows augmentation strategies from the corresponding SSL method papers, which were largely designed for natural-image pretraining. The physics-aligned pipeline ($\mathcal{T}_{\text{phys}}$) replaces or supplements these operations with perturbations motivated by electron-microscopy image formation and acquisition variability.

All augmentations operate on single-channel grayscale images that are percentile-normalized to $[0,1]$. Both regimes share common preprocessing (resize to a fixed canvas, then random crop to model input size; Section~\ref{sec:supp_models}). Consequently, the primary difference between regimes is the set of stochastic spatial and acquisition-aware perturbations applied during training. For 4D-STEM in particular, intensity and blur ranges are kept tighter than in real-space EM to preserve Bragg-peak structure.

\begin{figure*}[htb!]
\centering
\includegraphics[width=0.98\textwidth]{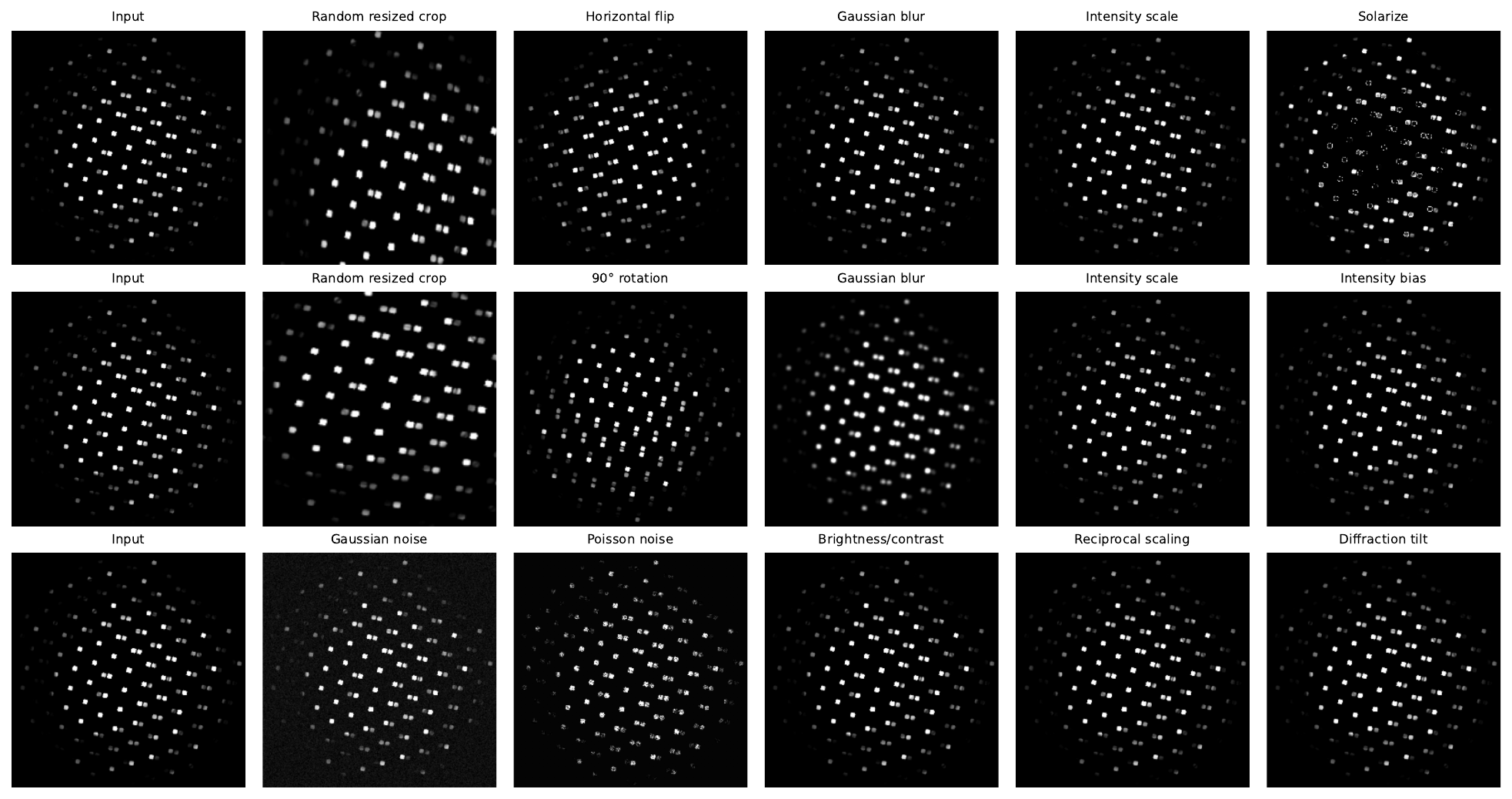}
\caption{\textbf{4D-STEM augmentation effects (individual transforms).} Layout matches Fig.~\ref{fig:supp_aug_cem500k}: input + five transforms per row, with $\mathcal{T}_{\mathrm{orig}}$ in the first row and $\mathcal{T}_{\mathrm{phys}}$ in the second/third rows. The physics-aligned regime includes diffraction-consistent perturbations (e.g., reciprocal scaling and tilt) while omitting mirror flips.}
\label{fig:supp_aug_4dstem}
\end{figure*}

Table~\ref{tab:supp_aug_summary} reports the exact parameter ranges and application probabilities used in each regime. Figures~\ref{fig:supp_aug_cem500k} and~\ref{fig:supp_aug_4dstem} provide qualitative examples on CEM500K and 4D-STEM, respectively, showing both per-augmentation effects and sampled outputs from the full pipelines.

\subsubsection{Dataset-specific augmentations}
For real-space CEM500K, $\mathcal{T}_{\text{phys}}$ includes acquisition-artifact perturbations such as scanline dropout and charging streaks, together with physically motivated intensity/noise variations. For reciprocal-space 4D-STEM, $\mathcal{T}_{\text{phys}}$ emphasizes diffraction-consistent perturbations (e.g., reciprocal-space scaling and small diffraction tilt) and disables mirror flips to avoid transformations that can alter crystallographic orientation semantics.

\paragraph{A note on $90^\circ$ rotations for 4D-STEM.}
We include discrete $90^\circ$ rotations in $\mathcal{T}_{\text{phys}}$ for both modalities, but their justification differs. For real-space cellular EM, which is rotation-equivariant, they are a valid symmetry transformation. For 4D-STEM, the simulated LiNiO$_2$ specimens are rhombohedral ($R\overline{3}m$), so a $90^\circ$ rotation is not a point-group symmetry of the sample; we therefore include it as an empirical regulariser rather than a symmetry-consistent transform. Our ablation (Section~\ref{subsec:ablation}) shows it is not harmful in this setting. This is an instance where the operator test of our procedure (Section~\ref{sec:procedure}) would exclude the transform from $\mathcal{T}_{\text{sym}}$ even though it remains empirically useful, illustrating why the procedure validates the composed pipeline via ablation rather than trusting any single transform in isolation.

\begin{figure*}[t]
\centering
\includegraphics[width=0.98\textwidth]{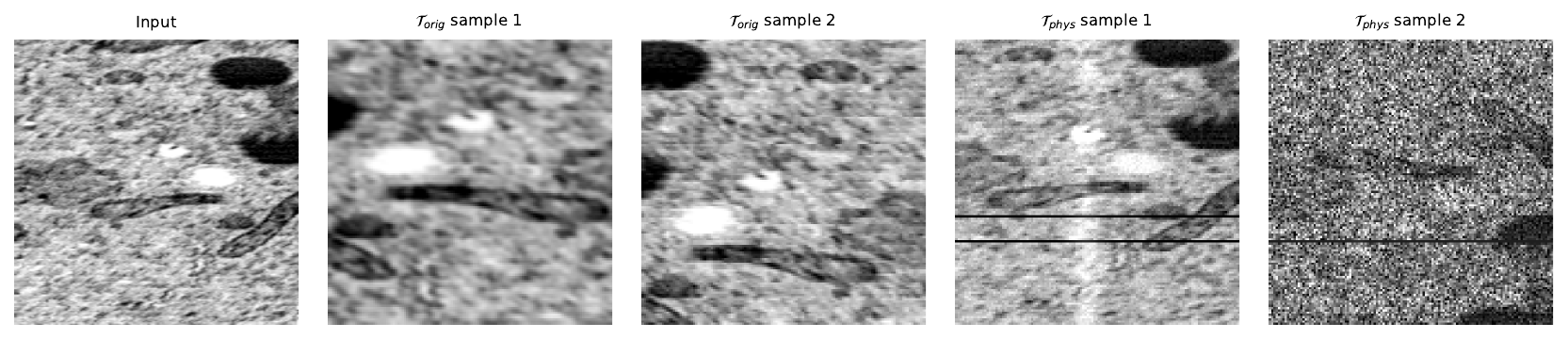}\\[0.4em]
\includegraphics[width=0.98\textwidth]{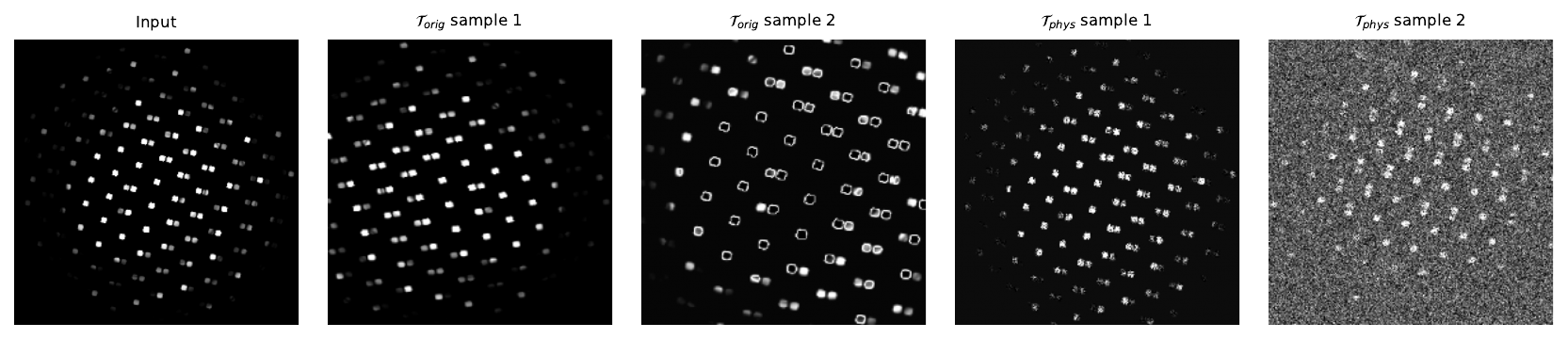}
\caption{\textbf{Sampled outputs from full augmentation pipelines.} For each dataset, we show one input plus random realizations from $\mathcal{T}_{\mathrm{orig}}$ and $\mathcal{T}_{\mathrm{phys}}$. These examples complement Figs.~\ref{fig:supp_aug_cem500k}--\ref{fig:supp_aug_4dstem} by visualizing the aggregate effect of composing multiple stochastic augmentations. Top row shows examples for CEM500K and bottom row for 4D-STEM data.}
\label{fig:supp_aug_pipeline_samples}
\end{figure*}

\section{Additional Results}
\subsection{Qualitative Results}
\label{sec:supp_qualitative}

We present qualitative examples and per-class confusion matrices for the NFFA
classification task.  All results are from a single training run, as opposed to averaged results reported in the main paper.
\subsubsection{NFFA Classification Accuracy}
\label{sec:supp_nffa_acc}

Table~\ref{tab:nffa_acc} reports test-set accuracy on NFFA for all five
backbone architectures under both augmentation regimes, as well as the
scratch baseline.  Physics-aligned pretraining ($\mathcal{T}_{\text{phys}}$)
yields higher accuracy than natural-image pretraining
($\mathcal{T}_{\text{orig}}$) for DINOv2 ({+}10.7 pp), I-JEPA ({+}1.1 pp),
and VICRegL ({+}3.3 pp), while MAE ($-$5.4 pp) and SimCLR ($-$0.5 pp) favour
$\mathcal{T}_{\text{orig}}$.  Both augmentation regimes comfortably exceed the
scratch baseline (69.4\%).

\begin{table}[h]
\centering
\caption{NFFA test accuracy (\%) for all methods and augmentation regimes,
  seed~42.  $\Delta$ = $\mathcal{T}_{\text{phys}} - \mathcal{T}_{\text{orig}}$.}
\label{tab:nffa_acc}
\setlength{\tabcolsep}{8pt}
\begin{tabular}{lccc}
\toprule
Method
  & $\mathcal{T}_{\text{orig}}$
  & $\mathcal{T}_{\text{phys}}$
  & $\Delta$ \\
\midrule
DINOv2   & 66.2 & 76.9 & \textbf{+10.7} \\
I-JEPA   & 73.2 & 74.3 &  {+1.1}  \\
MAE      & 81.3 & 75.9 &  {$-$5.4} \\
SimCLR   & 89.7 & 89.2 &  {$-$0.5} \\
VICRegL  & 73.0 & 76.3 &  {+3.3}  \\
\midrule
Scratch  & \multicolumn{2}{c}{69.4} & --- \\
\bottomrule
\end{tabular}
\end{table}

\begin{figure*}[htb!]
\centering
\includegraphics[width=0.95\textwidth]{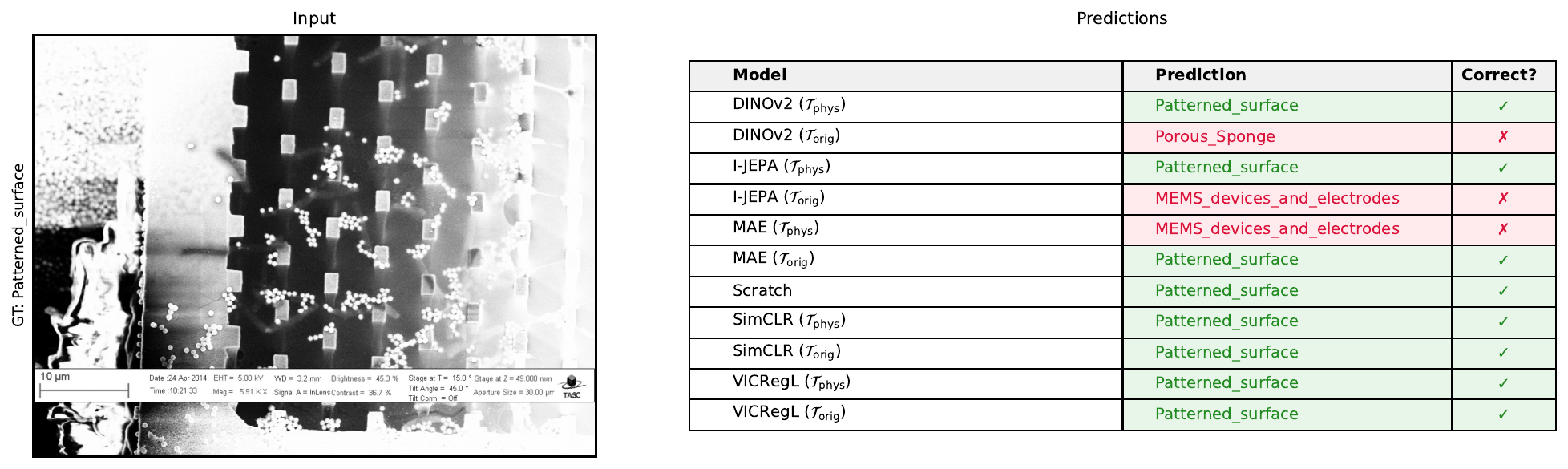}
\caption{%
  \textbf{Random NFFA qualitative examples.}
  Each row shows one test image (left) and a prediction table (right) listing
  all models.  Correct predictions are highlighted in green; incorrect in red.
}
\label{fig:qual_nffa_random}
\end{figure*}

\subsubsection{Qualitative Examples}
\label{sec:supp_nffa_qual}

\begin{figure*}[htb!]
\centering
\includegraphics[width=0.95\textwidth]{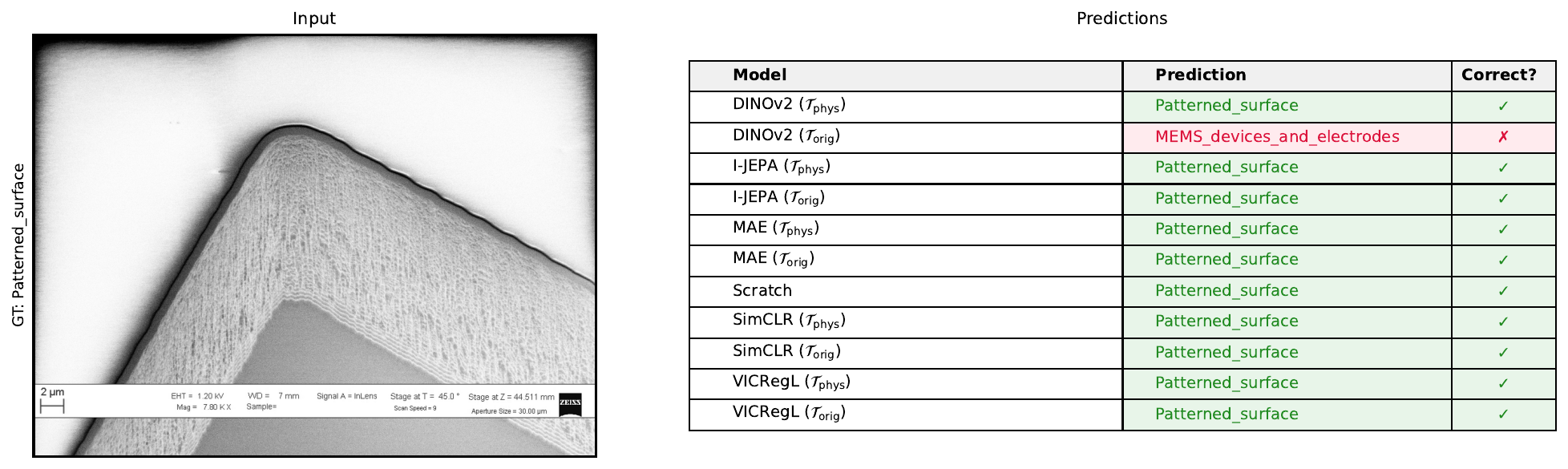}
\caption{%
  \textbf{Physics-aligned advantage examples (NFFA).}
  All $\mathcal{T}_{\text{phys}}$ models predict correctly; at least one
  $\mathcal{T}_{\text{orig}}$ model fails.  These examples predominantly
  show structurally distinctive EM morphologies (nanowires, porous networks,
  surface patterns) where domain-specific augmentation priors provide a
  recognisable benefit.
}
\label{fig:qual_nffa_domain_wins}\end{figure*}

\begin{figure*}[htb!]
\centering
\includegraphics[width=0.95\textwidth]{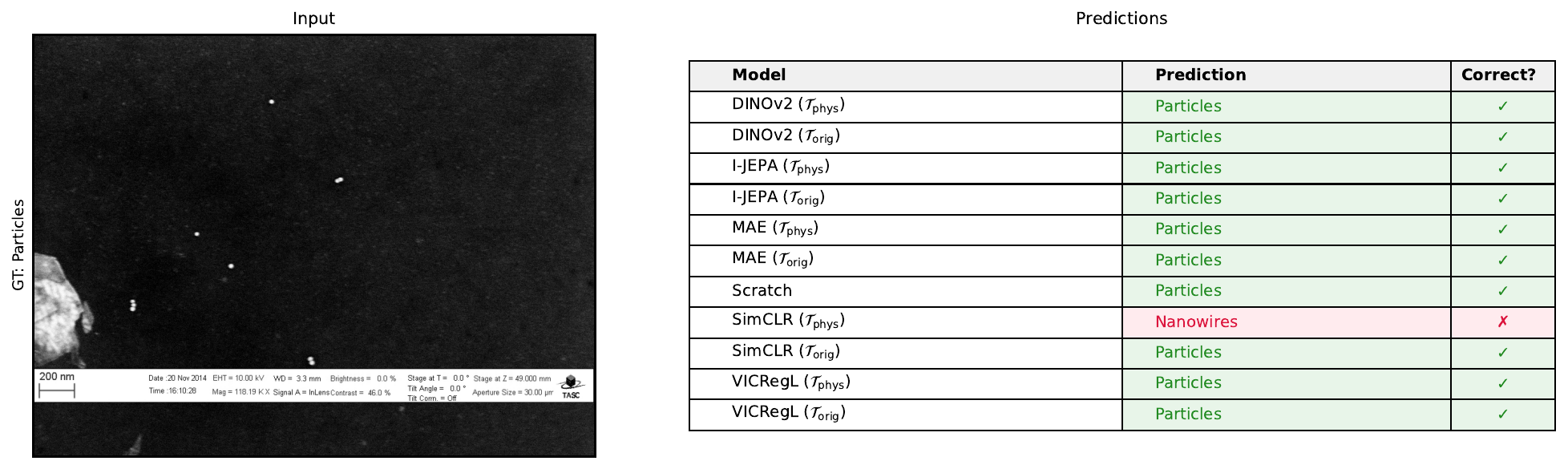}
\caption{%
  \textbf{Natural-image augmentation advantage examples (NFFA).}
  All $\mathcal{T}_{\text{orig}}$ models predict correctly; at least one
  $\mathcal{T}_{\text{phys}}$ model fails.  These represent cases where
  natural-image diversity in pretraining is more beneficial than domain
  specialisation.
}
\label{fig:qual_nffa_orig_wins}\end{figure*}

Figures~\ref{fig:qual_nffa_random}--\ref{fig:qual_nffa_orig_wins} illustrate
three complementary selection criteria.
Figure~\ref{fig:qual_nffa_random} shows randomly sampled test images with
predictions from all models.
Figure~\ref{fig:qual_nffa_domain_wins} highlights cases where
\emph{all} $\mathcal{T}_{\text{phys}}$ models are correct yet at least one
$\mathcal{T}_{\text{orig}}$ model fails; these tend to be structurally
informative images (nanostructures, patterned surfaces) where
physics-domain priors are most beneficial.
Figure~\ref{fig:qual_nffa_orig_wins} shows the converse: all
$\mathcal{T}_{\text{orig}}$ models correct, at least one
$\mathcal{T}_{\text{phys}}$ model wrong.

Qualitative inspection reveals broad variation in per-image difficulty and
no single model dominates visually.  We therefore complement these examples
with class-level confusion matrices, which
give a more complete picture of systematic error patterns.

\subsubsection{Confusion Matrices}
\label{sec:supp_nffa_cm}

\begin{figure*}[htb!]
     \centering
     \begin{subfigure}[b]{0.40\textwidth}
         \centering
         \includegraphics[width=\textwidth]{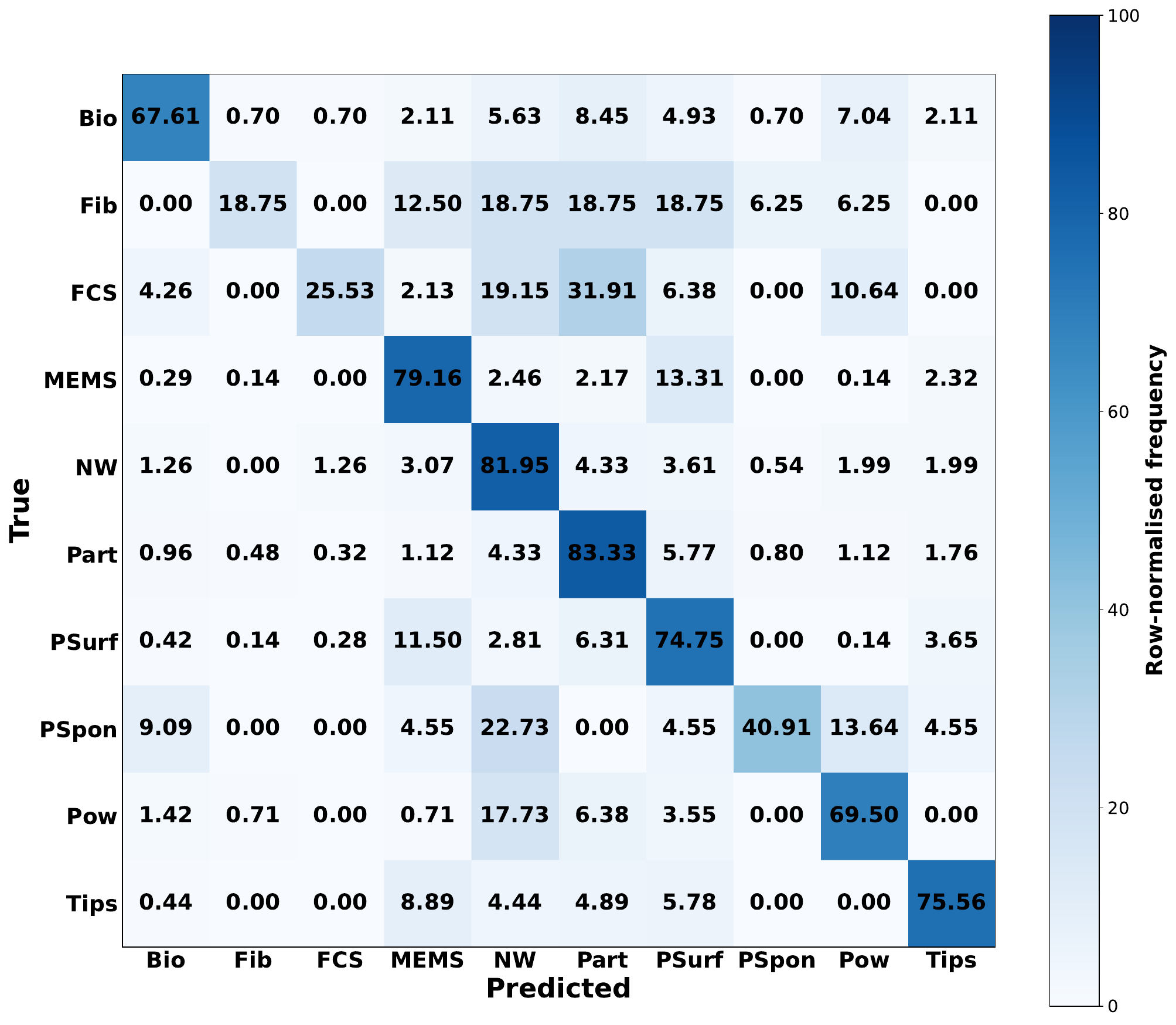}
         \caption{DINOv2 ($\mathcal{T}_{\text{phys}}$)}
         \label{fig:cm_dinov2_domain}
     \end{subfigure}
     \hfill
     \begin{subfigure}[b]{0.40\textwidth}
         \centering
         \includegraphics[width=\textwidth]{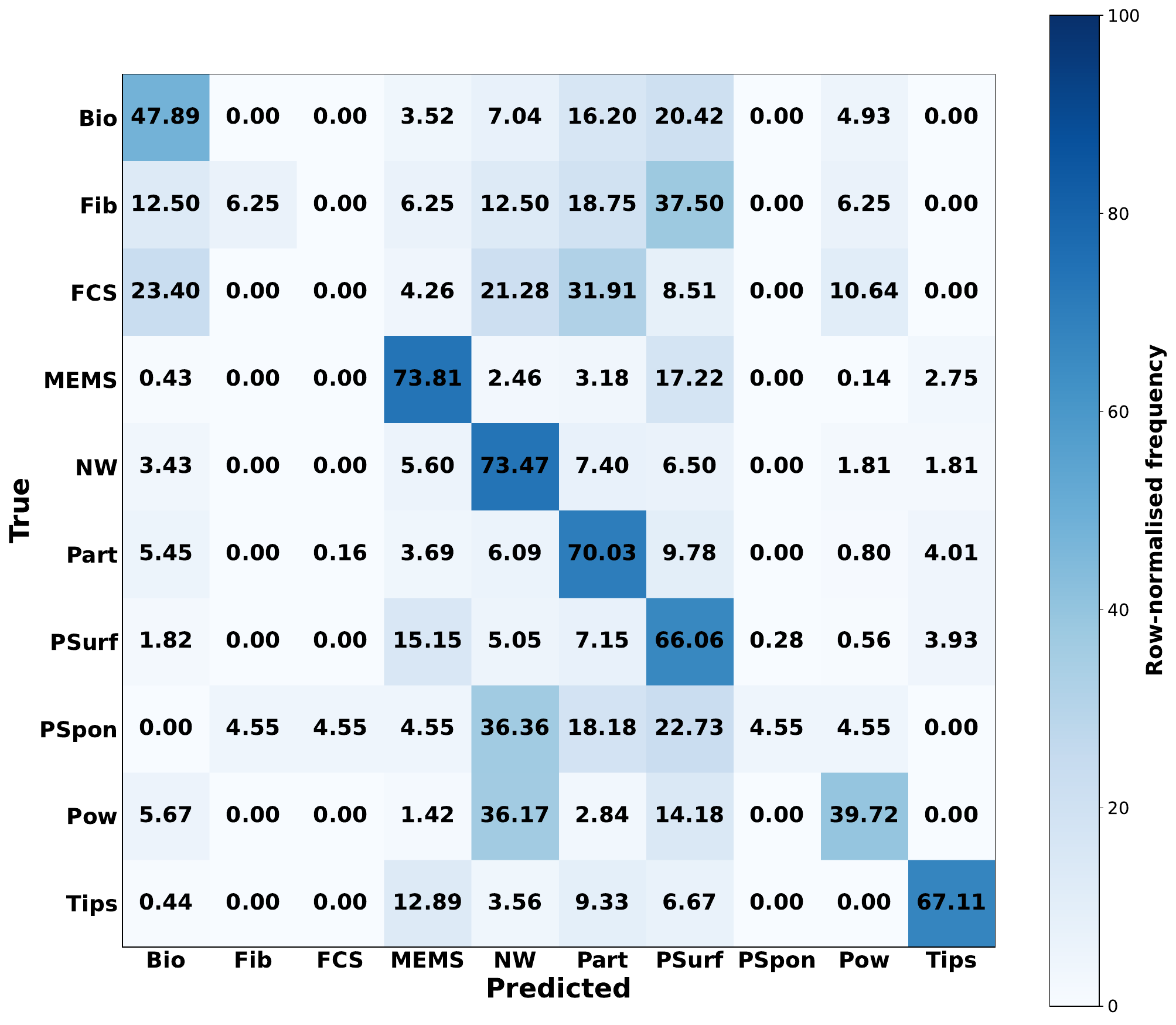}
         \caption{DINOv2 ($\mathcal{T}_{\text{orig}}$)}
         \label{fig:cm_dinov2_original}
     \end{subfigure}
        \caption{Row-normalized confusion matrices (\%) for DINOv2 on NFFA (seed~42).}
        \label{fig:cm_dinov2}
\end{figure*}

\begin{figure*}[htb!]
     \centering
     \begin{subfigure}[b]{0.40\textwidth}
         \centering
         \includegraphics[width=\textwidth]{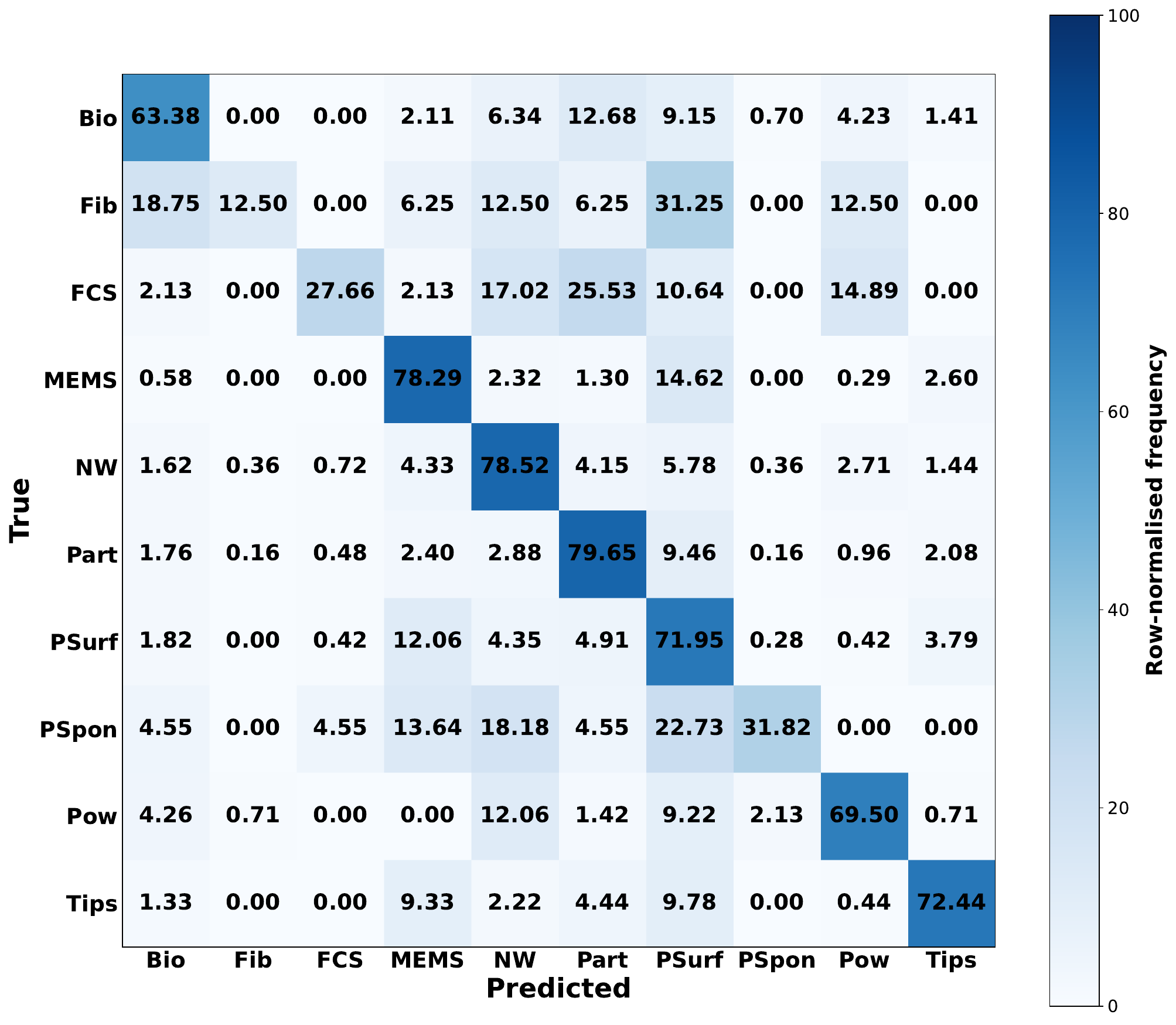}
         \caption{I-JEPA ($\mathcal{T}_{\text{phys}}$)}
         \label{fig:cm_ijepa_domain}
     \end{subfigure}
     \hfill
     \begin{subfigure}[b]{0.40\textwidth}
         \centering
         \includegraphics[width=\textwidth]{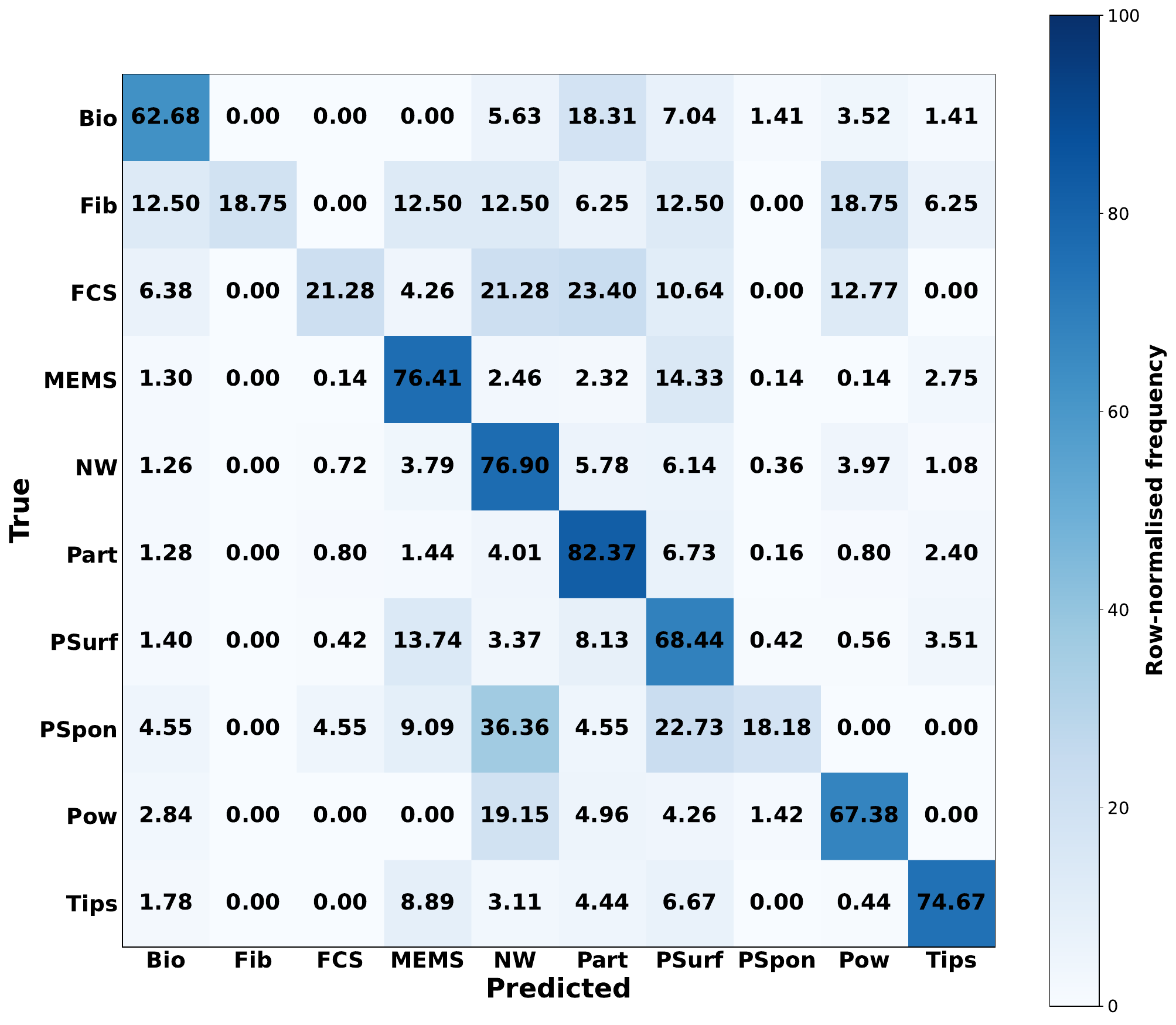}
         \caption{I-JEPA ($\mathcal{T}_{\text{orig}}$)}
         \label{fig:cm_ijepa_original}
     \end{subfigure}
        \caption{Row-normalized confusion matrices (\%) for I-JEPA on NFFA (seed~42).}
        \label{fig:cm_ijepa}
\end{figure*}

\begin{figure*}[htb!]
     \centering
     \begin{subfigure}[b]{0.40\textwidth}
         \centering
         \includegraphics[width=\textwidth]{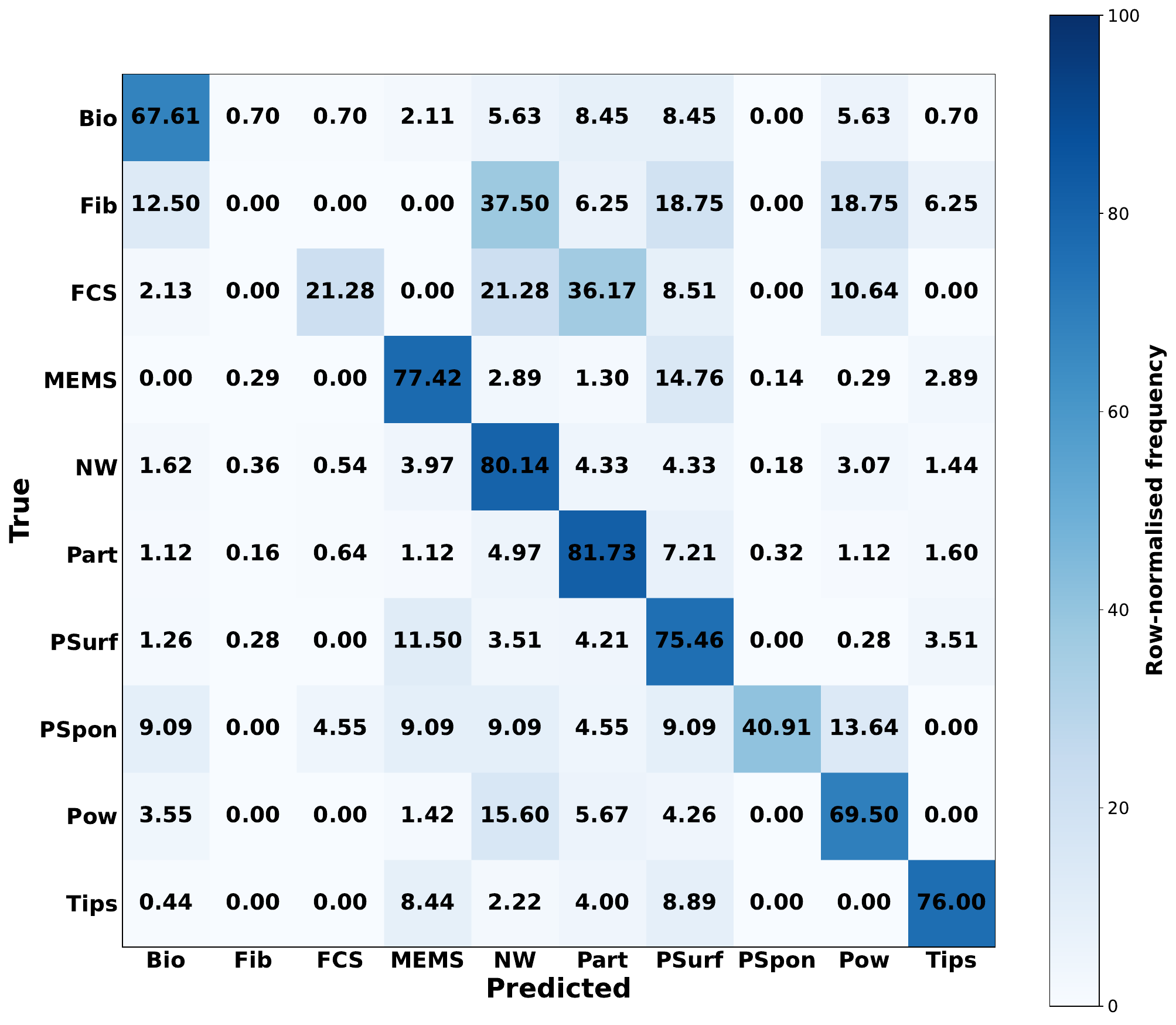}
         \caption{MAE ($\mathcal{T}_{\text{phys}}$)}
         \label{fig:cm_mae_domain}
     \end{subfigure}
     \hfill
     \begin{subfigure}[b]{0.40\textwidth}
         \centering
         \includegraphics[width=\textwidth]{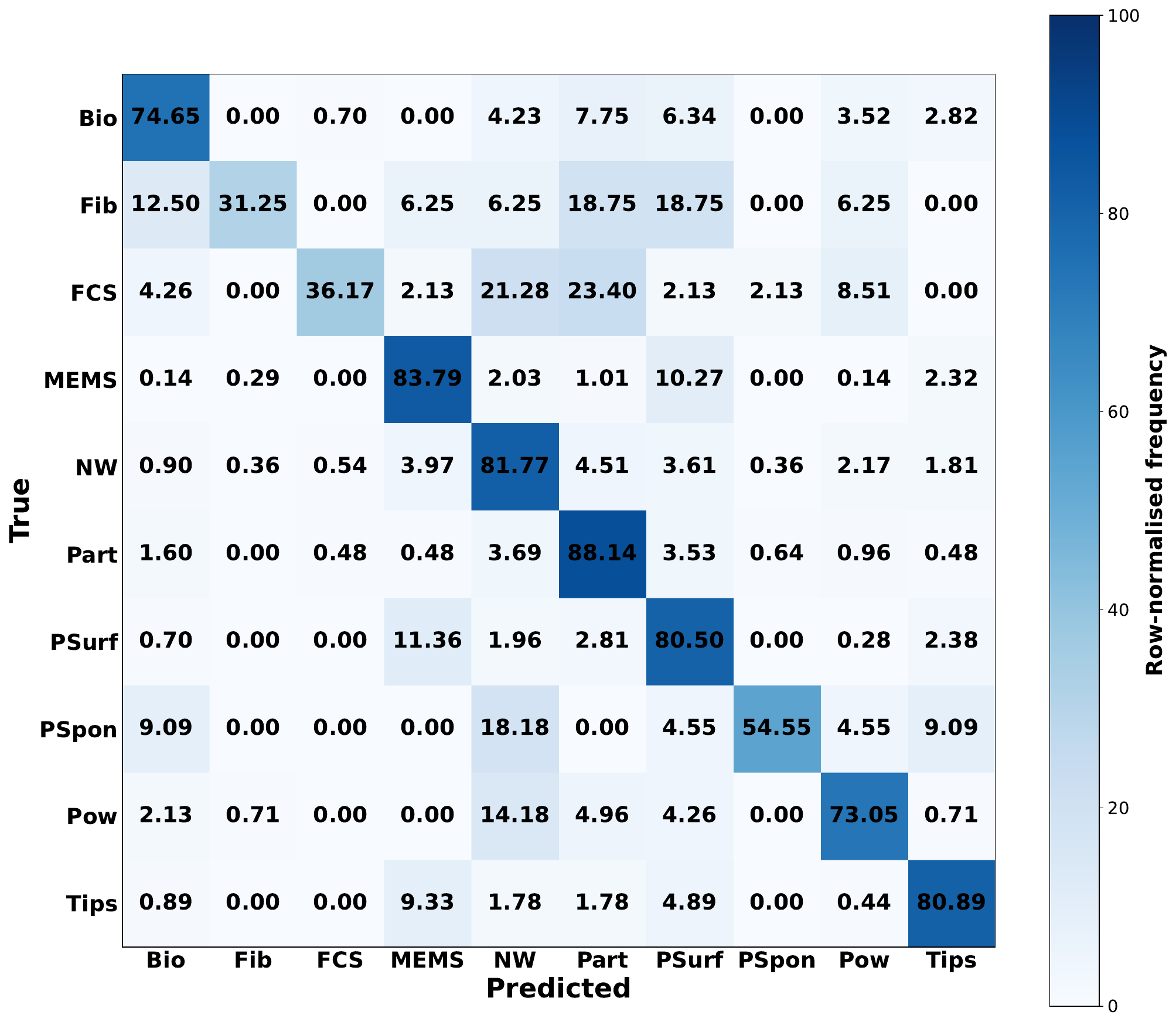}
         \caption{MAE ($\mathcal{T}_{\text{orig}}$)}
         \label{fig:cm_mae_original}
     \end{subfigure}
        \caption{Row-normalized confusion matrices (\%) for MAE on NFFA (seed~42).}
        \label{fig:cm_mae}
\end{figure*}

\begin{figure*}[htb!]
     \centering
     \begin{subfigure}[b]{0.40\textwidth}
         \centering
         \includegraphics[width=\textwidth]{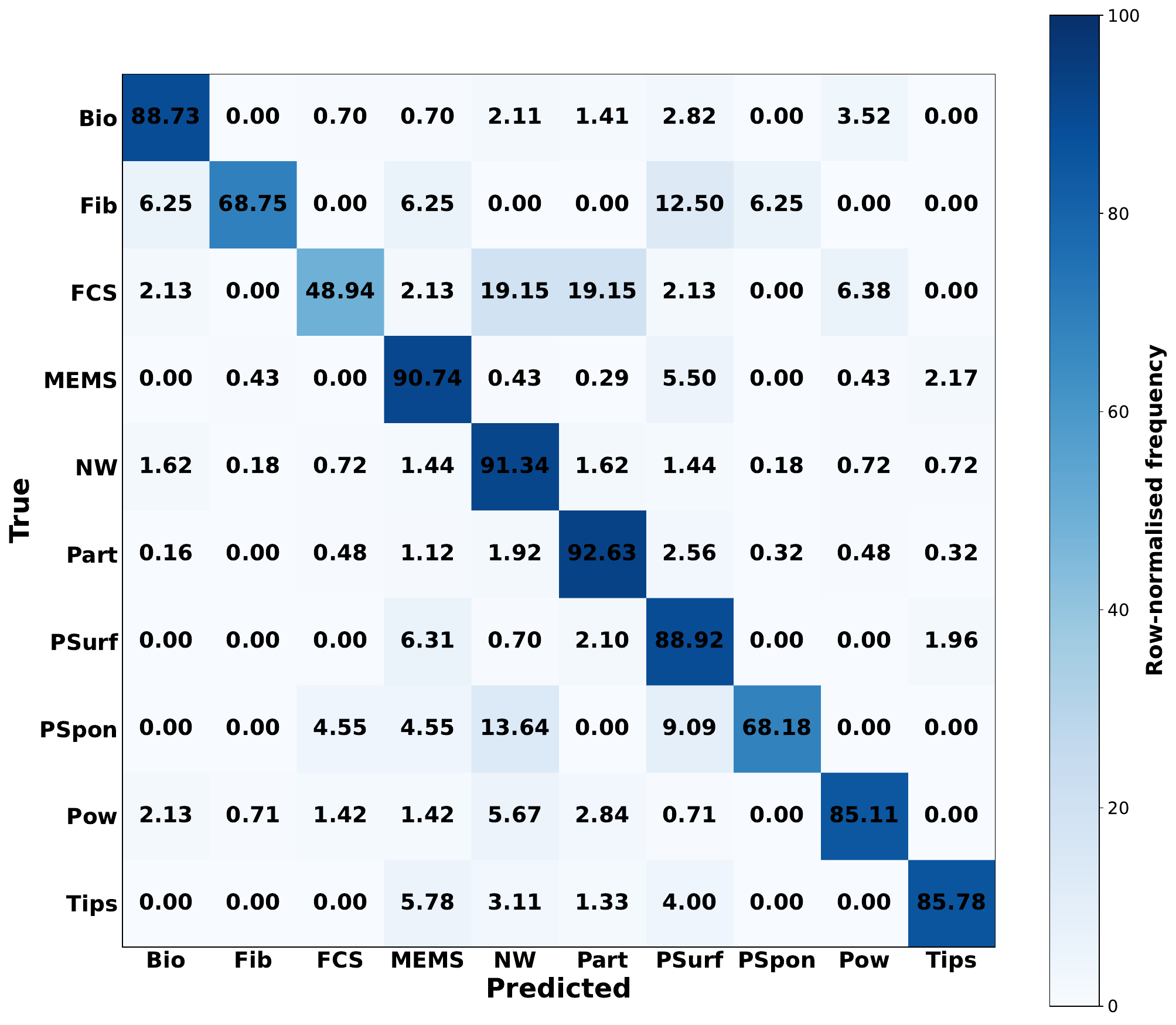}
         \caption{SimCLR ($\mathcal{T}_{\text{phys}}$)}
         \label{fig:cm_simclr_domain}
     \end{subfigure}
     \hfill
     \begin{subfigure}[b]{0.40\textwidth}
         \centering
         \includegraphics[width=\textwidth]{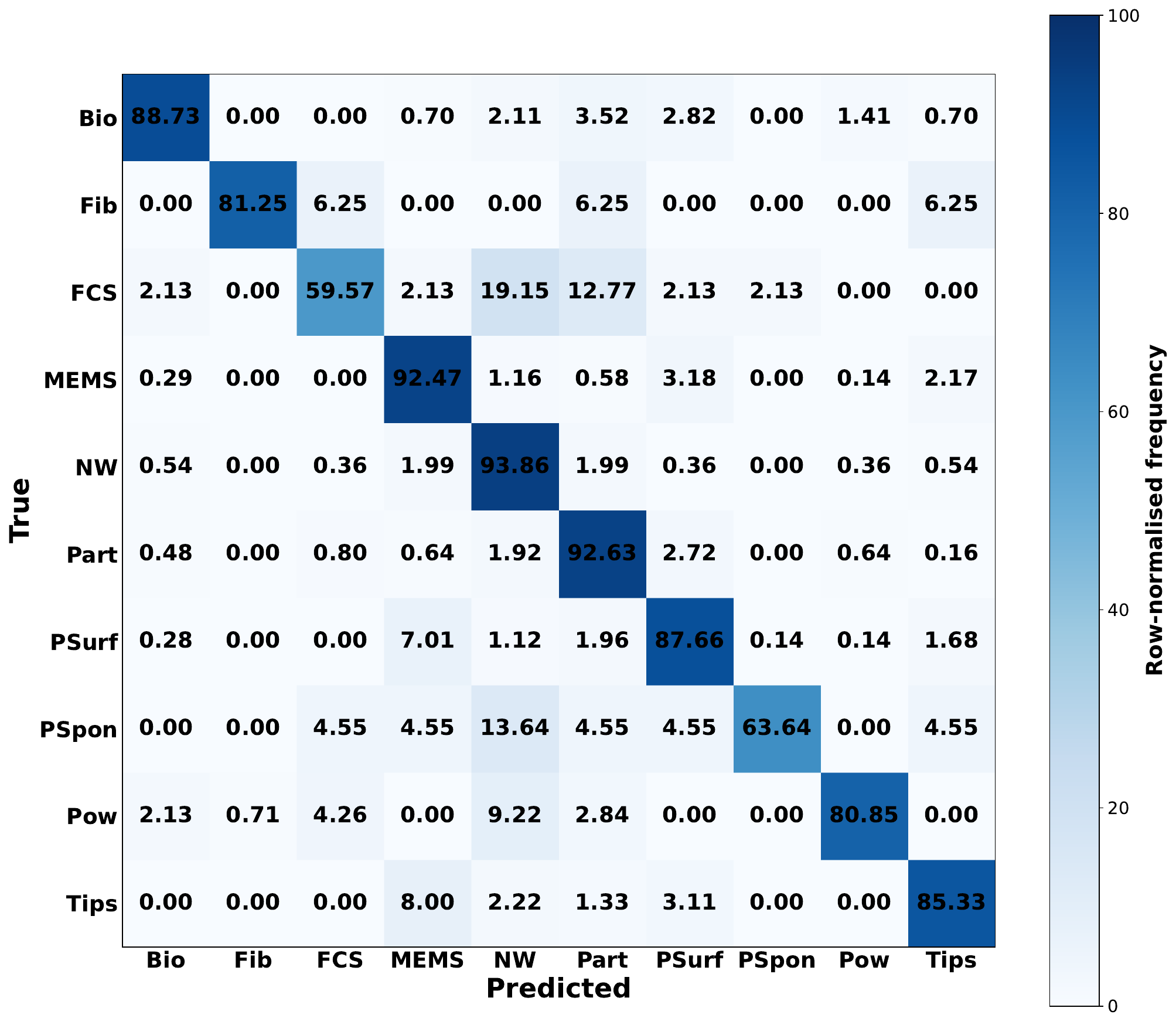}
         \caption{SimCLR ($\mathcal{T}_{\text{orig}}$)}
         \label{fig:cm_simclr_original}
     \end{subfigure}
        \caption{Row-normalized confusion matrices (\%) for SimCLR on NFFA (seed~42).}
        \label{fig:cm_simclr}
\end{figure*}

\begin{figure*}[htb!]
     \centering
     \begin{subfigure}[b]{0.40\textwidth}
         \centering
         \includegraphics[width=\textwidth]{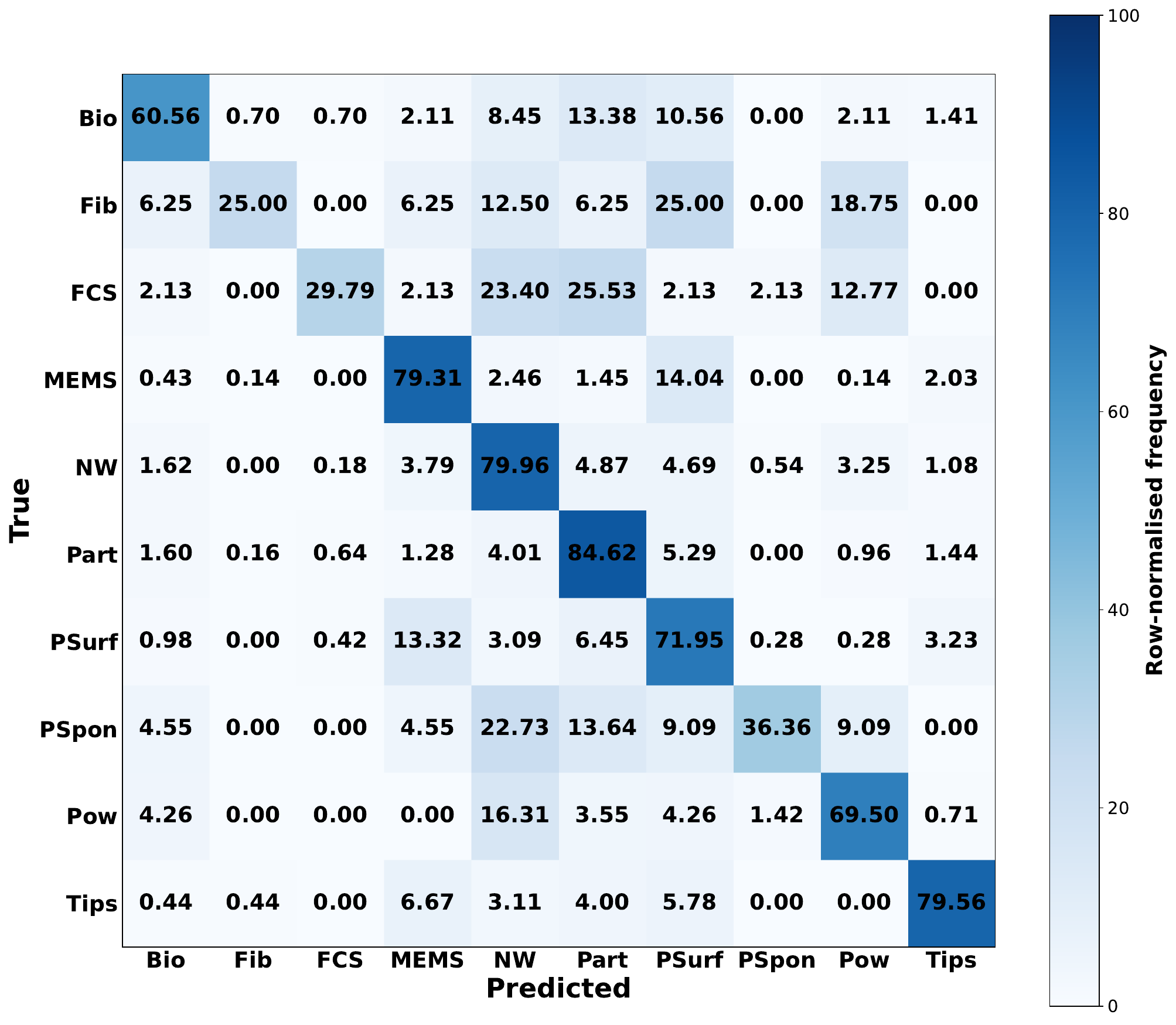}
         \caption{VICRegL ($\mathcal{T}_{\text{phys}}$)}
         \label{fig:cm_vicregl_domain}
     \end{subfigure}
     \hfill
     \begin{subfigure}[b]{0.40\textwidth}
         \centering
         \includegraphics[width=\textwidth]{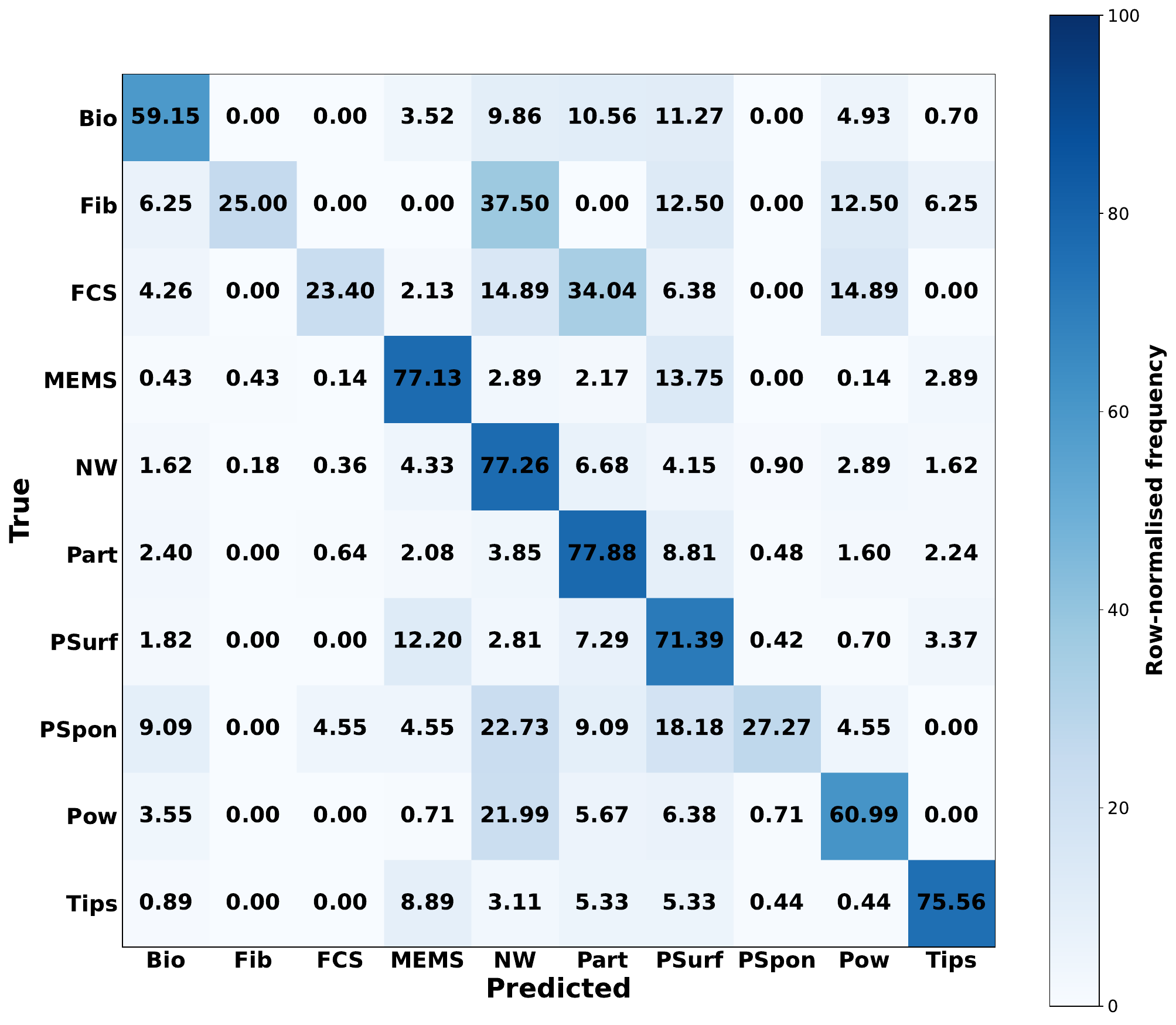}
         \caption{VICRegL ($\mathcal{T}_{\text{orig}}$)}
         \label{fig:cm_vicregl_original}
     \end{subfigure}
        \caption{Row-normalized confusion matrices (\%) for VICRegL on NFFA (seed~42).}
        \label{fig:cm_vicregl}
\end{figure*}

\begin{figure*}[htb!]
\centering
\includegraphics[width=0.40\textwidth]{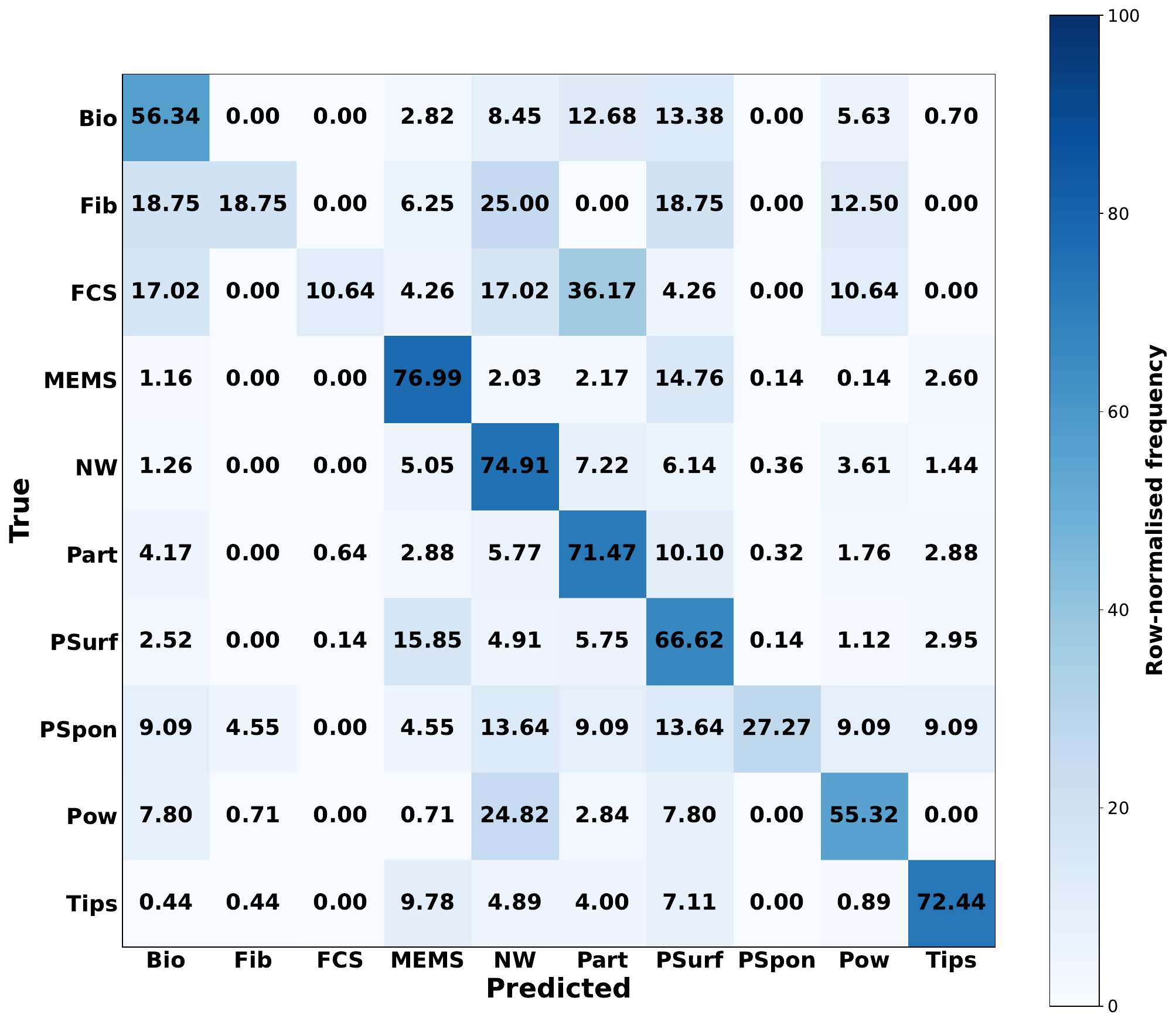}
\caption{Row-normalized confusion matrix (\%) for Scratch on NFFA (seed~42).}
\label{fig:cm_scratch_table}
\end{figure*}

We report confusion matrices as figure heatmaps. Figures~\ref{fig:cm_dinov2}--\ref{fig:cm_scratch_table}
provide row-normalized confusion values (\%) for all 11 models on the NFFA
test set, with class abbreviation mapping given in Table~\ref{tab:cm_class_key}.
Misclassification is most frequent between visually similar categories
(\emph{Porous\_Sponge} / \emph{Nanowires}, \emph{Films} / \emph{Coated Surface}),
and this pattern is consistent across methods. Higher-accuracy models show
stronger diagonal concentration, while lower-accuracy models spread errors more
broadly across classes.

\begin{table*}[t]
\centering
\footnotesize
\setlength{\tabcolsep}{6pt}
\begin{tabular}{ll}
\toprule
Abbr. & Class name \\
\midrule
Bio & Biological \\
Fib & Fibres \\
FCS & Films\_Coated\_Surface \\
MEMS & MEMS\_devices\_and\_electrodes \\
NW & Nanowires \\
Part & Particles \\
PSurf & Patterned\_surface \\
PSpon & Porous\_Sponge \\
Pow & Powder \\
Tips & Tips \\
\bottomrule
\end{tabular}
\caption{Class abbreviation key used in Figs.~\ref{fig:cm_dinov2}--\ref{fig:cm_scratch_table}.}
\label{tab:cm_class_key}
\end{table*}

\clearpage

\subsubsection{4D-STEM Orientation Prediction}
\label{sec:supp_4dstem_qual}

For 4D-STEM quaternion regression, we report mean geodesic error together with
Acc@5$^\circ$ and Acc@10$^\circ$ (Table~\ref{tab:4dstem_reg}).
Physics-aligned pretraining improves orientation accuracy for most backbones,
with especially large gains for DINOv2 and I-JEPA relative to
$\mathcal{T}_{\text{orig}}$. MAE performs strongly in both regimes, achieving
the lowest mean geodesic error overall.

\begin{table}[h]
\centering
\caption{4D-STEM quaternion regression results (seed~42). Mean geodesic error is in degrees; lower is better.}
\label{tab:4dstem_reg}
\setlength{\tabcolsep}{6pt}
\begin{tabular}{lccc}
\toprule
Method & Mean geodesic$\downarrow$ & Acc@5$^\circ\uparrow$ & Acc@10$^\circ\uparrow$ \\
\midrule
DINOv2 ($\mathcal{T}_{\text{phys}}$)   & 4.89  & 92.79 & 94.41 \\
DINOv2 ($\mathcal{T}_{\text{orig}}$)   & 10.68 & 74.21 & 83.58 \\
MAE ($\mathcal{T}_{\text{phys}}$)      & \textbf{1.40} & \textbf{98.85} & \textbf{99.03} \\
MAE ($\mathcal{T}_{\text{orig}}$)      & 1.60  & 98.60 & 98.86 \\
I-JEPA ($\mathcal{T}_{\text{phys}}$)   & 10.18 & 78.39 & 84.49 \\
I-JEPA ($\mathcal{T}_{\text{orig}}$)   & 12.04 & 66.82 & 79.42 \\
SimCLR ($\mathcal{T}_{\text{phys}}$)   & 10.05 & 80.28 & 85.24 \\
SimCLR ($\mathcal{T}_{\text{orig}}$)   & 10.85 & 80.10 & 83.96 \\
VICRegL ($\mathcal{T}_{\text{phys}}$)  & 8.31  & 83.94 & 88.59 \\
VICRegL ($\mathcal{T}_{\text{orig}}$)  & 8.19  & 83.30 & 88.74 \\
\midrule
Scratch                              & 11.43 & 72.36 & 82.32 \\
\bottomrule
\end{tabular}
\end{table}

\begin{figure*}[htb!]
\centering
\begin{subfigure}[t]{0.99\textwidth}
  \includegraphics[width=\linewidth]{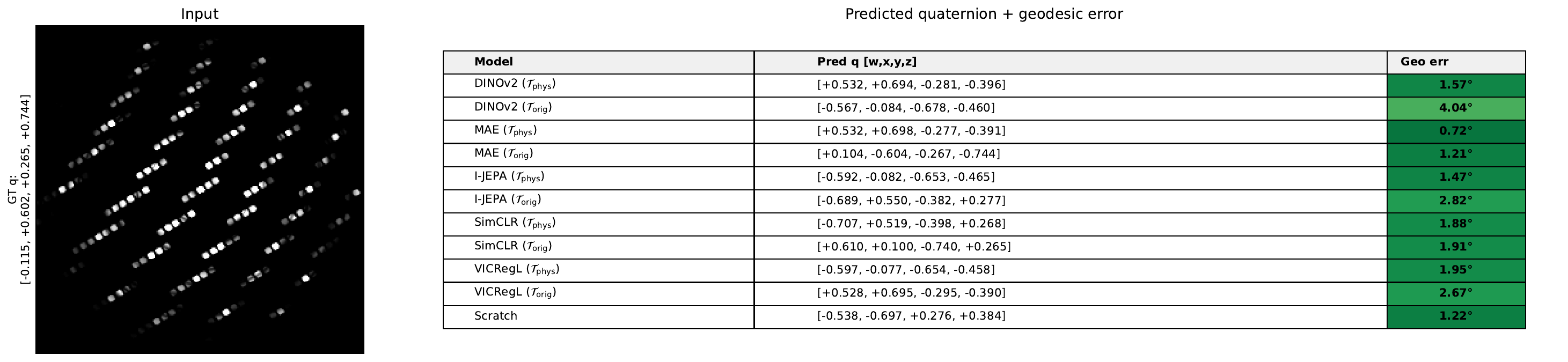}
\end{subfigure}
\vspace{0.4em}

\begin{subfigure}[t]{0.99\textwidth}
  \includegraphics[width=\linewidth]{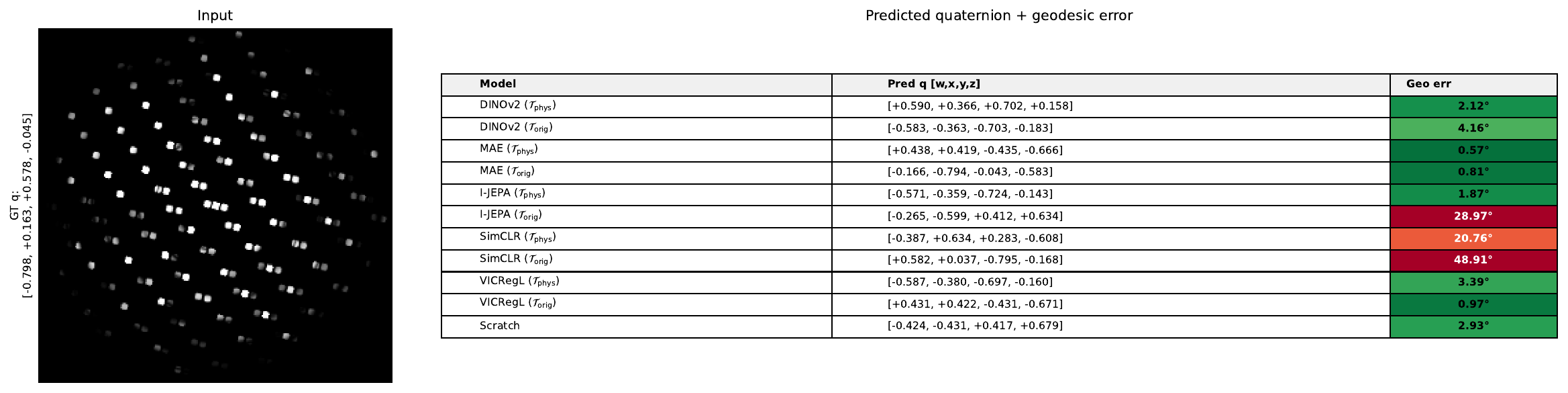}
\end{subfigure}

\vspace{0.4em}

\begin{subfigure}[t]{0.99\textwidth}
  \includegraphics[width=\linewidth]{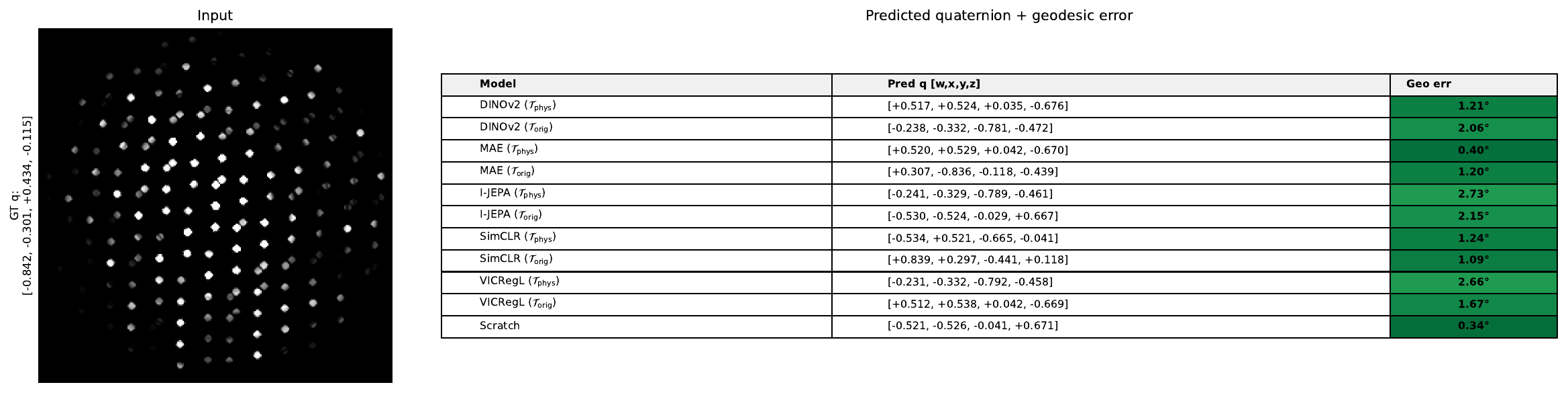}
\end{subfigure}

\vspace{0.4em}

\begin{subfigure}[t]{0.99\textwidth}
  \includegraphics[width=\linewidth]{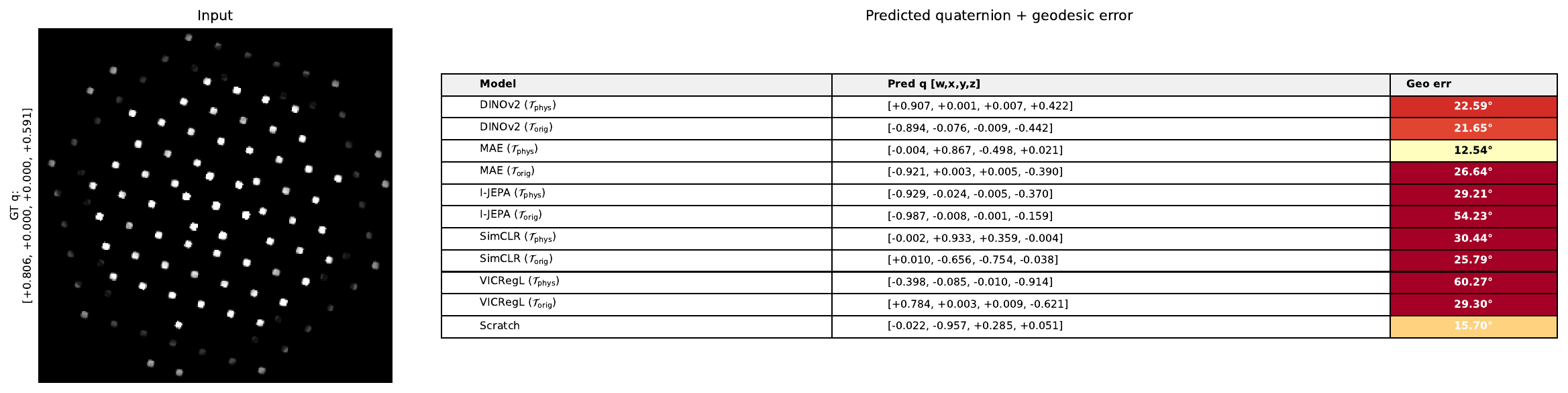}
\end{subfigure}
\vspace{0.4em}

\begin{subfigure}[t]{0.99\textwidth}
  \includegraphics[width=\linewidth]{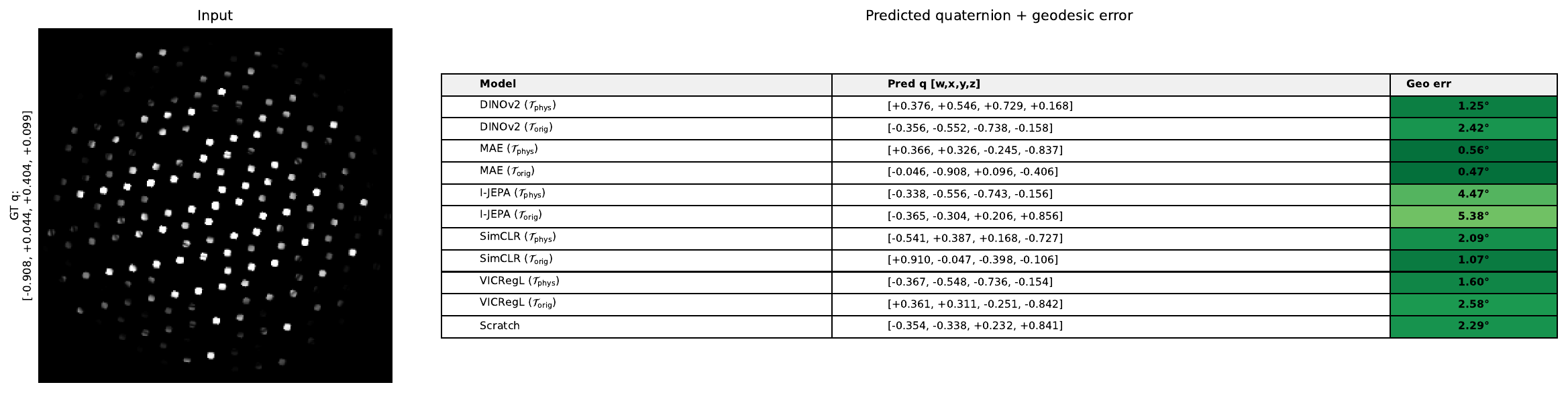}
\end{subfigure}
\vspace{0.4em}

\caption{%
\textbf{Qualitative 4D-STEM quaternion regression examples.}
Each subfigure shows a generated qualitative panel with diffraction inputs and
a per-model table of predicted quaternions and geodesic angular error.
Lower-error entries indicate closer agreement with ground-truth orientation,
enabling direct visual comparison of robustness across pretraining strategies.
}
\label{fig:qual_4dstem}
\end{figure*}

Figure~\ref{fig:qual_4dstem} presents qualitative examples. Each panel contains one diffraction pattern (left) and the
per-model predicted quaternions with geodesic error (right). This view supports
direct per-sample comparison across all methods and highlights where
physics-aligned pretraining reduces angular error on challenging patterns.

\clearpage
\subsection{Full Robustness Results}
\label{sec:supp_robustness}

This section reports the complete robustness evaluation for the 4D-STEM orientation prediction task. We evaluate model performance under controlled perturbations that simulate common sources of acquisition variability in electron microscopy. In particular, we consider global intensity scaling (detector gain variation) and Gaussian blur (resolution degradation).

All models are evaluated without retraining under the same perturbation levels used in the main paper. In addition to the plots presented in the main text, we provide the full numerical results for all methods and perturbation levels.

\subsubsection{Intensity Scaling:}

We first evaluate robustness to global intensity scaling with gain factors $g \in \{0.8, 0.9, 1.0, 1.1, 1.2\}$. Table~\ref{tab:supp_intensity_results} reports the mean geodesic orientation error for each perturbation level.

\begin{table*}[htb!]
\centering
\small
\caption{Robustness to intensity scaling in 4D-STEM orientation prediction. Comparison of mean geodesic error ($^\circ$) and accuracies (\%) across intensity gains. Lower error and higher accuracy indicate better performance.}
\label{tab:supp_intensity_results}
\resizebox{\columnwidth}{!}{%
\small
\begin{tabular}{ll cc cc cc cc cc}
\toprule
 & & \multicolumn{2}{c}{Gain = 0.8} & \multicolumn{2}{c}{Gain = 0.9} & \multicolumn{2}{c}{Gain = 1.0} & \multicolumn{2}{c}{Gain = 1.1} & \multicolumn{2}{c}{Gain = 1.2} \\
\cmidrule(lr){3-4} \cmidrule(lr){5-6} \cmidrule(lr){7-8} \cmidrule(lr){9-10} \cmidrule(lr){11-12}
Method & Aug & Err $\downarrow$ & A@5 $\uparrow$ & Err $\downarrow$ & A@5 $\uparrow$ & Err $\downarrow$ & A@5 $\uparrow$ & Err $\downarrow$ & A@5 $\uparrow$ & Err $\downarrow$ & A@5 $\uparrow$ \\
\midrule
DINOv2 & $\mathcal{T}_{\text{phys}}$ & 5.27 & 91.12 & 4.99 & 92.19 & 4.93 & 92.41 & 5.06 & 92.28 & 5.37 & 91.52 \\
 & $\mathcal{T}_{\text{orig}}$ & 11.29 & 70.80 & 10.77 & 72.45 & 10.61 & 73.11 & 11.00 & 72.66 & 11.77 & 71.12 \\
\midrule
I-JEPA & $\mathcal{T}_{\text{phys}}$ & 10.31 & 76.52 & 10.17 & 77.41 & 10.12 & 77.86 & 10.24 & 77.39 & 10.42 & 76.69 \\
 & $\mathcal{T}_{\text{orig}}$ & 12.54 & 63.30 & 12.07 & 65.39 & 11.96 & 65.66 & 12.15 & 65.09 & 12.92 & 63.13 \\
\midrule
MAE & $\mathcal{T}_{\text{phys}}$ & 1.91 & 98.56 & 1.84 & 98.67 & 1.83 & 98.71 & 1.84 & 98.66 & 1.92 & 98.55 \\
 & $\mathcal{T}_{\text{orig}}$ & 2.01 & 98.25 & 1.95 & 98.31 & 1.95 & 98.33 & 1.97 & 98.31 & 2.01 & 98.19 \\
\midrule
SimCLR & $\mathcal{T}_{\text{phys}}$ & 11.33 & 76.45 & 10.24 & 78.99 & 10.01 & 79.64 & 10.25 & 79.29 & 10.85 & 78.18 \\
 & $\mathcal{T}_{\text{orig}}$ & 10.97 & 78.41 & 10.86 & 79.30 & 10.83 & 79.61 & 10.87 & 79.53 & 11.13 & 79.05 \\
\midrule
VICRegL & $\mathcal{T}_{\text{phys}}$ & 8.88 & 81.03 & 8.37 & 82.84 & 8.27 & 83.43 & 8.55 & 82.93 & 9.24 & 81.38 \\
 & $\mathcal{T}_{\text{orig}}$ & 8.77 & 80.53 & 8.26 & 82.16 & 8.14 & 82.69 & 8.30 & 82.29 & 9.08 & 80.64 \\
\bottomrule
\end{tabular}
}
\end{table*}

\begin{figure*}[htb!]
\centering
\begin{subfigure}{0.49\textwidth}
\centering
\includegraphics[width=\linewidth]{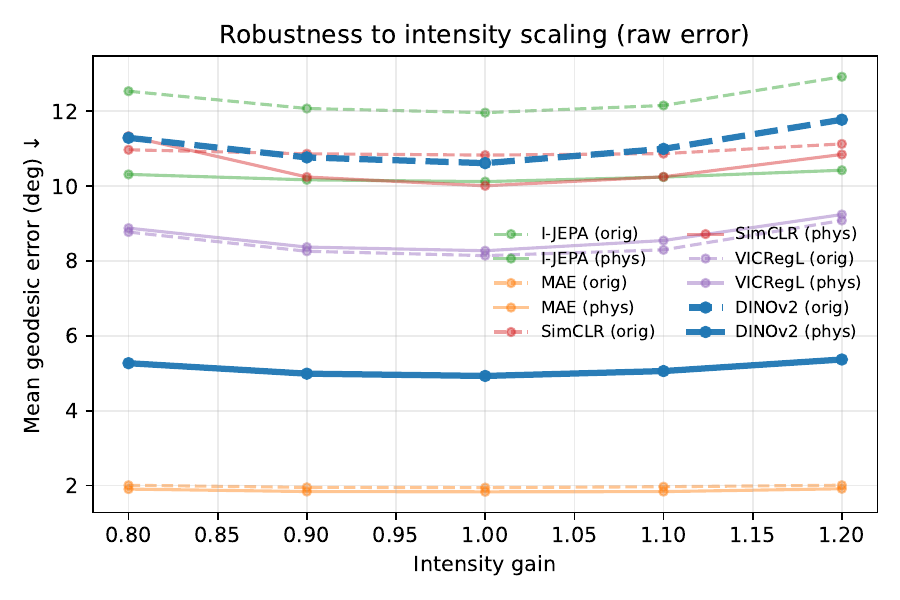}
\caption{Raw geodesic error under intensity scaling.}
\end{subfigure}
\hfill
\begin{subfigure}{0.49\textwidth}
\centering
\includegraphics[width=\linewidth]{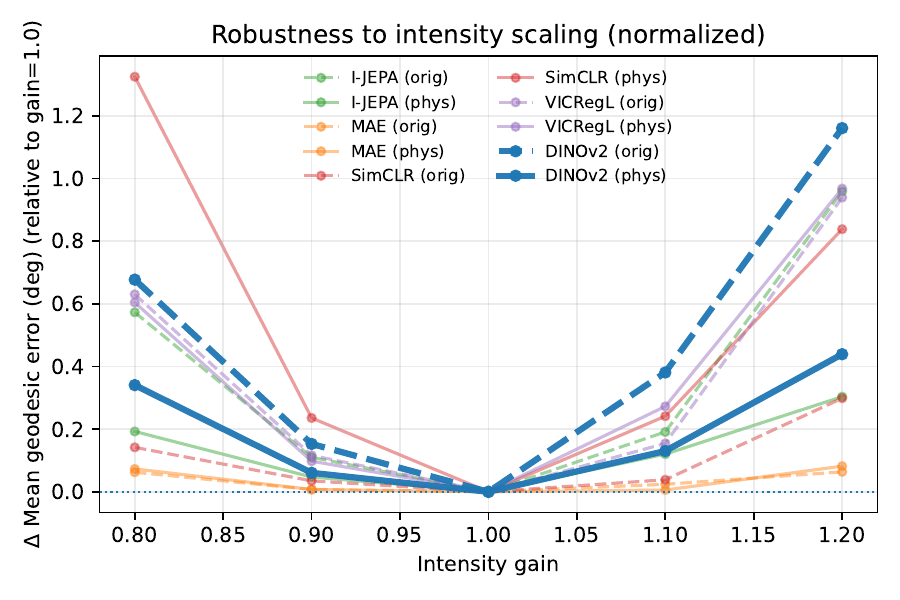}
\caption{Normalized degradation relative to the nominal gain ($g=1.0$).}
\end{subfigure}
\caption{Robustness to detector gain variation (4D-STEM orientation prediction).}
\label{fig:supp_robustness_intensity}
\end{figure*}

\subsubsection{Blur Robustness:}

We next evaluate robustness to Gaussian blur with $\sigma \in \{0, 0.5, 1.0, 1.5, 2.0\}$. Table~\ref{tab:supp_blur_results} reports the full numerical results.

\begin{table*}[htb!]
\centering
\caption{Robustness to Gaussian blur in 4D-STEM orientation prediction. Comparison of mean geodesic error ($^\circ$) and accuracies (\%) across blur levels. Lower error and higher accuracy indicate better performance.}
\label{tab:supp_blur_results}
\resizebox{\columnwidth}{!}{%
\begin{tabular}{ll cc cc cc cc cc}
\toprule
 & & \multicolumn{2}{c}{Blur = 0.0} & \multicolumn{2}{c}{Blur = 0.5} & \multicolumn{2}{c}{Blur = 1.0} & \multicolumn{2}{c}{Blur = 1.5} & \multicolumn{2}{c}{Blur = 2.0} \\
\cmidrule(lr){3-4} \cmidrule(lr){5-6} \cmidrule(lr){7-8} \cmidrule(lr){9-10} \cmidrule(lr){11-12}
Method & Aug & Err $\downarrow$ & A@5 $\uparrow$ & Err $\downarrow$ & A@5 $\uparrow$ & Err $\downarrow$ & A@5 $\uparrow$ & Err $\downarrow$ & A@5 $\uparrow$ & Err $\downarrow$ & A@5 $\uparrow$ \\
\midrule
DINOv2 & $\mathcal{T}_{\text{phys}}$ & 4.93 & 92.41 & 4.97 & 92.34 & 8.03 & 81.86 & 12.66 & 65.21 & 15.75 & 56.31 \\
 & $\mathcal{T}_{\text{orig}}$ & 10.61 & 73.11 & 10.90 & 72.23 & 16.00 & 52.91 & 22.78 & 32.19 & 26.07 & 23.59 \\
\midrule
I-JEPA & $\mathcal{T}_{\text{phys}}$ & 10.12 & 77.86 & 10.19 & 77.52 & 12.44 & 66.15 & 15.37 & 51.21 & 17.34 & 43.09 \\
 & $\mathcal{T}_{\text{orig}}$ & 11.96 & 65.66 & 12.18 & 64.53 & 16.91 & 45.87 & 22.08 & 30.51 & 24.54 & 24.63 \\
\midrule
MAE & $\mathcal{T}_{\text{phys}}$ & 1.83 & 98.71 & 1.81 & 98.76 & 2.27 & 97.48 & 4.00 & 92.85 & 5.71 & 88.49 \\
 & $\mathcal{T}_{\text{orig}}$ & 1.95 & 98.33 & 1.95 & 98.41 & 2.67 & 96.54 & 5.00 & 87.28 & 7.00 & 79.25 \\
\midrule
SimCLR & $\mathcal{T}_{\text{phys}}$ & 10.01 & 79.64 & 10.12 & 79.07 & 15.54 & 63.95 & 22.11 & 45.20 & 25.32 & 35.77 \\
 & $\mathcal{T}_{\text{orig}}$ & 10.83 & 79.61 & 10.97 & 79.20 & 13.91 & 70.36 & 18.41 & 55.55 & 21.24 & 46.24 \\
\midrule
VICRegL & $\mathcal{T}_{\text{phys}}$ & 8.27 & 83.43 & 8.50 & 82.94 & 12.29 & 68.86 & 17.18 & 51.70 & 20.03 & 43.18 \\
 & $\mathcal{T}_{\text{orig}}$ & 8.14 & 82.69 & 8.28 & 81.72 & 12.26 & 66.01 & 18.65 & 43.72 & 22.44 & 32.73 \\
\bottomrule
\end{tabular}
}
\end{table*}

\begin{figure*}[htb!]
\centering
\begin{subfigure}{0.49\textwidth}
\centering
\includegraphics[width=\linewidth]{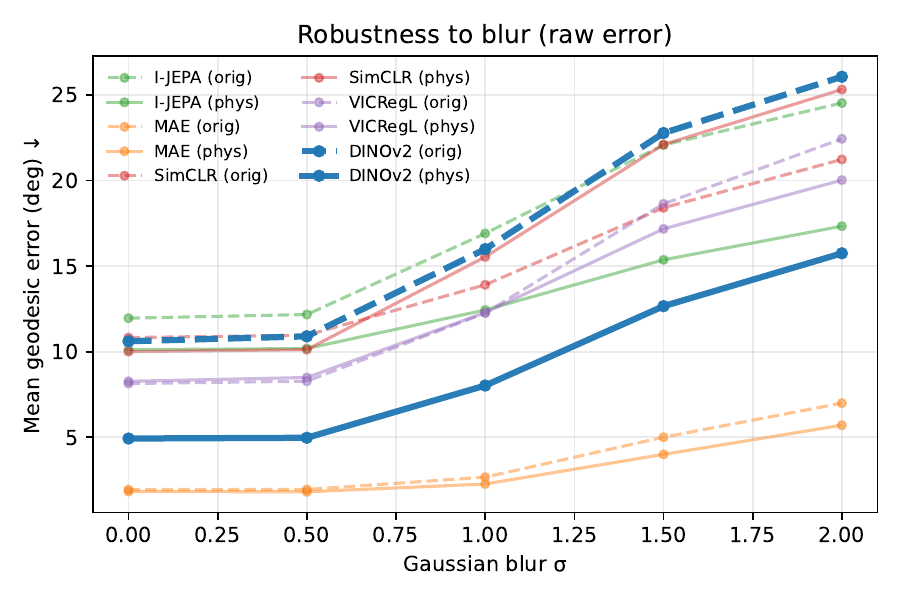}
\caption{Raw geodesic error under increasing blur.}
\end{subfigure}
\hfill
\begin{subfigure}{0.49\textwidth}
\centering
\includegraphics[width=\linewidth]{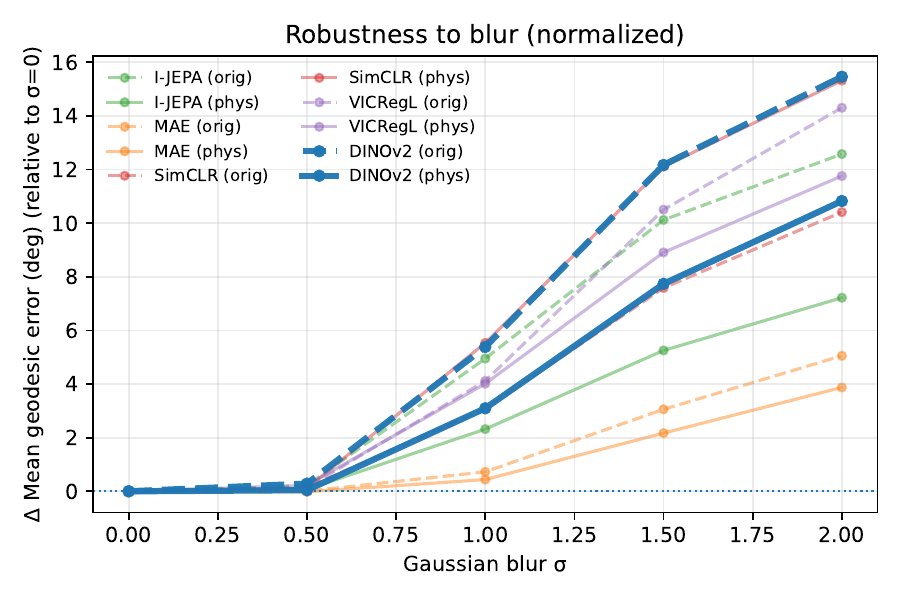}
\caption{Normalized degradation relative to $\sigma=0$.}
\end{subfigure}
\caption{Robustness to resolution degradation modeled by Gaussian blur (4D-STEM orientation prediction).}
\label{fig:supp_robustness_blur}
\end{figure*}

\section{Analysis}

\subsection{Full Representation Geometry Tables}
\label{sec:supp_geometry_tables}

Tables~\ref{tab:nffa_repr_geom} and~\ref{tab:4dstem_repr_geom} report the
complete representation-geometry metrics summarized in the main paper, for NFFA
classification and 4D-STEM orientation regression respectively.

\begin{table}[htb!]
\centering
\small
\caption{Representation geometry analysis on NFFA embeddings. We report effective rank (higher indicates broader spectral utilization), embedding uniformity~\cite{wang2020alignment} (more negative indicates stronger dispersion), collapse ratio (lower indicates less dimensional collapse), and kNN accuracy. Uniformity magnitudes are set by each objective's loss (e.g., VICRegL's covariance term drives strongly negative values) and are therefore comparable across augmentation regimes within a method, not across methods.}
\label{tab:nffa_repr_geom}
\setlength{\tabcolsep}{4.5pt}
\begin{tabular}{lccccc}
\toprule
Method & Aug & Eff.\ Rank $\uparrow$ & Uniformity $\downarrow$ & Collapse $\downarrow$ & kNN $\uparrow$ \\
\midrule
DINOv2 & $\mathcal{T}_{\text{phys}}$ & 1.8 & -0.003 & 0.917 & 0.499 \\
       & $\mathcal{T}_{\text{orig}}$ & 1.4 & -0.000 & 0.964 & 0.552 \\
\midrule
I-JEPA & $\mathcal{T}_{\text{phys}}$ & 4.7 & -0.691 & 0.519 & 0.392 \\
       & $\mathcal{T}_{\text{orig}}$ & 4.5 & -0.581 & 0.557 & 0.389 \\
\midrule
MAE    & $\mathcal{T}_{\text{phys}}$ & 36.6 & -0.096 & 0.443 & 0.569 \\
       & $\mathcal{T}_{\text{orig}}$ & 29.5 & -0.050 & 0.500 & 0.585 \\
\midrule
SimCLR & $\mathcal{T}_{\text{phys}}$ & 80.6 & -0.513 & 0.242 & 0.707 \\
       & $\mathcal{T}_{\text{orig}}$ & 121.3 & -0.644 & 0.195 & 0.679 \\
\midrule
VICRegL& $\mathcal{T}_{\text{phys}}$ & 298.9 & -2.437 & 0.053 & 0.738 \\
       & $\mathcal{T}_{\text{orig}}$ & 382.8 & -2.750 & 0.040 & 0.742 \\
\bottomrule
\end{tabular}
\end{table}

\begin{table}[htb!]
\centering
\caption{Representation geometry analysis on 4D-STEM embeddings used for quaternion regression. As in the NFFA table, uniformity magnitudes are objective-dependent and should be read within a method, not across methods.}
\label{tab:4dstem_repr_geom}
\setlength{\tabcolsep}{4.5pt}
\begin{tabular}{lcccc}
\toprule
Method & Aug & Eff.\ Rank $\uparrow$ & Uniformity $\downarrow$ & Collapse $\downarrow$ \\
\midrule
DINOv2 & $\mathcal{T}_{\text{phys}}$ & 3.1 & -0.010 & 0.830 \\
       & $\mathcal{T}_{\text{orig}}$ & 1.2 & -0.000 & 0.975 \\
\midrule
I-JEPA & $\mathcal{T}_{\text{phys}}$ & 1.7 & -0.007 & 0.927 \\
       & $\mathcal{T}_{\text{orig}}$ & 2.1 & -0.001 & 0.910 \\
\midrule
MAE    & $\mathcal{T}_{\text{phys}}$ & 62.9 & -0.056 & 0.410 \\
       & $\mathcal{T}_{\text{orig}}$ & 84.3 & -0.198 & 0.317 \\
\midrule
SimCLR & $\mathcal{T}_{\text{phys}}$ & 118.8 & -0.829 & 0.173 \\
       & $\mathcal{T}_{\text{orig}}$ & 91.7 & -1.310 & 0.150 \\
\midrule
VICRegL& $\mathcal{T}_{\text{phys}}$ & 269.6 & -2.995 & 0.040 \\
       & $\mathcal{T}_{\text{orig}}$ & 224.5 & -2.869 & 0.047 \\
\bottomrule
\end{tabular}
\end{table}

\subsection{Representation Collapse vs Downstream Performance}

\begin{figure}[htb!]
\centering
\includegraphics[width=0.6\linewidth]{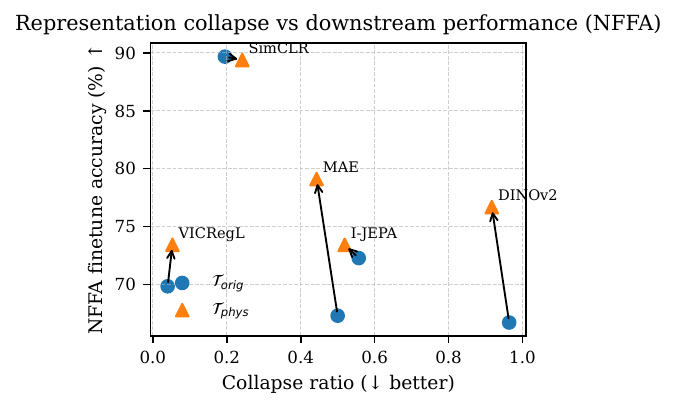}
\caption{Relationship between representation collapse and downstream performance on NFFA classification. Each point corresponds to a pretrained model under natural-image ($\mathcal{T}_{\text{orig}}$) or physics-aligned ($\mathcal{T}_{\text{phys}}$) augmentations.}
\label{fig:geometry_performance}
\end{figure}

We further analyze the relationship between representation geometry and downstream task performance. Figure~\ref{fig:geometry_performance} plots the collapse ratio of the learned embeddings against NFFA classification accuracy for each method and augmentation regime.

Each point corresponds to a pretrained model evaluated using the frozen embedding representation. While some degree of dimensional compression occurs for all methods, models with more balanced feature spectra generally achieve higher downstream accuracy. This observation is consistent with the representation geometry analysis presented in the main paper.

\subsection{Training Dynamics}
\label{sec:supp_training_dynamics}

\begin{figure}[htb!]
\centering
  \includegraphics[width=\linewidth]{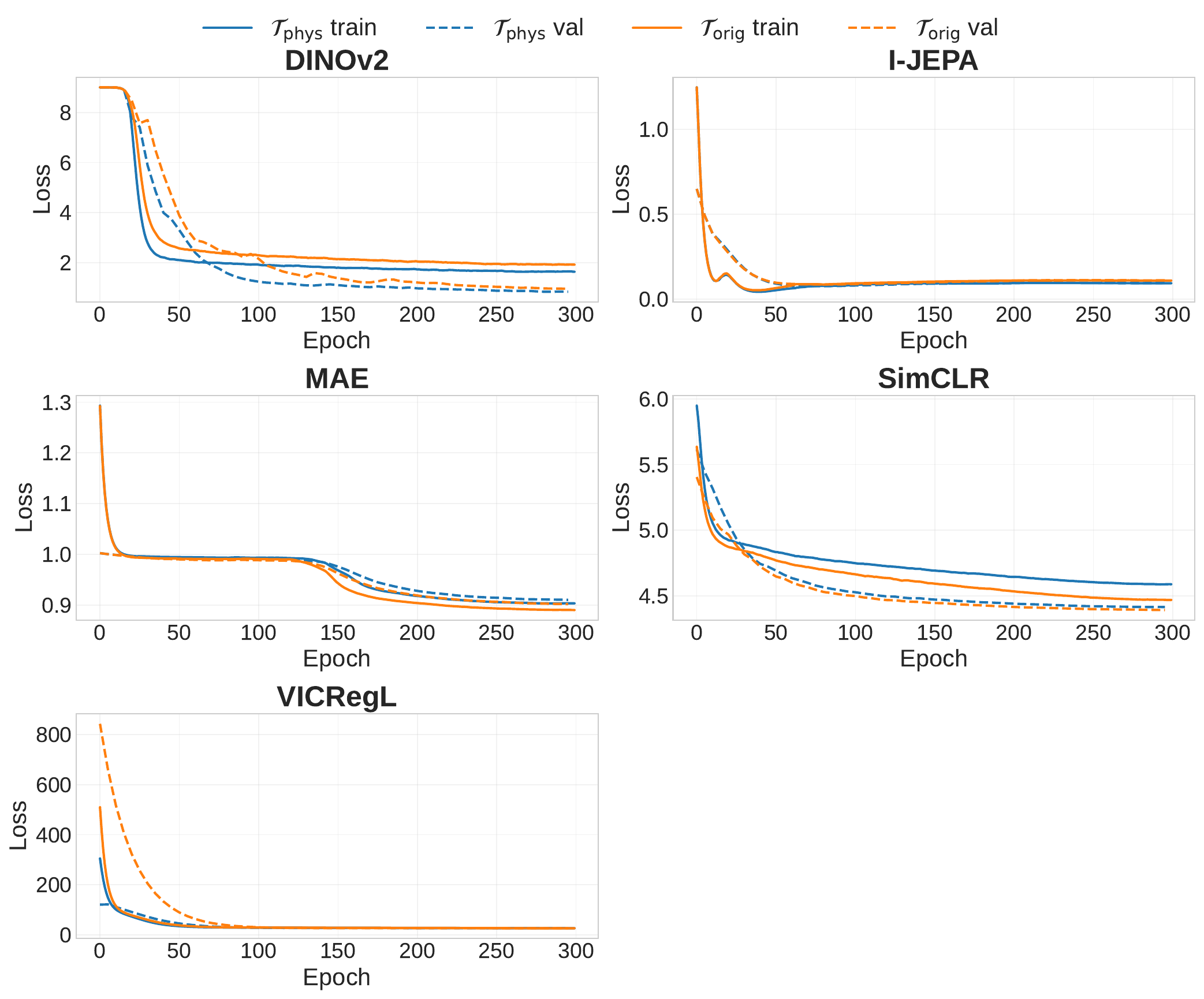}
\caption{%
\textbf{SSL pretraining loss dynamics on CEM500K.}
Train (solid) and validation (dashed) loss for
$\mathcal{T}_{\mathrm{phys}}$ and $\mathcal{T}_{\mathrm{orig}}$ across methods.
Despite stronger augmentations, $\mathcal{T}_{\mathrm{phys}}$ generally reaches
lower loss trajectories.
}  \label{fig:ssl_loss_cem500k}
\end{figure}

\begin{figure}[htb!]
\centering
  \includegraphics[width=\linewidth]{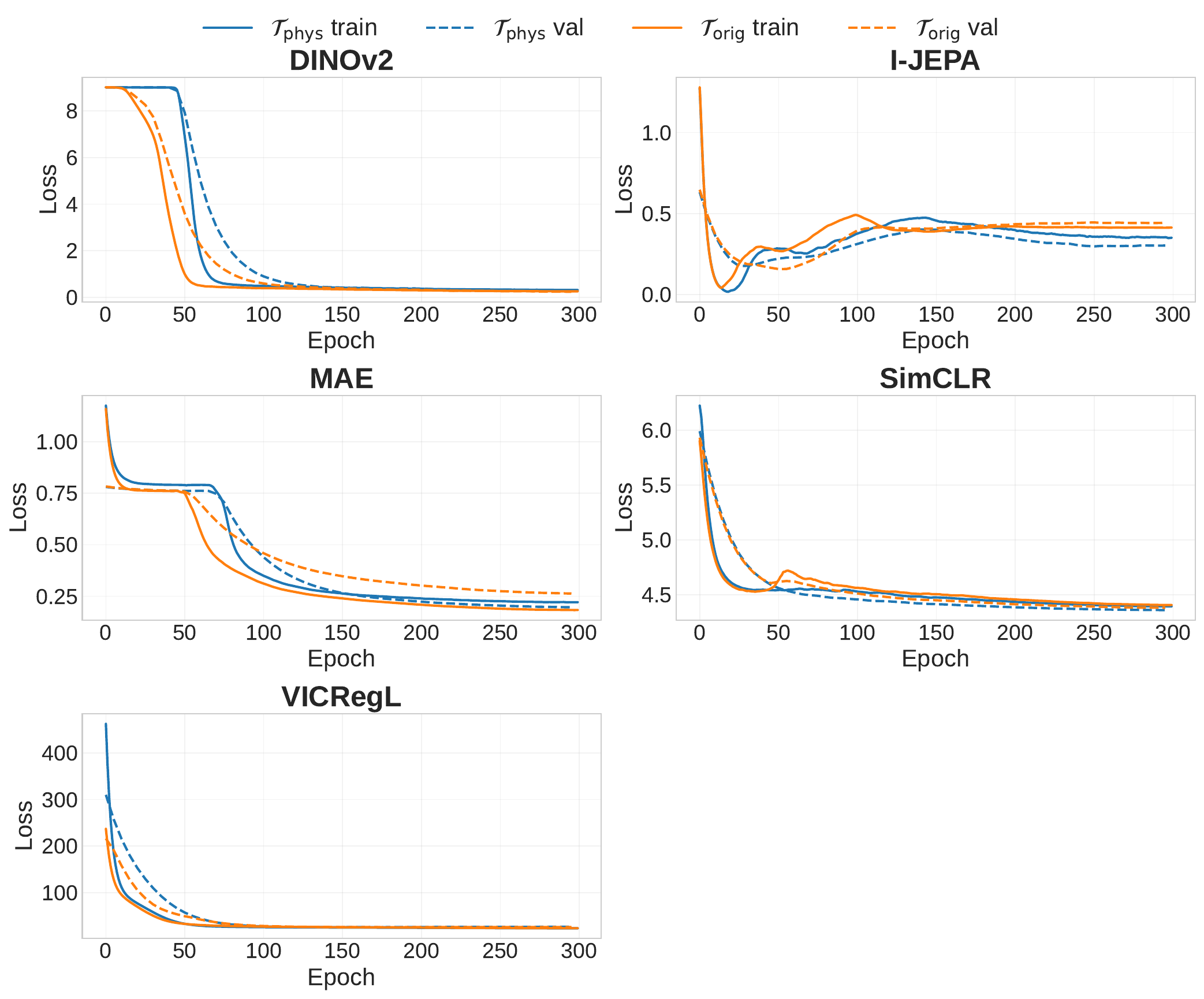}
\caption{%
\textbf{SSL pretraining loss dynamics on 4D-STEM.}
Train (solid) and validation (dashed) loss for
$\mathcal{T}_{\mathrm{phys}}$ and $\mathcal{T}_{\mathrm{orig}}$ across methods.
Despite stronger augmentations, $\mathcal{T}_{\mathrm{phys}}$ generally reaches
lower loss trajectories.
}  \label{fig:ssl_loss_4dstem}
\end{figure}

We next compare optimization dynamics under natural-image augmentations
($\mathcal{T}_{\text{orig}}$) versus physics-aligned augmentations
($\mathcal{T}_{\text{phys}}$).
Although $\mathcal{T}_{\text{phys}}$ introduces stronger distortions and thus
defines a harder pretraining objective, the SSL curves in
Figures~\ref{fig:ssl_loss_cem500k} and~\ref{fig:ssl_loss_4dstem} show that
$\mathcal{T}_{\text{phys}}$ achieves lower (or comparable) train and
validation loss for most backbones on both CEM500K and 4D-STEM.
This indicates that physically grounded augmentations improve representation
learning efficiency rather than merely increasing task difficulty.

\begin{figure}[htb!]
\centering
  \includegraphics[width=\linewidth]{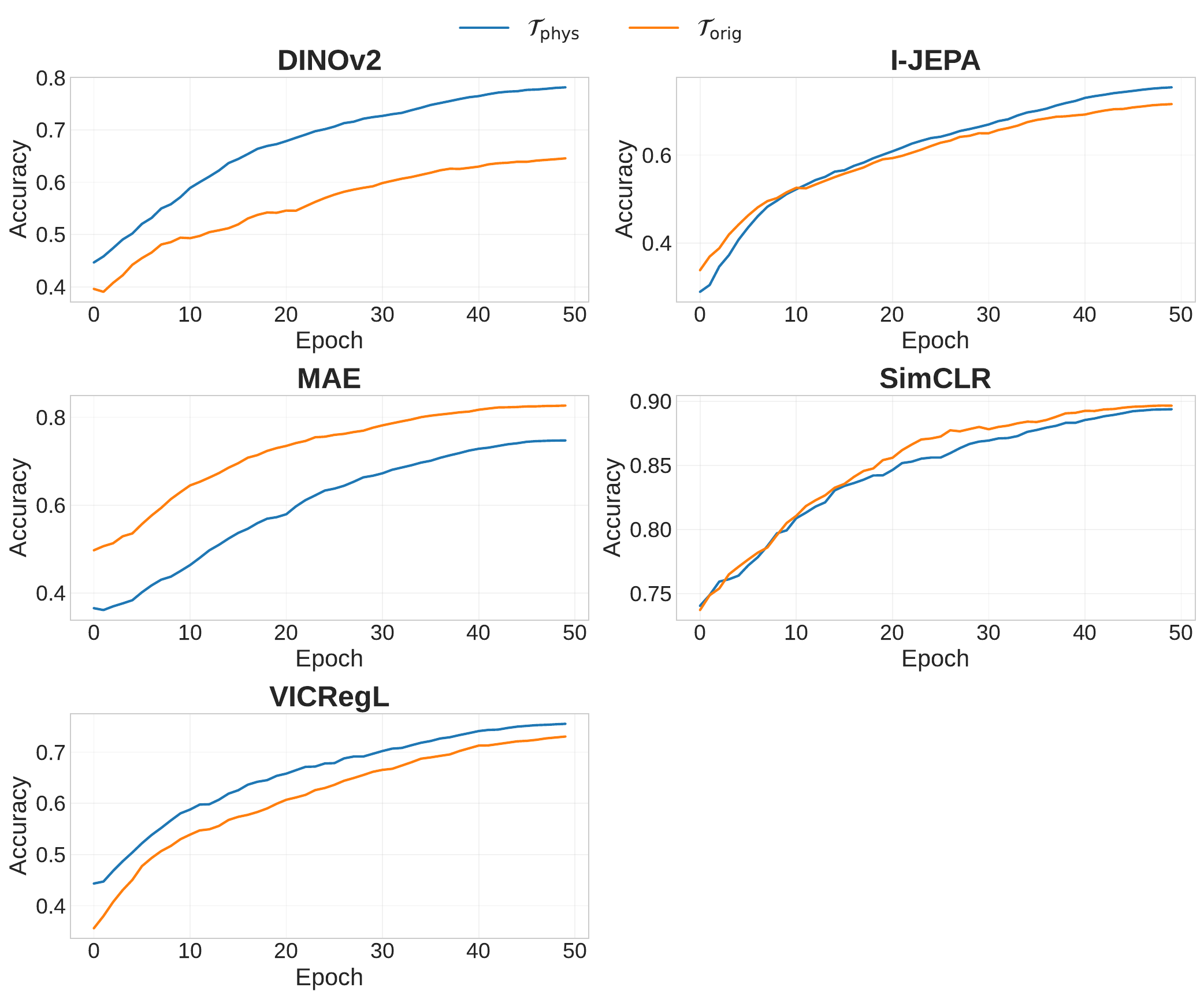}
\caption{%
\textbf{Downstream NFFA validation accuracy dynamics.}
Validation accuracy during finetuning (seed~42), comparing
$\mathcal{T}_{\mathrm{phys}}$ and $\mathcal{T}_{\mathrm{orig}}$ initializations
for each backbone.
}  \label{fig:downstream_nffa_acc}
\end{figure}

\begin{figure}[htb!]
\centering
  \includegraphics[width=\linewidth]{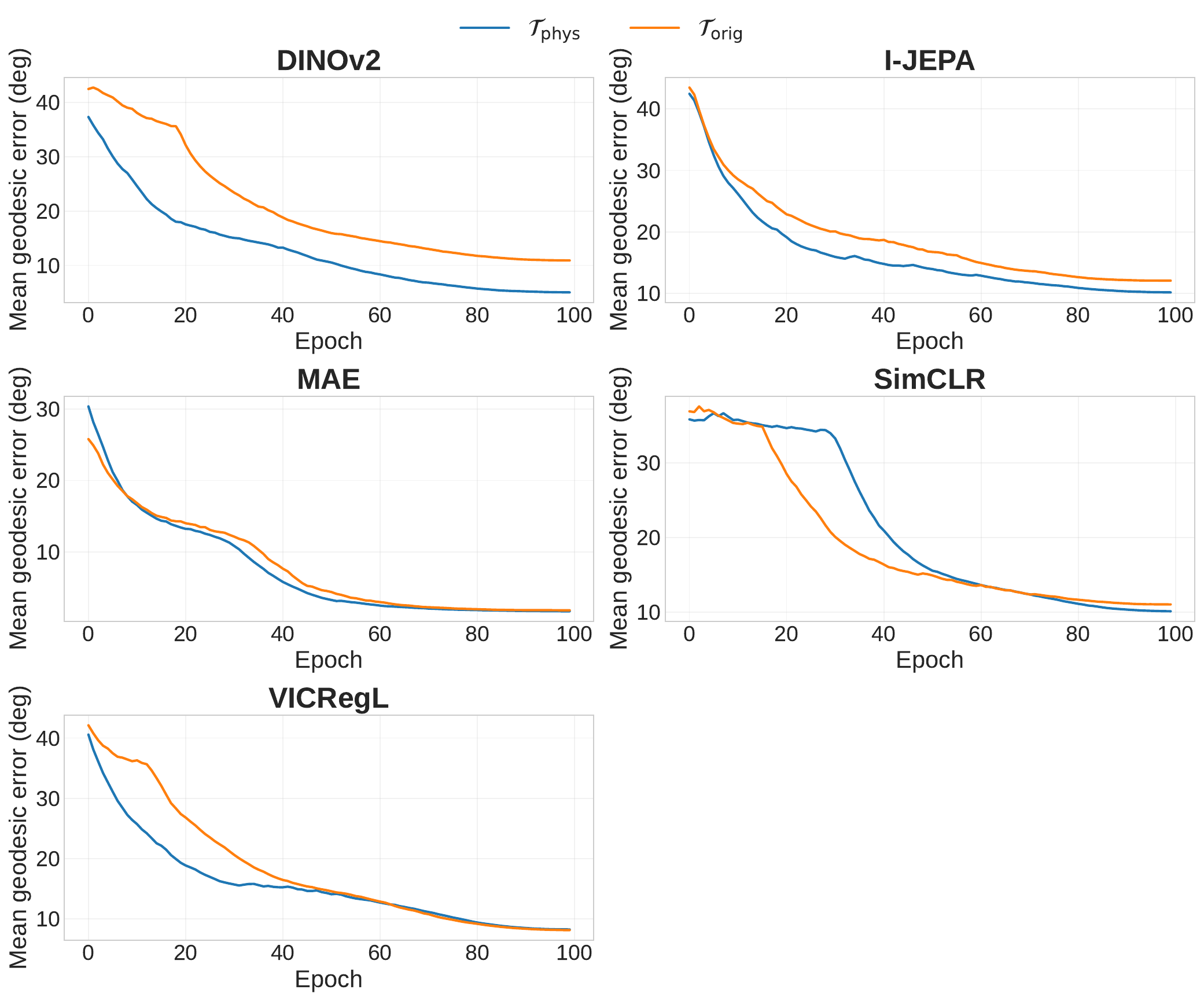}
\caption{%
\textbf{Downstream 4D-STEM validation mean geodesic error dynamics.}
Lower is better. Across methods, $\mathcal{T}_{\mathrm{phys}}$ typically
converges to lower angular error.
}  \label{fig:downstream_quat_geo}
\end{figure}

\begin{figure}[htb!]
\centering
  \includegraphics[width=\linewidth]{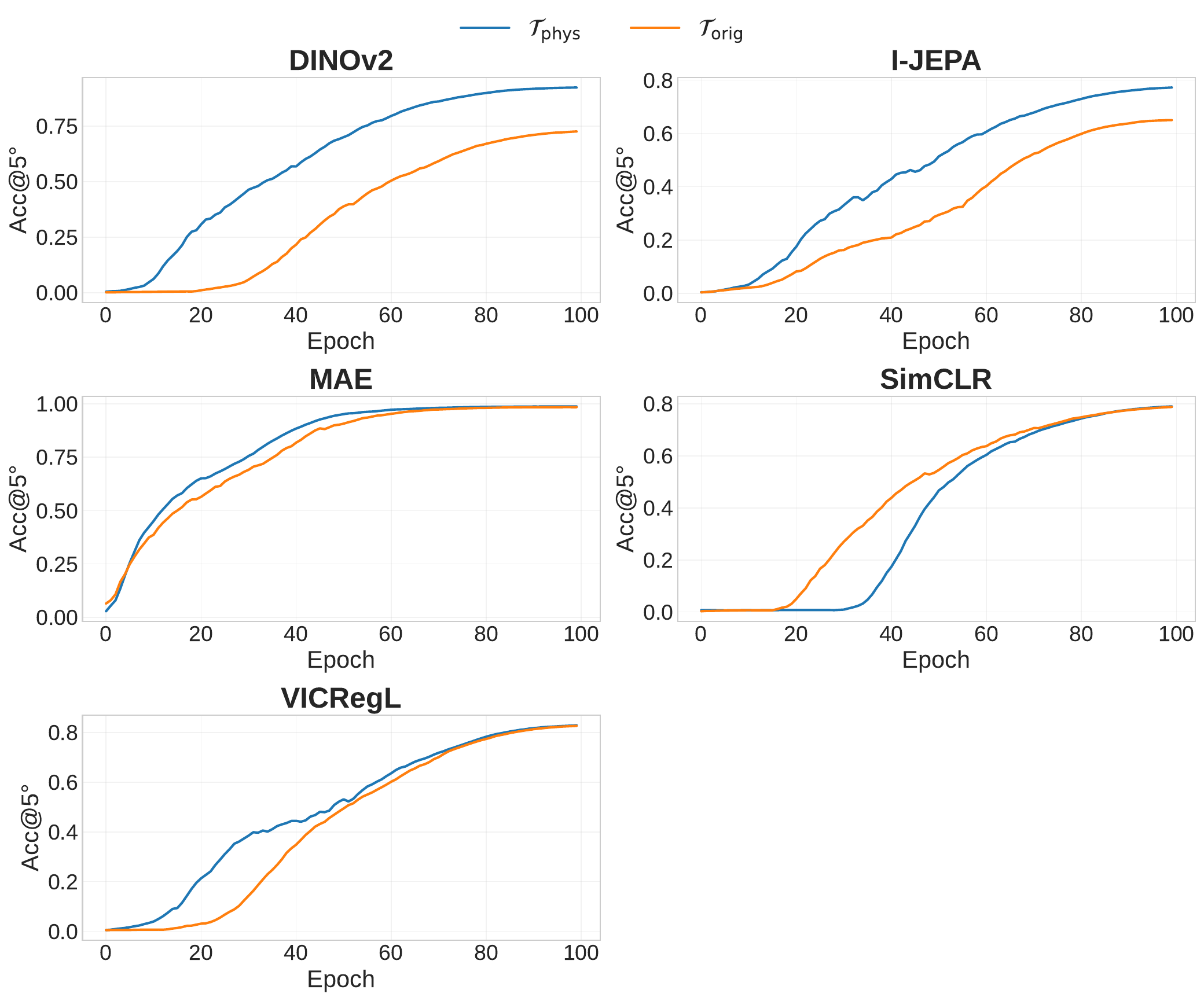}
\caption{%
\textbf{Downstream 4D-STEM validation Acc@5$^\circ$ dynamics.}
Higher is better. $\mathcal{T}_{\mathrm{phys}}$ yields consistently stronger
high-precision orientation accuracy.
} \label{fig:downstream_quat_acc5}
\end{figure}

\begin{figure}[htb!]
\centering
  \includegraphics[width=\linewidth]{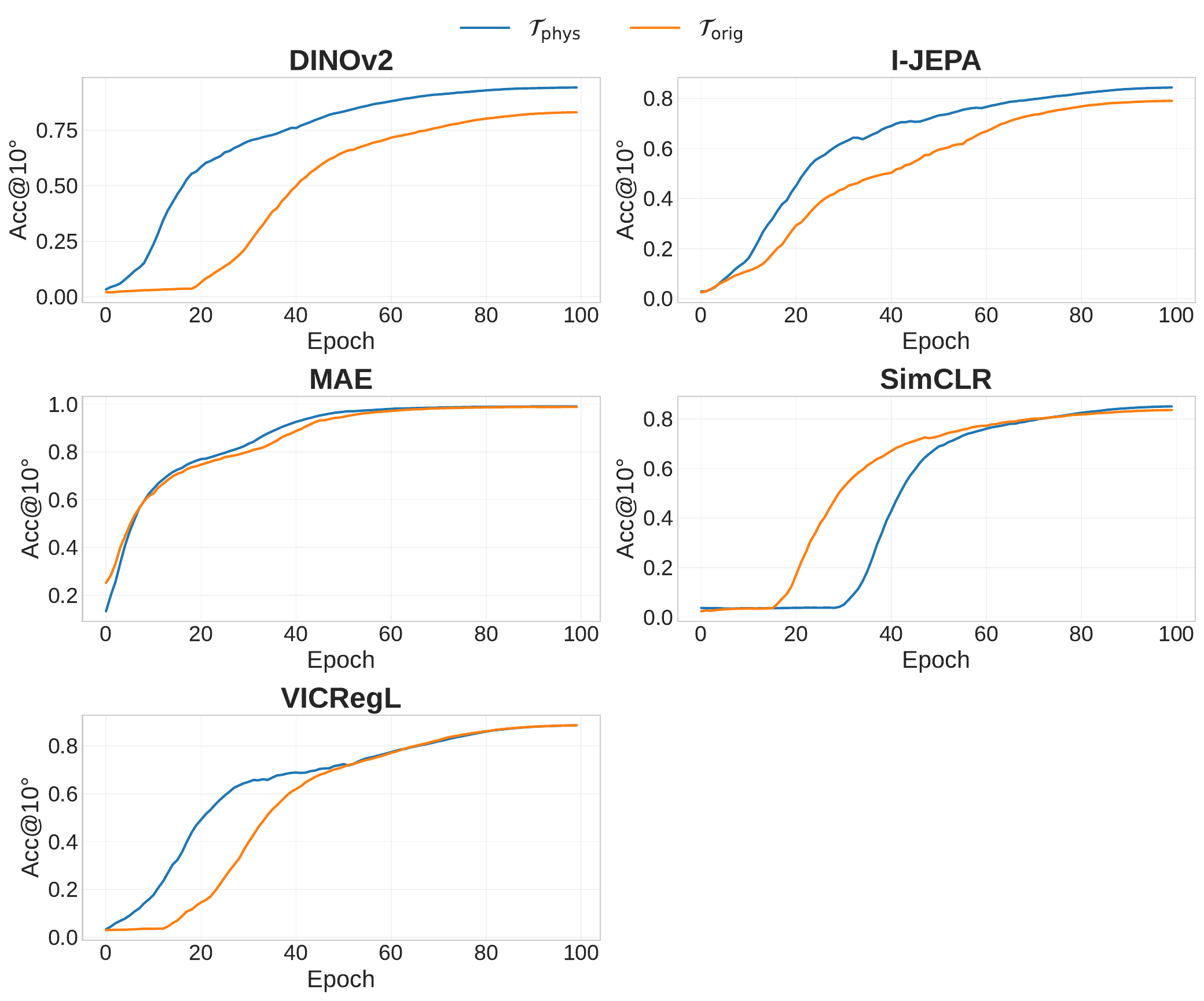}
\caption{%
\textbf{Downstream 4D-STEM validation Acc@10$^\circ$ dynamics.}
Higher is better. The improvement trend of
$\mathcal{T}_{\mathrm{phys}}$ persists at the wider angular threshold.
}\label{fig:downstream_quat_acc10}
\end{figure}

In downstream finetuning (seed~42), we observe task-dependent transfer
behavior consistent with the final metrics. For 4D-STEM quaternion regression,
the validation geodesic error and angular accuracies
(Acc@5$^\circ$, Acc@10$^\circ$) consistently favor
$\mathcal{T}_{\text{phys}}$ across methods
(Figures~\ref{fig:downstream_quat_geo}--\ref{fig:downstream_quat_acc10}).
For NFFA classification, the validation-accuracy dynamics
(Figure~\ref{fig:downstream_nffa_acc}) reveal that MAE and SimCLR are notable
cases where $\mathcal{T}_{\text{orig}}$ can match or exceed
$\mathcal{T}_{\text{phys}}$, aligning with the endpoint comparisons in
Table~\ref{tab:nffa_acc}.

Overall, the dynamics support a consistent picture: physically aligned
augmentations improve both optimization and transfer in geometry-sensitive
settings, yielding lower SSL losses and stronger downstream orientation
prediction, while NFFA remains partially method-dependent with MAE and SimCLR
as notable exceptions where $\mathcal{T}_{\text{orig}}$ is competitive or
slightly better. This pattern suggests that augmentation-task alignment is a
key factor in determining transfer gains.

\end{document}